\documentclass[10pt,twocolumn,letterpaper]{article}

\usepackage[pagenumbers]{cvpr}      % To produce the REVIEW version
\definecolor{cvprblue}{rgb}{0.21,0.49,0.74}
\usepackage[pagebackref,breaklinks,colorlinks,allcolors=cvprblue]{hyperref}

\usepackage{xspace}
\usepackage[table]{xcolor}
\usepackage{amsmath,bm} % or amsmath alone if you use \boldsymbol
\usepackage{multirow}
\usepackage{booktabs}
\usepackage{colortbl}
\usepackage{graphicx}
\usepackage{arydshln}
\usepackage{tikz}
\usepackage{pgf-pie}
\usepackage{makecell}
\usepackage{marvosym}       

\makeatletter
\def\@fnsymbol#1{%
  \ifcase#1%
    \or *%              1
    \or \Letter%        2
    \or \ddagger%       3
    \or \mathsection%   4
    \or \mathparagraph% 5
    \or \|%             6
    \or **%             7
    \or \dagger\dagger% 8
    \or \ddagger\ddagger% 9
  \else\@ctrerr\fi}
\makeatother

\usepackage{cuted}   % 提供 strip 环境
\usepackage{capt-of} % 提供 \captionof

\definecolor{citecolor}{HTML}{0071bc}

\definecolor{scolor}{RGB}{111,168,220}
\definecolor{hcolor}{RGB}{111,176,81}
\definecolor{ocolor}{RGB}{224,103,102}
\definecolor{wcolor}{RGB}{246,178,107}
\newcommand{\SHOW}{\textcolor{scolor}{S}\textcolor{hcolor}{h}\textcolor{ocolor}{o}\textcolor{wcolor}{w}UI-$\pi$}
\newcommand{\our}{ShowUI-$\pi$\xspace}
\newcommand{\bench}{ScreenDrag\xspace}

\renewcommand{\eg}{\textit{e.g.,}}
\newcolumntype{L}[1]{>{\raggedright\arraybackslash}m{#1}}
\newcolumntype{C}[1]{>{\centering\arraybackslash}m{#1}}

\definecolor{HeaderGray}{HTML}{F2F2F2}   % header cell behind "Methods"
\definecolor{BaselineGray}{HTML}{F5F6F7} % pale gray for predefined-JSON baselines
\definecolor{BandBlue}{HTML}{E8F3FF}     % light-blue band for the last (ours) row
\definecolor{OursBlue}{HTML}{0B5394}     % dark blue for our method name
\definecolor{RuleGray}{HTML}{C4C4C4}     % neutral gray for rules/dashed lines
\definecolor{LabelGray}{HTML}{6B7280}    % muted gray for subsection labels
\newcommand{\best}[1]{\textbf{#1}}

\def\paperID{1917} % *** Enter the Paper ID here
\def\confName{CVPR}
\def\confYear{2026}

\title{\SHOW: Flow-based Generative Models as GUI Dexterous Hands}

\author{%
    Siyuan Hu\footnotemark[1]\quad
    Kevin Qinghong Lin\footnotemark[1]\quad
    Mike Zheng Shou\footnotemark[2]
    \vspace{2mm}\\
    Show Lab, National University of Singapore\\
    [2mm]
    \href{https://showlab.github.io/showui-pi}{\texttt{https://showlab.github.io/showui-pi}}\\    
}

\begin{document}
\maketitle
\begingroup
\renewcommand{\thefootnote}{\fnsymbol{footnote}}
\footnotetext[1]{Equal contribution.}
\footnotetext[2]{Corresponding author.}
\endgroup

\begin{strip}
    \centering
    \includegraphics[width=0.99\textwidth]{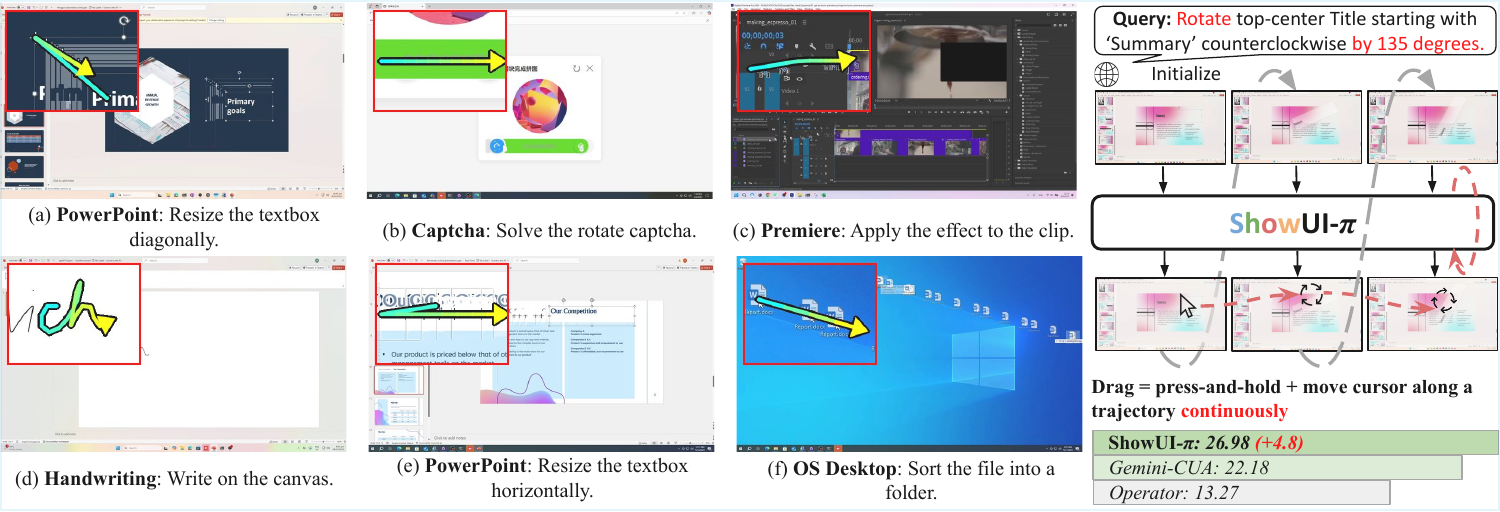}
    % 如果需要 caption / 引用：
    \captionof{figure}{Drag refers to a \emph{continuous} interaction where the cursor maintains contact with the UI element while moving along a trajectory, rather than a single \emph{discrete} click. \textbf{Left: Visualization of \bench data domains.} \textbf{Right: \our is a lightweight flow-based generative model for GUI Automation} that handles dragging actions requiring on-the-fly observation, such as drawing and Captcha solving. 
    Given a query, \our efficiently generates corresponding continuous trajectory from streaming visual observations.}
    % \label{fig:teaser}
\end{strip}

\begin{abstract}
Building intelligent agents capable of dexterous manipulation is essential for achieving human-like automation in both robotics and digital environments.
However, existing GUI agents rely on discrete click predictions $(x,y)$, which prohibits free-form, closed-loop trajectories (\eg~dragging a progress bar) that require continuous, on-the-fly perception and adjustment.
In this work, we develop \textup{\our}, the first flow-based generative model as GUI dexterous hand,
featuring the following designs:
\textbf{(i) Unified Discrete–Continuous Actions}, integrating discrete clicks and continuous drags within a shared model, enabling flexible adaptation across diverse interaction modes;
\textbf{(ii) Flow-based Action Generation} for drag modeling, which predicts incremental cursor adjustments from continuous visual observations via a lightweight action expert, ensuring smooth and stable trajectories;
\textbf{(iii) Drag Training data and Benchmark}, where we manually collect and synthesize 20K drag trajectories across five domains (\eg~PowerPoint, Adobe Premiere Pro), and introduce ScreenDrag, a benchmark with comprehensive online and offline evaluation protocols for assessing GUI agents’ drag capabilities.
Our experiments show that proprietary GUI agents still struggle on ScreenDrag (\eg~Operator scores 13.27, and the best Gemini-2.5-CUA reaches 22.18). In contrast, \our achieves 26.98 with \textbf{only 450M parameters}, underscoring both the difficulty of the task and the effectiveness of our approach. 
We hope this work advances GUI agents toward human-like dexterous control in digital world. The code is available at {\url{https://github.com/showlab/showui-pi}}. 
\end{abstract}
        
\section{Introduction}
Building intelligent assistants capable of dexterous manipulation is essential for achieving human-like automation in both physical and digital environments~\cite{showui,openvla,magma}. For example, in the physical world, robotic dexterous hands are deployed for manipulation tasks \eg, sorting objects on physical desktops~\cite{rt1,rt2,pi_0,pi_0_5}.
Analogously, in the digital world, Graphical User Interfaces (GUI) serve as the primary medium through which people interact with computers, and automating GUI interactions holds substantial promise for enhancing productivity~\cite{workarena}, reducing workload~\cite{webshop}, and improving accessibility~\cite{mind2web}. 

% \begin{figure}[t]
%     \centering
%     \includegraphics[width=\linewidth]{figure/teaser_cropped_big.pdf}
%     \caption{\textbf{\our is a lightweight flow-based generative model for GUI Automation} that handles dragging actions requiring on-the-fly observation, such as drawing and Captcha solving. 
%     Given a query, \our efficiently generates corresponding continuous trajectory from streaming visual observations.}
%     \label{fig:teaser}
%     \vspace{-2em}
% \end{figure}

Recent advances in large language models (LLMs)~\cite{gpt5, claude4.5, webagentplan, webarena, multimodalweb} and vision-language models (VLMs)~\cite{qwen2vl, cogagent, seeclick, uground} have accelerated progress in GUI agents~\cite{showui,uitars1.5}. These emerging visual-centric agents directly perceive screen observations and output actions such as clicking or typing.
However, most existing GUI agents~\cite{aguvis, opencua} are obtained by fine-tuning foundation VLMs without architecture adaptation, which represent action's coordinate in a discrete, tokenized form.
This representation fundamentally restricts agents from executing complex high-degree-of-freedom dragging, such as creative drawing or Captcha-solving by rotation \ie, tasks inherently demanding continuous, real-time visual observation and responsible for incremental trajectory adjustment, 
By contrast, Vision-Language-Action models~\cite{diffusion_policy, pi_0, pi_0_5} in robotics leverage flow-based generative methods (\eg~diffusion policy, flow matching) to enable achieving such continuous, fine-grained control on the fly.

Motivated by how humans control the mouse for fine-grained cursor movement,~\ie, continuously perceive and adjust actions, we wonder: \textit{can we construct such a digital dexterous hand in GUI?}
We propose {\our}, the first flow-based GUI model designed for continuous trajectory. 
Specifically, \our highlights the following architecture:
\textbf{(i) Unified Discrete-Continuous Actions:} \our casts discrete clicks as drags with negligible movements, and integrates them with continuous drags into a unified modeling. Under this formulation, both action types are represented by a sequence of $(x,y,m)$ triplets, where $(x,y)$ are cursor coordinates and $m\in\{\texttt{down},\texttt{up}\}$ is the mouse button state. This unified design allows \our to handle both drag and click tasks with a single shared model, adapting without task-specific head selection.
\textbf{(ii) Flow-based Action Generation:} Different from existing GUI agents predicting discrete, tokenized actions from language decoding,~\eg, \texttt{click(x,y)} and \texttt{drag(start, end)}, \our is a flow-based generative model, and employs a lightweight action expert to incrementally predict cursor adjustments from continuous visual observations. Built on a transformer backbone, the action expert is trained with flow matching to generate stable and precise action trajectories; 
\textbf{(iii) \bench~benchmark:} We construct a benchmark specially for continuous GUI tasks, including 505 real-world drag tasks across five domains covering both professional control and daily usage, such as PowerPoint, OS Desktop and file manager, Handwriting on canvas, Adobe Premiere Pro, and Captcha solving, with 101 tasks for each domain. 
Moreover, to fully evaluate continuous trajectories, we introduce two complementary modes: an offline open-loop evaluation using average trajectory error and endpoint accuracy, and an online closed-loop evaluation using task success rate.
\textbf{(iv) Continuous Trajectory Training Data:} To advance the model training, we construct a training dataset of 20K manually collected and synthesized drag trajectories across the five aforementioned domains and 11 categories of tasks, and all trajectories have recorded UI states and dense coordinates.

Our experiments show that proprietary GUI agents still struggle on ScreenDrag (\eg~Operator scores 13.27, and the best Gemini-2.5-CUA reaches 22.18). In contrast, \our achieves 26.98 with only 450M parameters, underscoring both the difficulty of the task and the effectiveness of our approach. Moreover, our ablation studies further reveal the superficial nature of standard flow-matching training and highlight the substantial impact of individual designs. {Our contributions are three-fold:} 
\begin{itemize}
    \item \textbf{The first flow-based GUI agent for continuous trajectories.} To the best of our knowledge, we are \textit{the first work} to tackle continuous trajectory-based drags in GUI automation, revealing core limitations of discrete, tokenized actions. We propose \our, a lightweight 450M flow-based VLA that unifies discrete clicks and continuous drags within a shared modeling.
    \item \textbf{ScreenDrag training dataset and benchmark suite.} We introduce \bench, a benchmark tailored for continuous GUI manipulation, with 505 real-world drag tasks across five domains and 20K manually collected and synthesized drag trajectories over 11 task categories. \bench supports both offline open-loop metrics and online closed-loop success evaluation.
    \item \textbf{Comprehensive evaluation and key insights.} Experiments show that proprietary GUI agents still struggle on ScreenDrag (\eg, Operator 13.27, best Gemini-2.5-CUA 22.18), while \our reaches 26.98 with only 450M parameters, highlighting both task difficulty and model effectiveness. Ablations further demonstrate the impact of our individual design choices for achieving robust continuous control.
\end{itemize}

\section{Related Work}

\begin{figure*}[!t]
    \centering
    \includegraphics[width=\textwidth]{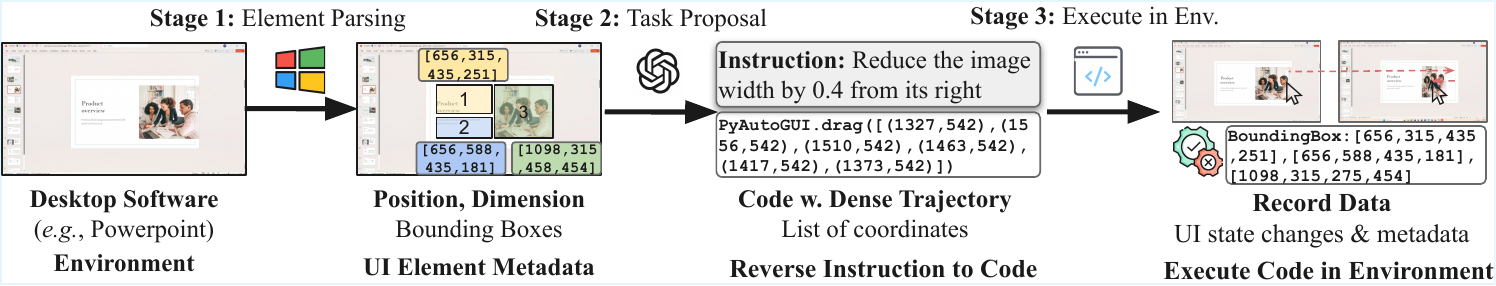}
    \vspace{-2em}
    \caption{\textbf{\bench Automated Data Collection Pipeline.} \bench automated data generation pipeline for continuous trajectory-based GUI interaction data. The pipeline includes three stages: \textbf{(i) Element Parsing:} The software application UI is parsed with UI Automation of Windows SDK in order to retrieve the UI element metadata. \textbf{(ii) Task Proposal:} Given the UI element metadata, an LLM will be prompted to generate a drag instruction, the expected metadata change and the drag code with dense trajectory. \textbf{(iii) Trajectory Synthesis:} The drag code will be executed in the software environment. A rule-based verifier will check the parsed metadata from UI states before and after the drag to ensure that the metadata change satisfies the expectation.}
    \vspace{-2em}
    \label{fig:data}
\end{figure*}

\subsection{Digital GUI Automation.}
LLM-based agents in digital environments have evolved from reasoning and tool-use paradigms, such as Chain-of-Thought~\cite{chain_of_thought}, ReAct~\cite{react}, and related prompting frameworks~\cite{socratic,gpt4tools,assistgpt}, toward task-oriented GUI automation. Existing methods include \emph{training-free} pipelines which plan with a VLM and execute via external tools~\cite{assistgui,seeact}, and \emph{training-based} models jointly learning perception and action from screenshots and instructions~\cite{seeclick,cogagent,ferretui}. 
% Mobile GUI agents have similarly been explored~\cite{gpt4v_wonderland,mobile_agent,aitw}. 
Existing GUI agents, including ShowUI, decode actions into discrete text tokens, simplifying integration with VLM planners but limiting control to simple clicks or short drags. \our departs from these methods by directly modeling continuous spatial trajectories for long, smooth, temporally coherent manipulations.

\subsection{Physical Robotic Manipulation.}
Generalist Vision-Language-Action (VLA) models map natural-language instructions and visual observations to action policies by leveraging large multimodal backbones and robotics data at scale. Representative lines include RT-1~\cite{rt1} and successors~\cite{rt2} which transfer knowledge from web-scale vision-language pretraining to robot control, open-source VLA backbones and compact variants such as OpenVLA~\cite{openvla}, PaliGemma~\cite{paligemma}, and TinyVLA~\cite{tinyvla}, as well as recent works enriching training signals with visual traces and language alignment~\cite{llarva,llara}. 

\noindent\textbf{Discrete Language Models} formulations predict low-level controls from decoded discrete action tokens. Representative models include RT-2~\cite{rt2} and OpenVLA~\cite{openvla}, which discretize continuous actions into 256 bins per dimension and overwrite the least-used tokens in the language model vocabulary to represent these action bins as autoregressively generated tokens; and Magma~\cite{magma}, which applies discrete action tokenization across both UI navigation and robotic manipulation tasks. Language-modeling maintains compatibility with instruction-tuned VLMs but suffers from temporal quantization when fine motor control is needed~\cite{fast}.
% \kevin{we are the first work leverage robotic-like dex hand for GUI agent}

\noindent\textbf{Continuous Generative Models} parameterize a time-indexed transport from a base distribution to action trajectories. Within this paradigm, (i) \emph{Diffusion Policies} instantiate the transport via score-based denoising, demonstrating strong capabilities in visuomotor control due to stable optimization and high-dimensional handling~\cite{denoising,nonequilibrium,diffusion_policy,scalingdp}.
Alternative frameworks such as (ii) \emph{Flow Matching} and \emph{Rectified Flow} directly regress a time-conditioned velocity field along a predefined probability path, eliminating explicit score estimation and iterative denoising for deterministic ODE-based sampling. Rectification further simplifies trajectory modeling and enhances sample efficiency~\cite{flow_matching,rectified_flow}. Such properties have inspired their integration into robotics-oriented flow-matching VLA policies, combining a flow-based action head with VLM backbones for smooth, real-time continuous control~\cite{pi_0,pi_0_5}. The learning objective is:
\begin{equation}
    \mathcal{L}_{\text{flow matching}} = \frac{1}{T}\sum_{t=1}^{T} \| \mathbf{v}_\theta(\hat{\mathbf{a}}_t, t \mid o_t,\mathcal{Q}) - \mathbf{u}_t \|^2,
    \label{eq:flow-matching-loss}
\end{equation}
where $\mathbf{v}_\theta(\hat{\mathbf{a}}_t, t \mid o_t,\mathcal{Q})$ denotes the predicted velocity field at time $t$ conditioned on the current observation $o_t$ and task instruction $\mathcal{Q}$, $\mathbf{u}_t$ is the target velocity along the probability path, $\hat{\mathbf{a}}_t$ is the intermediate action state, and $T$ is the number of training timesteps.
% \sy{what is relationship between u and v}

% This approach significantly reduces control latency by amortizing inference to a few ODE steps per update, seamlessly integrating with large-scale multi-task datasets and evaluation frameworks established by RT-1, RT-2, Octo, and Open X-Embodiment~\cite{rt1,rt2,octo,open_x_embodiment_rt_x}.

% Eq. of flow-based action modeling

\section{\bench Dataset}

% too simple, add details:
% 1.5x height, textbox - screenshot correspondence by bbox
% instruction: 
% - 0.4: show in the trace
% - more natural
% double-col, concrete example

\subsection{Problem Definitions}
% Continuous interaction constitutes a fundamental capability for generalist GUI agents aspiring to human-level automation. Such interaction involves executing dynamic, visually-guided spatial trajectories rather than discrete point-and-click actions. Formally, we define the continuous GUI control task as follows: given an initial GUI observation $o_{0}$ and a natural language instruction $\mathcal{Q}$, an agent policy $\pi$ sequentially predicts a continuous trajectory $\tau = \{a_t\}_{t=0}^{T}$, where each action $a_t=[x,y]$ specifies a spatial coordinate in the GUI. The predicted action sequence autonomously terminates at timestep $T$, aiming to minimize cumulative spatial discrepancies with the ground truth trajectory:
% % \begin{equation}
% % \mathcal{L}(\tau, \tau^{gt}) = \mathbb{E}\left[\sum_{t=0}^{T}\|a_{t}-a_{t}^{gt}\|_{2}\right],
% % \label{eq:trajectory_loss}
% % \end{equation}
% where $a_t^{gt}$ denotes the ground-truth spatial coordinates. This formulation closely aligns with standard trajectory-matching objectives used in recent continuous-control models.

% Existing GUI benchmarks predominantly emphasize discrete, atomic actions and fall short of assessing the nuanced complexity involved in continuous spatial trajectories. To bridge this gap, we introduce \bench, a specialized benchmark explicitly designed to rigorously evaluate GUI agents' capabilities in predicting and executing continuous trajectories.
% \textbf{Challeges} high resolution, fine-grained in-the-fly, compare with simple click, simple drag start/end might ignore
% How to evaluate is still open.

Unlike discrete actions such as clicking, which can be completed with one command based on an initial observation, continuous actions—such as dragging or rotation—require on-the-fly visually grounded trajectories in real time.
% Continuous interaction `drag' is critical for generalist GUI agents aspiring to precise, adaptive, and dexterous automation. 
Formally, given an initial observation $o_{0}$ and a language instruction $\mathcal{Q}$, a GUI agent policy $\pi$ sequentially predicts a continuous trajectory $\tau = \{a_t\}_{t=0}^{T}$, where each action $a_t = [x, y]$ denotes the spatial coordinates at time step $t$.

% This task presents several unique challenges: 
% \textbf{\textit{a}.} In drag scenarios, screen states will change during the interaction and how to keep perceiving the state change is unclear.

However, existing benchmarks reduce drag-like interactions to discrete start–end point pairs and provide only a single screenshot per atomic action, ignoring the rich intermediate state changes that occur during continuous dragging. In addition, many drag scenarios allow multiple valid trajectories to complete a task, yet this flexibility is not captured in current formulations.
To address these limitations, we introduce \bench, a benchmark designed to evaluate GUI agents’ capabilities in performing continuous actions.

% \textbf{\textit{b}.} Many drag scenarios have multiple ways to complete the tasks, where a simple start-end coordinate pair cannot judge the task completion. Meanwhile existing benchmarks model drag such interactions as discrete start-end endpoint pair actions, and provide only a single screenshot with an atomic action, while the rich state changes during drag trajectories are discarded.
% \textbf{\textit{c}.} Many drag scenarios involve free-form trajectories, but these are not included nor evaluated in existing benchmarks. 

% To bridge this gap, we introduce \bench, a specialized benchmark explicitly designed to rigorously evaluate GUI agents\' capabilities in predicting and executing drag behaviors.

\subsection{Dataset Construction}
% \noindent \textbf{Task Source.} \bench comprises a diverse array of realistic tasks necessitating precise and continuous spatial control, including creative manipulation (PowerPoint object arrangement), dynamic file organization (File Manager), precise curve tracing (handwriting input), video editing (Adobe Premiere Pro clip adjustments), and captcha-solving tasks involving intricate spatial trajectories.
% % 

% \noindent \texpie chart heretbf{Data Curation.} We construct \bench by initially extracting structured metadata using UI Automation (UIA) to accurately localize interactive GUI elements. Leveraging these structured annotations, automated scripts systematically generate demonstrations of continuous trajectories. Each demonstration includes a high-frequency sequence of screenshots, finely sampled continuous spatial coordinates, and clearly defined conditions dictating autonomous action termination.

% \noindent \textbf{Observation and Action Space.} The observation space consists of sequential high-resolution screenshots ($1920\times1080$ pixels) captured at consistent intervals. The action space explicitly encodes continuous spatial trajectories as sequences of densely sampled 2D coordinates and mouse button state, enhancing the representation of nuanced interactions. Action sequences autonomously terminate when predefined spatial conditions are satisfied, mirroring conditions found in continuous-control robotic tasks.

% 不写UIA, cost efficient
\noindent\textbf{Data Source.}
\bench comprises a diverse set of real-world drag tasks including (1) PowerPoint element manipulation such as rotation, (2) File drag sorting in OS desktop and File manager, (3) Handwriting on canvas, (4) Premiere Pro asset manipulation between workbench and tracks, as well as (5) Captcha-solving.
These tasks cover both daily usage and professional control, requiring precise and continuous spatial control, and observation on-the-fly during the task execution.

% Why them? because professional control, daily usage,

% TODO: Insert visualization (e.g., pie chart) illustrating task diversity and distribution.

\noindent\textbf{Data Curation.}
% One navie way is reocrd human demonstration, but labor-costly.
% THus, we aim to find a scalable solution. but this require xxx
% (i) obj. coord
% (ii) instruct. sytheissi
% (iii) video record
Notably, high-quality drag data must capture screen state changes, dense cursor trajectories, and diverse task instructions. A straightforward approach is to manually record human demonstrations, but this is labor-intensive, thus not scalable. 
To overcome this limitation, we propose a scalable data collection pipeline that leverages automated drag execution, as illustrated in Fig.~\ref{fig:data}.
Our data curation approach includes three stages: \textbf{(i) Element Parsing:} We first extract the element bounding box metadata using UI Automation of Windows SDK~\cite{ufo}, to collect a set of candidate elements to be dragged,~\eg, an image in a PowerPoint slides. \textbf{(ii) Task Proposal:} Then we use an LLM (Qwen-2.5-72B~\cite{qwen2.5}) to propose candidate instructions,~\eg, reduce the image width by 0.4 from its right. 
\textbf{(iii) Trajectory Synthesis:} Next, we synthesize drag \texttt{PyAutoGUI} code containing dense trajectory coordinates and sending it to a code executor, which performs the drag actions on the real operating system and software applications. To complement this automated process, we also collect human demonstrations of drag interactions. Each human demonstration includes a high-resolution 60 FPS screen recording that captures UI state changes, the executed dense trajectory, and the corresponding instruction.

\begin{figure}[t]
    \centering
    \includegraphics[width=\linewidth]{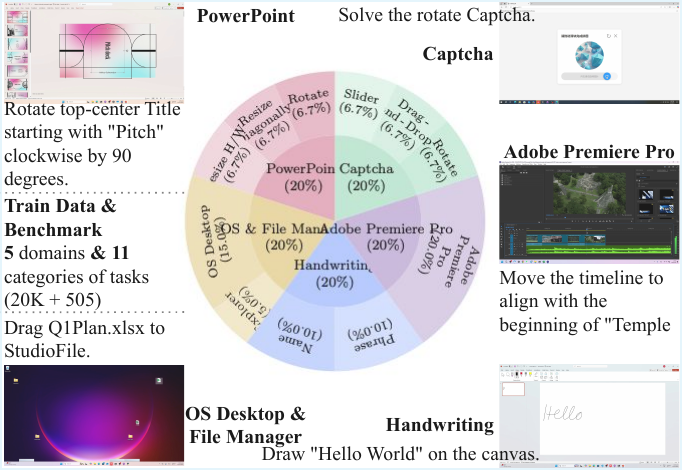}
    \vspace{-1em}
    \caption{\textbf{\bench Data Distribution.} The inner ring indicates the five equally distributed domains. The outer ring demonstrates per-category breakdowns with shares of the full dataset.}
    \label{fig:data_distribution}
    \vspace{-2em}
\end{figure}

\noindent\textbf{Data Analysis.}
As shown in Fig.~\ref{fig:data_distribution}, 
\bench comprises a training dataset of 20K manually collected and synthesized drag trajectories, with high resolution screen recording and metadata of the UI changes, as well as dense trajectories during the drag. The training dataset spans across five aforementioned domains and 11 categories of tasks. Moreover, \bench includes a benchmark of 505 trajectories for evaluation (101 tasks for each domain). The average duration of screen recordings is 9.62 seconds, and the average number of frames is 577 frames per recording. Refer to the Supplementary Material for more detailed analysis.
% , and proposes a data-driven close-loop approach to efficiently evaluate the drag task completion, to provide the model with UI changes responsive to their actions, without the requirement to setup complex OS or software environments, further enhancing reproducibility.

\subsection{Evaluation Metrics}
% challenging is still unclear hwo to define drag, one way is traj-alignment, other is outcome-aware.
Besides preparing training data, an effective evaluation protocol for model-generated drag trajectories is also essential. 
As noted earlier, the benchmark must expose UI state changes to the model throughout the drag process. Moreover, evaluation should go beyond assessing a simple start–end linear drag and instead offer a dynamic environment in which the model must perform drag actions incrementally.
To this end, \bench introduces two complementary evaluation modes for drag tasks: {offline evaluation} and {online evaluation}.
\noindent\textbf{Offline Evaluation} is performed as an open-loop~\cite{openloop_closeloop} assessment of the policy, measuring the policy's ability to replicate the ground-truth stepwise behavior without compounding errors. Given a static environment where screenshot, task and oracle previous state are provided, offline evaluation assesses whether the policy can successfully reach the next trajectory point, and the amount of the error, which can be represented by two metrics:

\noindent\textit{(i) Average Trajectory Error} evaluates the mean square error of the entire trajectory (aligned with Eq.\ref{eq:flow-matching-loss}), defined as the average Euclidean distance between predicted and ground-truth coordinates across all timesteps:
% TODO: add cap 
\begin{equation}
    % \text{ATE}\bigl(\hat{A}, A^{\text{gt}}\bigr) = \frac{1}{T}\sum_{t=0}^{T}\|\hat{a_{t}}-a_{t}^{gt}\|_{2},
    % \label{eq:ate}
    \text{ATE} = \frac{1}{T}\sum_{t=0}^{T} \bigl\| \hat{\mathbf{a}}_{t} - \mathbf{a}^{\mathrm{gt}}_{t} \bigr\|_{2},
    \label{eq:ate}
\end{equation}
providing a detailed assessment of the spatial accuracy and smoothness of interactions step by step throughout trajectory execution.

\noindent
\textit{(ii) Trajectory Endpoint Accuracy} emphasizes endpoint precision, measuring the proportion of predicted trajectories whose endpoints fall within a predefined spatial tolerance radius $\epsilon$,~\eg, 20 pixels from the ground-truth endpoints. This metric highlights the agent's ability to accurately reach the intended final interaction points. Defined as:
% \begin{equation}
% \text{TEA}=\mathbf{1}\left(a_\text{end}\in \mathcal{R}\right)
% \end{equation}
\begin{equation}
\text{TEA}
= \frac{1}{T}\sum_{t=0}^{T} 
\mathbf{1}\bigl(\hat{\mathbf{a}}_{t} \in \mathcal{R}_{t}\bigr),
\end{equation}
% offline: MSE, offset
% 1112: 分割线
where $\hat{\mathbf{a}}_t$ is the predicted intermediate action state, $\mathbf{a}_t^{gt}$ is the ground-truth intermediate action state, $\mathcal{R}_{t}$ is the corresponding acceptance region, $t$ is the timestamp, and $T$ is the number of timesteps. 
\begin{figure}[!t]
    \centering
    \includegraphics[width=\linewidth]{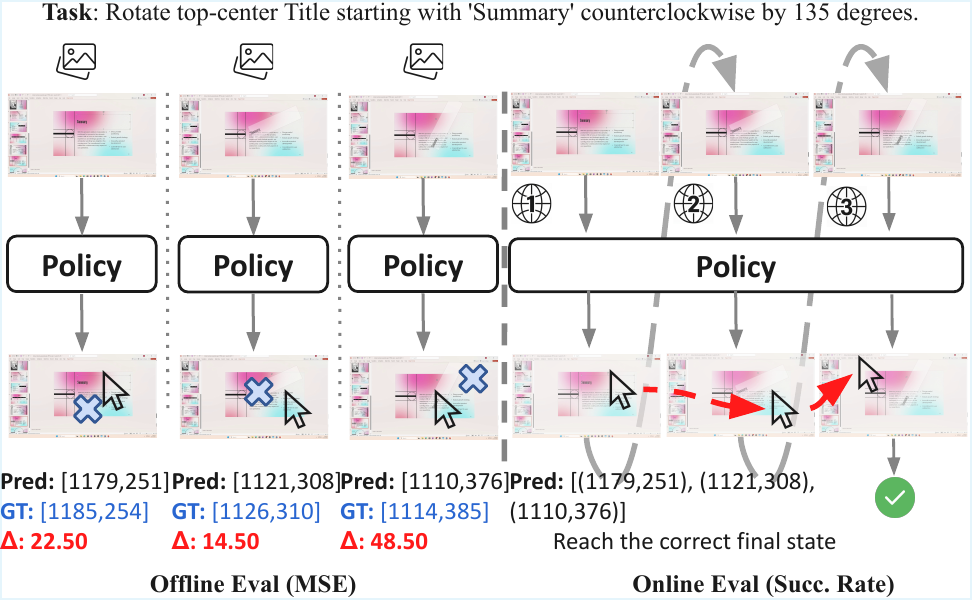}
    \vspace{-2em}
    \caption{\textbf{Comparison between offline evaluation and online evaluation pipelines of \bench:} \textbf{(i) Offline evaluation} is based on the distance of prediction and ground-truth in independent trunks; \textbf{(ii) Online evaluation} is incremental over sequential trunks, and based on the final outcome.}
    \label{fig:offline_online}
    \vspace{-2em}
\end{figure}

\begin{figure*}[!t]
    \centering
    \includegraphics[width=\textwidth]{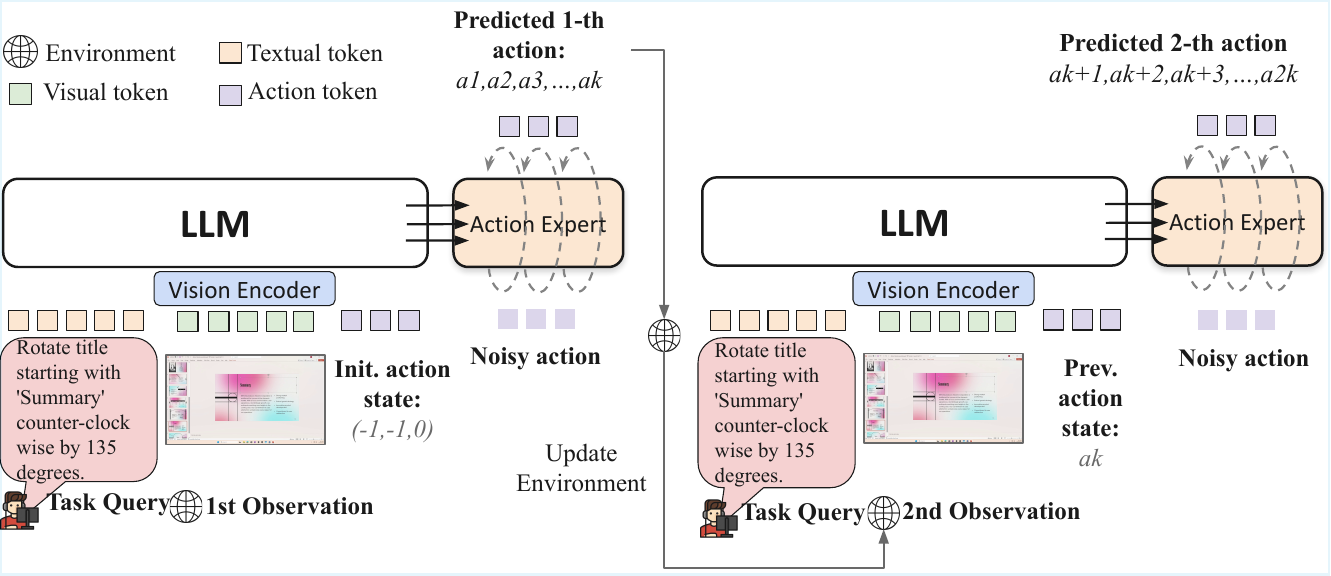}
    \vspace{-2em}
    \caption{\textbf{Overview of \our.}
    Given a task query and visual observations, the model first processes them through the VLM to obtain intermediate hidden states, which are then attended by the action expert. During interaction, the predicted actions update the environment, the next observation is encoded, and a new action chunk is produced—enabling fine-grained, closed-loop cursor control.
}
\label{fig:overview}
\vspace{-2em}
\end{figure*}

\noindent\textbf{Online Evaluation.}
Offline evaluation metrics are inherently limited as they solely measure static stepwise trajectory precision, neglecting dynamic closed-loop~\cite{openloop_closeloop} interactions and UI state changes in real-world GUI drag tasks, which are missing from offline evaluation, or naive endpoint-pair evaluation in existing benchmarks. 
% Therefore, we propose an efficient data-driven online evaluation method, without the complexity of setting up OS and software applications.
Notably, online evaluation naturally captures real-world trajectory variance, where agents can successfully achieve task goals despite diverse trajectory patterns.

% one sentence
However, constructing realistic online evaluation environments~\cite{offline_online} is non-trivial, such environments should provide diversity across multiple domains, while preserving the reproducibility. One naive approach is to set up the OS and software applications as environment, however, such designs either by manual setup or virtual machine image are costly and difficult to reproduce as the initial state of drag tasks are highly diverse. 
Therefore, we design a \textbf{data-driven} approach to enable policy closed-loop rollouts. For each drag task, we store the video recording, task specification, and dense drag trajectory, providing extensive possible GUI states encountered during dragging. During rollouts, the model’s predicted action is matched to the nearest recorded state if it falls within a tolerance $\epsilon$ (\eg~within 20 pixels of a ground-truth waypoint), upon which the corresponding next observation is retrieved. See the Supp. for more details.
\noindent\textit{(iii) Task Success Rate:}
In this data-driven online environment, the agent receives the current visual observation and task instruction, generates an action, and subsequently obtains an updated visual observation from the approximated corresponding state from the extensive states.
We measure performance primarily through binary task success (\emph{success or failure}) per rollout, computed as the number of successful tasks against the total number of tasks. 

\begin{equation}
\text{Success Rate}
= \frac{1}{N}\sum_{i=1}^{N}
\mathbf{1}\bigl(\hat{\mathbf{a}}_{i, T_i} \in \mathcal{R}^{\text{goal}}_{i}\bigr),
\label{eq:tsr}
\end{equation}
where $N$ is the number of online rollouts (tasks), $\hat{\mathbf{a}}_{i, T_i}$ is the final action state of the $i$-th rollout, and $\mathcal{R}^{\text{goal}}_{i}$ is the goal acceptance region for that task (\eg~all end states within $\epsilon$ pixels of the ground-truth goal). See Supp. for illustrations.

\begin{figure}[htbp]
    \centering
    \includegraphics[width=\linewidth]{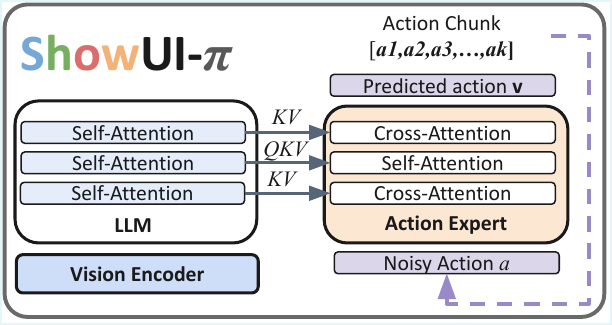}
    \vspace{-2em}
    \caption{\textbf{Architecture of \our.}
    \our uses an LLM with cross-attention to a lightweight action expert to generate unified action chunks that handle both discrete clicks and continuous drag segments.
}
\label{fig:arch}
\vspace{-1em}
\end{figure}
\section{\SHOW Model}

\noindent\textbf{Overview.}
% compress first 2 into 1
As outlined in Fig.~\ref{fig:arch}, \our~is built upon the SmolVLA-450M~\cite{smolvla} architecture, with tailored designs for streaming GUI control. \our consists of two primary integrated components: 
(\textit{i}) a pretrained VLM, initialized by SmolVLM-2~\citep{smolvlm} and (\textit{ii}) a flow matching action expert. 
The VLM efficiently processes multimodal inputs, encoding screenshot $o$ into visual token representations in a unified embedding space together with projected action states and language instructions $\mathcal{Q}$. 
The action expert is built on a transformer with the same number of layers as the VLM backbone (\ie~16). 
During action prediction, the corresponding layers of the VLM backbone and the action expert perform interleaved self-attention and cross-attention. The \textit{action state} $a$ from the previous step is projected back into the VLM backbone to condition subsequent predictions; it is initialized as $a_0=[-1,-1]$ and later updated using the last action $a_k$. The action expert is trained with flow matching, executed $k$ times per trajectory, where each step refines noisy actions into clean predictions, enabling smooth and deterministic trajectory generation.
We next illustrate our three designs for effectively modeling:
(\textit{i}) Unified action representation;
{(\textit{ii})} Flow-based trajectory generation;
{(\textit{iii})} Directional regularization for stable trajectory prediction.
% \textbf{(i) Flow-based Trajectory Generation with Temporal Weighting}, incrementally integrating a conditional vector field to produce horizon-free trajectories, naturally supporting short clicks and extended drags without explicit duration thresholds;
% \textbf{(ii) Unified Action Representation}, casting discrete clicks and continuous drags into a unified sequence format $(x,y,m)$;
% \textbf{(iii) Vision-Language-Action Streaming Policy}, interleaving continuous visual inputs with action trajectories to achieve low-latency, closed-loop responses in dynamically evolving UI environments.

\subsection{Unified Continuous and Discrete Actions}
\label{sec:unify}

Clicks and drags exhibit different temporal and spatial dynamics, making it nontrivial to integrate them within a unified model. Yet such unification is crucial: a single model capable of handling both interaction types can flexibly adapt to diverse GUI tasks that require switching between discrete and continuous actions.
Building upon our trajectory generation, we aim to unify discrete clicks and continuous drags under a unified action modeling framework. 
% Observing that \textit{a click is essentially a drag with negligible movement}, we model all GUI interactions as sequences of coordinate with atomic up and down.
Observing that “\textit{a click is essentially a drag with negligible movement}”, we thus unify click or drag action interactions as sequences $\mathcal{A}$ of $(x,y,m)$ triplets, where $(x,y)$ are cursor coordinates and $m$ is an atomic up and down mouse button state.
\[
\mathcal{A} = [\mathbf{a}_1,\dots,\mathbf{a}_H],
\mathbf{a}_k = (x_k,y_k,m_k),
m_k\in\{\texttt{down},\texttt{up}\}.
\]
\textbf{(i)} Within this representation, clicks become minimal two-step trajectories:
$[(x1,y1,\texttt{down}),\,(x1,y1,\texttt{up})]$ while \textbf{(ii)} Drags, become naturally represented as extended incremental press-hold trajectories:
\[
[(x_{1},y_{1},\texttt{down}),\,(x_{2},y_{2},\texttt{down}),\,\dots,\,(x_{T},y_{T},\texttt{up})].
\]

This unified representation greatly simplifies the action space, removing the need for rigid, predefined action formats used in prior GUI agents (\eg language tokens). By modeling all interactions as continuous sequences, the policy naturally supports flexible multi-dataset co-training, seamlessly integrating both clicking~\cite{guicourse,showui} and dragging within a single unified framework.

\subsection{Flow-based Continuous Trajectory}
To achieve real-time interactions—which demand smooth and efficient trajectory generation—we adopt a flow-based incremental generation framework~\cite{flow_matching} driven by a lightweight conditional vector field $\mathbf{v}_\theta$:

\vspace{-0.5em}

\begin{equation}
    \frac{d\hat{\mathbf{a}}(s)}{ds} = \mathbf{v}_\theta\!\big(\hat{\mathbf{a}}(s), s \mid o_t,\mathcal{Q}\big),
    % , \quad \hat{\mathbf{a}}([-1,1,0])=\mathbf{a}_1,
    \label{eq:vector_field}
\end{equation}
where $o_t$ is the current visual observation, $\mathcal{Q}$ is the task instruction, and $\hat{\mathbf{a}}(s)=[x,y,m]$ represents the predicted action trajectory as in $\S$~\ref{sec:unify}. Moreover, $s\in[0,1]$ is a continuous parameter of the flow, smoothly moving the cursor from the segment's start where $s=0$ to its end where $s=1$.

Instead of treating all trajectory steps equally as in the naive flow-matching loss (Eq.~\ref{eq:flow-matching-loss}), we emphasize the steps that matter most for successful GUI interaction. In particular, early movements must be conditioned on the start point, and the final steps must land precisely on the intended endpoint. To capture this, we introduce a \textit{reweighting} scheme that increases the contribution of both the initial and terminal segments of the trajectory:

\begin{equation}
    \mathcal{L}_{\text{weighted}} = \sum_{t=1}^{T} w_t \, \mathcal{L}_{\text{flow matching}}^{(t)},
    \label{eq:weighted_flow_matching_loss}
\end{equation}
where $w_t$ is a constant that assigns larger weights to critical steps. We adopt a simple scheme that assigns a weight of 10 to the start and end points, otherwise 1. 
% An ablation of the choice of $w_t$ is reported in Tab.~\ref{tab:ablation_chunk_actions}.
% This reweighting scheme on top of the flow~\ref{eq:vector_field} encourages the model to prioritize correct initial movements and precise task completion, and preserve the stability of flow matching, aligning closely with goal-oriented GUI drag tasks.

% ======================= NEW: MAIN RESULTS (ACCURACY + OFFLINE ERROR) =======================
\begin{table*}[!h]
\centering
\setlength{\tabcolsep}{6pt}
\renewcommand{\arraystretch}{1.18}
\arrayrulecolor{black}
\resizebox{1.0\linewidth}{!}{%
\begin{tabular}{lcccccc|cccccc}
\toprule
\multirow{2}{*}{\textbf{Methods}} & \multicolumn{6}{c|}{\textbf{Offline Endpoint Accuracy (\%, ↑)}} & \multicolumn{6}{c}{\textbf{Offline Trajectory Error (↓)}} \\
\rule{0pt}{2.2ex} & OS & PPT & Premiere & Captcha & Handwriting & Overall & OS & PPT & Premiere & Captcha & Handwriting & Overall \\
\midrule
\multicolumn{13}{l}{\textbf{Action as Language} — \textit{Proprietary models}}\\[-0.3ex]
\rowcolor{BaselineGray}
Operator        & 51.49 & 3.96 & 0.00 & 0.00 & 0.00 & 11.09 & 46.85 & 244.25 & 602.00 & 0.00 & 795.58 & 422.17 \\
\rowcolor{BaselineGray}
UI\mbox{-}TARS\mbox{-}1.5\mbox{-}7B  & 60.40 & 0.99 & 0.00 & 3.96 & 0.00 & 13.07 & 224.57 & 235.50 & 564.89 & 172.53 & 703.54 & 380.21 \\
\rowcolor{BaselineGray}
Seed\mbox{-}1.6\mbox{-}Vision & 71.29 & 1.98 & 0.99 & 4.95 & 0.00 & 15.84 & 62.17 & 233.92 & 467.09 & 81.64 & 824.55 & 333.87 \\
\rowcolor{BaselineGray}
Gemini\mbox{-}2.5\mbox{-}CUA & 84.16 & 11.88 & 0.00 & 3.96 & 0.00 & 20.00 & 12.11 & 121.94 & 508.19 & 114.34 & 0.00 & 189.15 \\
\addlinespace[1pt]
\multicolumn{13}{l}{\textbf{Action as Language} — \textit{Open-source models}}\\[-0.3ex]
\rowcolor{BaselineGray}
OpenCUA\mbox{-}32B     & 96.04 & 6.93 & 0.00 & 0.00 & 0.00 & 20.59 & 10.95 & 165.70 & 907.41 & 62.06 & 791.88 & 387.60 \\
\rowcolor{BaselineGray}
OpenCUA\mbox{-}7B      & 99.01 & 4.95 & 0.00 & 3.96 & 0.00 & 21.58 & 4.42 & 186.65 & 762.77 & 214.81 & 959.10 & 425.55 \\
\rowcolor{BaselineGray}
Qwen3\mbox{-}VL\mbox{-}32B     & 71.29 & 1.98 & 0.99 & 0.00 & 0.00 & 14.85 & 149.77 & 212.02 & 527.23 & 236.83 & 646.73 & 354.52 \\
\rowcolor{BaselineGray}
Qwen3\mbox{-}VL\mbox{-}8B      & 8.91 & 1.98 & 0.00 & 0.00 & 0.00 & 2.18 & 660.33 & 228.47 & 532.71 & 257.44 & 662.37 & 468.26 \\
\rowcolor{BaselineGray}
Qwen3\mbox{-}VL\mbox{-}2B      & 3.96 & 0.00 & 0.99 & 0.00 & 0.00 & 0.99 & 708.68 & 248.29 & 551.15 & 258.07 & 562.62 & 465.76 \\
\multicolumn{13}{l}{\textbf{Flow-based action (Ours)}}\\
\rowcolor{BandBlue}
\textcolor{OursBlue}{\textbf{ShowUI-\ensuremath{\pi}-450M}} &
\best{61.39} & \best{85.06} & \best{56.92} & \best{96.30} & \best{93.07} & \best{78.55} &
\best{350.55} & \best{57.68} & \best{195.96} & \best{136.47} & \best{54.57} & \best{159.05} \\
\bottomrule
\end{tabular}%
}
\arrayrulecolor{black}
\vspace{-1em}
\caption{\textbf{Main Results on \bench\ offline evaluation.} Left: Endpoint Accuracy; Right: Trajectory Error. 
}
% \vspace{-1em}
\label{tab:main_results_offline} % keep original label to avoid breaking refs
\end{table*}

\subsection{Directional Regularization}
% Unlike robotics manipulation trajectories which are often naturally jittery, 
GUI actions inherently demand smooth, directionally coherent cursor movements.
Standard flow-based methods tend to optimize trajectory magnitude without explicitly enforcing directional consistency. This directional misalignment can cause critical failures in GUI tasks, such as incorrect cursor orientation or jitter.
To explicitly regularize directional alignment, we introduce a \textbf{directional regularization} loss term, defined as 
\begin{equation}
\mathcal{L}_{\text{reg}} = \frac{1}{T}\sum_{t=1}^{T} (1 - \cos(\hat{\mathbf{a}}_t, \mathbf{u}_t)),
\end{equation}
where $\hat{\mathbf{a}}_t$ and $\mathbf{u}_t$ represent predicted and ground-truth point.

% By combining this directional term with the weighted flow-matching objective, we significantly improve angular accuracy, yielding trajectories that are spatially precise and directionally consistent, thus ensuring robust GUI task execution.

% loss=L1+L2;
\noindent\textbf{Total Objective.} By combining the reweighting flow-matching objective with directional regularization, the final training objective is defined as:
\begin{equation}
\mathcal{L}_{\text{total}} = \mathcal{L}_{\text{weighted}} + \lambda\,\mathcal{L}_{\text{reg}},
\label{eq:final_loss}
\end{equation}
where $\lambda$ controls the balance between trajectory accuracy and directional consistency. We set $\lambda$ to 0.1 to ensure comparable magnitudes of loss terms.

% ======================= NEW: MAIN RESULTS (TRAJECTORY ERROR, SMALLER, SINGLE COLUMN) =======================
\begin{table}[!h]
\centering
\setlength{\tabcolsep}{3.5pt}
\renewcommand{\arraystretch}{1.05}
\arrayrulecolor{black}
\resizebox{\linewidth}{!}{%
\begin{tabular}{lcccccc}
\toprule
\multirow{2}{*}{\textbf{Methods}} & \multicolumn{6}{c}{\textbf{Online Success Rate (\%, ↑)}} \\
\rule{0pt}{2.2ex} & OS & PPT & Premiere & Captcha & Handwriting & Overall \\
\midrule
\multicolumn{7}{l}{\textbf{Action as Language} — \textit{Proprietary models}}\\[-0.3ex]
\rowcolor{BaselineGray}
Operator        & 53.47 & 9.90 & 2.97 & 0.00 & 0.00 & 13.27 \\
\rowcolor{BaselineGray}
UI\mbox{-}TARS\mbox{-}1.5\mbox{-}7B  & 73.27 & 1.98 & 1.98 & 7.92 & 0.00 & 17.03 \\
\rowcolor{BaselineGray}
Seed\mbox{-}1.6\mbox{-}Vision & 77.23 & 3.96 & 1.98 & 8.91 & 2.97 & 19.01 \\
\rowcolor{BaselineGray}
Gemini\mbox{-}2.5\mbox{-}CUA & 86.14 & 20.79 & 0.00 & 3.96 & 0.00 & 22.18 \\
\addlinespace[1pt]
\multicolumn{7}{l}{\textbf{Action as Language} — \textit{Open-source models}}\\[-0.3ex]
\rowcolor{BaselineGray}
OpenCUA\mbox{-}32B     & 97.03 & 6.93 & 0.00 & 0.00 & 0.00 & 20.79 \\
\rowcolor{BaselineGray}
OpenCUA\mbox{-}7B      & 99.01 & 4.95 & 0.00 & 5.94 & 0.00 & 21.98 \\
\rowcolor{BaselineGray}
Qwen3\mbox{-}VL\mbox{-}32B     & 83.17 & 4.95 & 3.96 & 2.97 & 0.00 & 19.01 \\
\rowcolor{BaselineGray}
Qwen3\mbox{-}VL\mbox{-}8B      & 24.75 & 5.94 & 6.93 & 0.00 & 0.00 & 7.52 \\
\rowcolor{BaselineGray}
Qwen3\mbox{-}VL\mbox{-}2B      & 13.86 & 6.93 & 2.97 & 0.00 & 0.00 & 4.75 \\
\multicolumn{7}{l}{\textbf{Flow-based action (ours)}}\\
\rowcolor{BandBlue}
\textcolor{OursBlue}{\textbf{ShowUI-\ensuremath{\pi}-450M}} &
\best{13.11} & \best{22.93} & \best{8.64} & \best{55.91} & \best{34.32} & \best{26.98} \\
\bottomrule
\end{tabular}%
}
\arrayrulecolor{black}
\vspace{-1em}
\caption{\textbf{Main Results on \bench\ (Online Success Rate)}.}
% \vspace{-2em}
\label{tab:main_results_online} % reusing the existing label to avoid breaking refs in text
\end{table}
\section{Experiments}
Our experiments systematically investigate the following research questions:
\textbf{Q1:} How does \our compare to existing mainstream GUI agents on performing dexterous operations?
\textbf{Q2:} What factors contribute to the performance of \our?
\textbf{Q3:} How well does \our generalize to different task domains under multi-dataset training?
% \subsection{Setup}

% \noindent\textbf{Benchmarks and Metrics.}
% We evaluate \our on both drag and click tasks to showcase it performance as a flow-based VLA with unified action representation.

% owning to our unificaition of click and drag, we are able Multi-task training

% We evaluate \our's drag capacity comprehensively across five task categories on \bench: PowerPoint, File drag sorting in OS and file manager, Premiere Pro, Captcha, and Handwriting. Metrics include open-loop offline Endpoint Accuracy and Trajectory Error (Eq.~\ref{eq:ate}), alongside online close-loop binary success rate.

% \noindent\textbf{Click.} we evaluate \our's click and grounding performance on Element Grounding UI-Vision~\cite{uivision}, a comprehensive computer use grounding benchmark across 83 software applications of six categories with diverse platforms.

\noindent\textbf{Baselines.}
We compare \our with several representative baselines, including both proprietary models and open-source models, including OpenAI Operator~\cite{openai_operator}, Gemini-2.5-CUA~\cite{gemini2.5cua}, Seed 1.6-Vision~\cite{seed1.6vl}, Qwen3-VL~\cite{qwen3vl}, UI-TARS 1.5~\cite{uitars1.5}, OpenCUA~\cite{opencua}, \etc. 
% Additionally, we also perform extensive ablation studies to provide more insights. 
% As the baseline models output actions as language, when performing offline evaluation on baseline models, we use their first action output for evaluation. When performing online evaluation on baseline models, they can iteratively interact with the data-driven close-loop online environment for multiple steps, and a rollout is counted as successful if the baseline models reach the goal state in any step. Given the size and computational cost of these large models, we cap the interaction horizon at three steps.
Because baseline models output actions as language, we use only their first predicted action for offline evaluation. 
For online evaluation, they interact with our data-driven closed-loop environment for fair comparison, and a rollout is marked successful if the goal is reached at any step. To control computational cost, we limit each baseline to three interaction steps.

\subsection{Main Results}

% As shown in Tab.~\ref{tab:main_results} and Tab.~\ref{tab:main_results_traj}, \our achieves the highest overall performance in \bench with only 450M parameters, surpassing the performance of various larger baseline models. 

\noindent\textbf{Online Evaluation.} In Tab.~\ref{tab:main_results_online}, \our attains an overall online closed-loop success rate of 26.98\%, surpassing the SOTA proprietary model Gemini-2.5-CUA by 4.8\%, and the SOTA open-source model OpenCUA-7B by 6.19\%.
It is worth noting that \our performs well on tasks involving free-form drag actions, \eg, PowerPoint where circular drag is required to rotate elements, and Handwriting where trajectories are non-linear. Moreover, \our performs well on Captcha tasks where observation on-the-fly is required. 
Interestingly, Operator failed all Captcha tasks as it refused to solve Captcha due to its safety policy, and Gemini-2.5-CUA always mistakenly calls open-browser tool when executing handwriting tasks. Furthermore, OpenCUA-7B reaches the second highest performance among all baseline models, demonstrating that increasing parameter size cannot always improve the action performance, \ie, drags and clicks, of computer use agents, and lightweight models such as OpenCUA-7B and \our are worth further development.

% ======================= NEW: Ablation on Chunk Size and Execution Steps (updated layout) =======================
\begin{table}[htbp]
\centering
\small
\setlength{\tabcolsep}{5pt}
\renewcommand{\arraystretch}{0.9} % 稍微压缩整体高度
\arrayrulecolor{black}
\resizebox{1\linewidth}{!}{%
\begin{tabular}{l c cc}
\toprule
% \rowcolor{white}
\multicolumn{1}{c}{\textbf{\shortstack{Chunk\\ Size}}} &
\multicolumn{1}{c}{\textbf{\shortstack{Exec.\\ Steps}}} &
\multicolumn{2}{c}{\textbf{\shortstack{Offline eval.}}} \\
\rowcolor{white}
% 这里用 \raisebox 把 length(a) 往上抬，让它在这一行里看起来居中
\multicolumn{1}{c}{\textcolor{gray!100}{\footnotesize \raisebox{0.3ex}{length($\hat{\mathbf{a}}$)}}} &
\multicolumn{1}{c}{\textcolor{gray!100}{\footnotesize \shortstack{before next \\ observation}}} &
{Endpoint Acc. (↑)} & {Traj. Error (↓)} \\[-0.3em]

\midrule
\multirow{3}{*}{10}  & 1   & 75.45 & 176.43 \\
                     & 2   & 72.28 & 193.81 \\
                     & 5   & 66.34 & 241.60 \\
\cmidrule(lr){1-4}
\multirow{3}{*}{20}  & 1   & 78.55 & \best{159.05} \\
                     & 2   & 76.35 & 177.87 \\
                     & 5   & \best{79.22} & 205.23 \\
\bottomrule
\end{tabular}%
}
\arrayrulecolor{black}
\vspace{-1em}
\caption{\textbf{Ablation on Chunk Size and Execution Steps.} Two chunk sizes ($n\in\{10,20\}$); each evaluated at execution steps $\{1,2,5\}$. We report offline Endpoint Acc. and Traj. Error.}
% \vspace{-2em}
\label{tab:ablation_chunk_actions}
\end{table}

% =========== Temporal Weights Ablation (Online) ===========
\begin{table}[htbp]
\centering
\setlength{\tabcolsep}{5pt}
\renewcommand{\arraystretch}{1.1}
\arrayrulecolor{black}
\resizebox{\linewidth}{!}{%
\begin{tabular}{lcccccc}
\toprule
\rowcolor{white}
\multirow{2}{*}{\centering \textbf{\shortstack{$w$}}} &
\multicolumn{6}{c}{\textbf{Online Success Rate (\%, ↑)}} \\
& OS & PPT & Premiere & Captcha & Handwriting & Overall \\
\midrule
{1} & 12.87 & 11.49 & 10.77 & 7.41 & 9.90 & 10.49 \\
\rowcolor{BandBlue}
\textcolor{OursBlue}{5} & 11.69 & 17.30 & 7.84 & 19.40 & 16.22 & 14.49 \\
\rowcolor{BandBlue}
\textcolor{OursBlue}{10} &
\best{13.11} & \best{22.93} & \best{8.64} & \best{55.91} & \best{34.32} & \best{26.98} \\
\rowcolor{BandBlue}
\textcolor{OursBlue}{15} &
26.73 & 11.49 & 7.69 & 33.33 & 24.75 & 20.80 \\
\bottomrule
\end{tabular}%
}
\arrayrulecolor{black}
\vspace{-1em}
\caption{\textbf{Effects of Temporal Weights.} Removing temporal weights reduces online success across domains.}
% \vspace{-2em}
\label{tab:ablation_temporal}
\end{table}

% ===================== Unified vs Separate Action Heads =====================
\begin{table}[htbp]
\centering
\scriptsize                    % 比 \footnotesize 更小
\setlength{\tabcolsep}{3pt}    % 列间距再缩一点
\renewcommand{\arraystretch}{0.85} % 压低行距
\arrayrulecolor{black}

% 调细 booktabs 横线
\setlength{\heavyrulewidth}{0.6pt}
\setlength{\lightrulewidth}{0.4pt}
\setlength{\cmidrulewidth}{0.4pt}

\begin{tabular*}{\columnwidth}{@{\extracolsep{\fill}}lccc@{}}
\toprule
\multirow{2}{*}{\textbf{Method}} & \multirow{2}{*}{\#Params} & \multicolumn{2}{c}{\textbf{Drag}} \\
\rule{0pt}{1.4ex} & & {Online SR.} & {Offline Acc.} \\
\midrule
Separate Heads & 550M & 23.25 & 79.22 \\

% \textbf{\textcolor{OursBlue}{Unified Head}} & 450M & \best{26.98} & \best{78.55} \\
\textbf{Unified Head} & 450M & \best{26.98} & \best{78.55} \\

\bottomrule
\end{tabular*}

\caption{\textbf{Unified vs.\ Separate Action Heads.} Unified head provides comparable performance to separate heads, while being more elegant and practical.}
\label{tab:ablation_action_head}
\end{table}

\begin{table}[htbp]
\centering
\setlength{\tabcolsep}{5pt}
\renewcommand{\arraystretch}{1.1}
\arrayrulecolor{black}
\resizebox{\linewidth}{!}{%
\begin{tabular}{lcccccc}
\toprule
\rowcolor{white}
\multirow{2}{*}{\centering \textbf{\shortstack{Directional \\ Reg.}}} &
\multicolumn{6}{c}{\textbf{Online Success Rate (\%, ↑)}} \\
& OS & PPT & Premiere & Captcha & Handwriting & Overall \\
\midrule
{w/o. ($\lambda=0$)} & 10.83 & 14.37 & 8.26 & 14.92 & 14.78 & 12.63 \\
\rowcolor{BandBlue}
\textcolor{OursBlue}{{w.} ($\lambda=0.1$)} &
\best{13.11} & \best{22.93} & \best{8.64} & \best{55.91} & \best{34.32} & \best{26.98} \\
\bottomrule
\end{tabular}%
}
\arrayrulecolor{black}
\caption{\textbf{Effects of Directional Regularization.} Directional regularization improves online success across all domains, especially in domains that are sensitive to drag direction,~\eg, Captcha.}
\label{tab:ablation_directional}
\end{table}

% ======================= NEW: Flow Matching vs Diffusion vs Pure LM =======================
\begin{table*}[htbp]
\centering
\small
\setlength{\tabcolsep}{6pt}
\renewcommand{\arraystretch}{1.18}
\arrayrulecolor{black}
\begin{tabular}{lcccccc}
\toprule
\rowcolor{white}
\multirow{2}{*}{\textbf{Trajectory modeling}} &
\multicolumn{6}{c}{\textbf{Offline Evaluation.} Avg. Endpoint Accuracy ($\uparrow$) / Trajectory Error ($\downarrow$).} \\
% \rowcolor{white}
\rule{0pt}{2.3ex} &
OS & PowerPoint & Premiere & Captcha & Handwriting & Overall \\
\midrule
% \multicolumn{7}{l}{\textbf{Action as Language }}\\
\rowcolor{BaselineGray}
SmolVLM  & 1.98 / 566.98 & 0.00 / 305.64 & 0.00 / 493.34 & 0.00 / 254.14 & 0.00 / 440.42 & 0.40 / 412.10 \\
% \multicolumn{7}{l}{\textbf{Flow-based policy}}\\
\rowcolor{BaselineGray}
Diffusion Policy & 43.47 / 553.66 & 45.17 / 154.11 & 29.70 / 347.88 & 64.44 / 203.46 & 53.56 / 80.49 & 47.33 / 267.92 \\
% \multicolumn{7}{l}{\textbf{Flow-based policy (ours)}}\\
\rowcolor{BandBlue}
\textcolor{OursBlue}{\textbf{Flow Matching (Our)}} &
\best{61.39} / \best{350.55} & \best{85.06} / \best{57.68} & \best{56.92} / \best{195.96} &
\best{96.30} / \best{136.47} & \best{93.07} / \best{54.57} & \best{78.55} / \best{159.05} \\
\bottomrule
\end{tabular}
\arrayrulecolor{black}
\vspace{-1em}
\caption{\textbf{Ablation of different modeling approaches under the same SmolVLM backbone.} We compare language modeling, diffusion policy, and flow matching, all trained on the same 20K training trajectories. Our flow-matching delivers the most effective performance.}
% \vspace{-2em}
\label{tab:ablation_flow_diffusion}
\end{table*}

\noindent\textbf{Offline Evaluation.} As shown in Tab.~\ref{tab:main_results_offline}, endpoint accuracy and trajectory error are computed in offline evaluation to evaluate whether the trajectory closely aligns with the ground-truth trajectory. For baseline models which output actions as language, only the endpoints are used to compute the trajectory error. For \our, all waypoints are used to evaluate whether the model learned to closely follow the trajectory. It can be observed from Tab.~\ref{tab:main_results_offline} that \our produces the least trajectory error, showing that the flow-based VLA can follow the learned trajectory pattern well.

\noindent\textbf{Which action modeling approach is most effective?}
Flow matching delivers the highest endpoint accuracy of 78.55\% and lowest trajectory error of 159.05\,px, surpassing diffusion policy by 31.22\% and lower error, and language modeling by 78.15\% and much lower error as shown in Tab.~\ref{tab:ablation_flow_diffusion}. This demonstrates flow matching's deterministic velocity field effectively captures complex, free-form GUI drags with its training stability.

\subsection{Key Ablations}
\noindent\textbf{How do chunk size and execution steps impact model performance?}
As shown in Tab.~\ref{tab:ablation_chunk_actions}, increasing chunk size,~\ie, length of steps in each prediction, from 10 to 20 generally boosts the performance, indicating predicting more action steps is helpful for learning to capture the action distribution. Interestingly, when the execution steps are large,~\ie, executing more steps from prediction before the next re-observation, increasing chunk size can help alleviate the degradation. When the execution steps are 5, increasing chunk size from 10 to 20 can improve its accuracy by 12.9\% and reduce trajectory error by 36.4\,px. Moreover, fewer execution steps generally produce the highest accuracy and least error, showing that frequent re-observation enhances precision. Therefore, chunk size 20 with execution step 1 achieves the tradeoff for reliable and precise actions.

\noindent\textbf{How does applying reweighting affect model performance?}
As shown in Eq.~\ref{eq:weighted_flow_matching_loss}, reweighting applies step-dependent coefficients to the flow matching loss across the predicted action horizon, prioritizing accuracy at important steps,~\ie, after drag starts and near drag ends. Reweighting at scale 10 yields the highest overall success rate, increasing from 10.49\% without weighting to 26.98\%, with dramatic gains on Captcha by 48.50\%, showcasing the importance of starting and ending steps for successful drags. Interestingly, overweighting,~\ie, with scale 15 benefits OS File drag but reduces overall performance, showing that overweighting may harm learning of intermediate actions, and moderate weighting optimally learns the entire drag, shown in Tab.~\ref{tab:ablation_temporal}.

\noindent\textbf{Unified action head or separate heads for drags and clicks?}
As shown in Tab.~\ref{tab:ablation_action_head}, the unified head matches separate heads in offline accuracy at 78.55\% versus 79.22\%, while improving online drag success by 3.7\% and reducing model size by 100M parameters. Though the two designs have comparable performance, the unified design is more elegant, and allows \our to handle both drag and click tasks without task-dependent head selection. Conversely, the separate head design introduces impractical head selection and requires an additional 100M parameters, resulting in higher computational overhead.

\noindent\textbf{Is Directional Regularization helpful?}
It can be observed from Tab.~\ref{tab:ablation_directional} shows that directional regularization consistently improves \our across all domains, as GUI drag tasks demand smooth and directionally correct trajectories,~\eg, Captcha-solving may fail to place the slider or rotate to the wrong angle if the drag direction is inaccurate.

% REQUIRE: \usepackage{booktabs} \usepackage{graphicx} \usepackage{array}

\begin{table}[t]
\centering
\setlength{\tabcolsep}{5pt}
\renewcommand{\arraystretch}{1.15}

% 列类型：左/中，固定宽度并支持自动换行
\newcolumntype{L}[1]{>{\raggedright\arraybackslash}m{#1}}
\newcolumntype{C}[1]{>{\centering\arraybackslash}m{#1}}

\def\imgH{1.05cm} % 缩略图高度（调这个就行）

\resizebox{\columnwidth}{!}{%
\begin{tabular}{L{0.2\linewidth} L{0.4\linewidth} C{0.4\linewidth}}
\toprule
\textbf{Domain} & \textbf{Task Instruction} & \textbf{Trajectory Visualization} \\
\midrule
PowerPoint & Rotate center Fox clockwise by 45 degrees. &
\includegraphics[width=\linewidth]{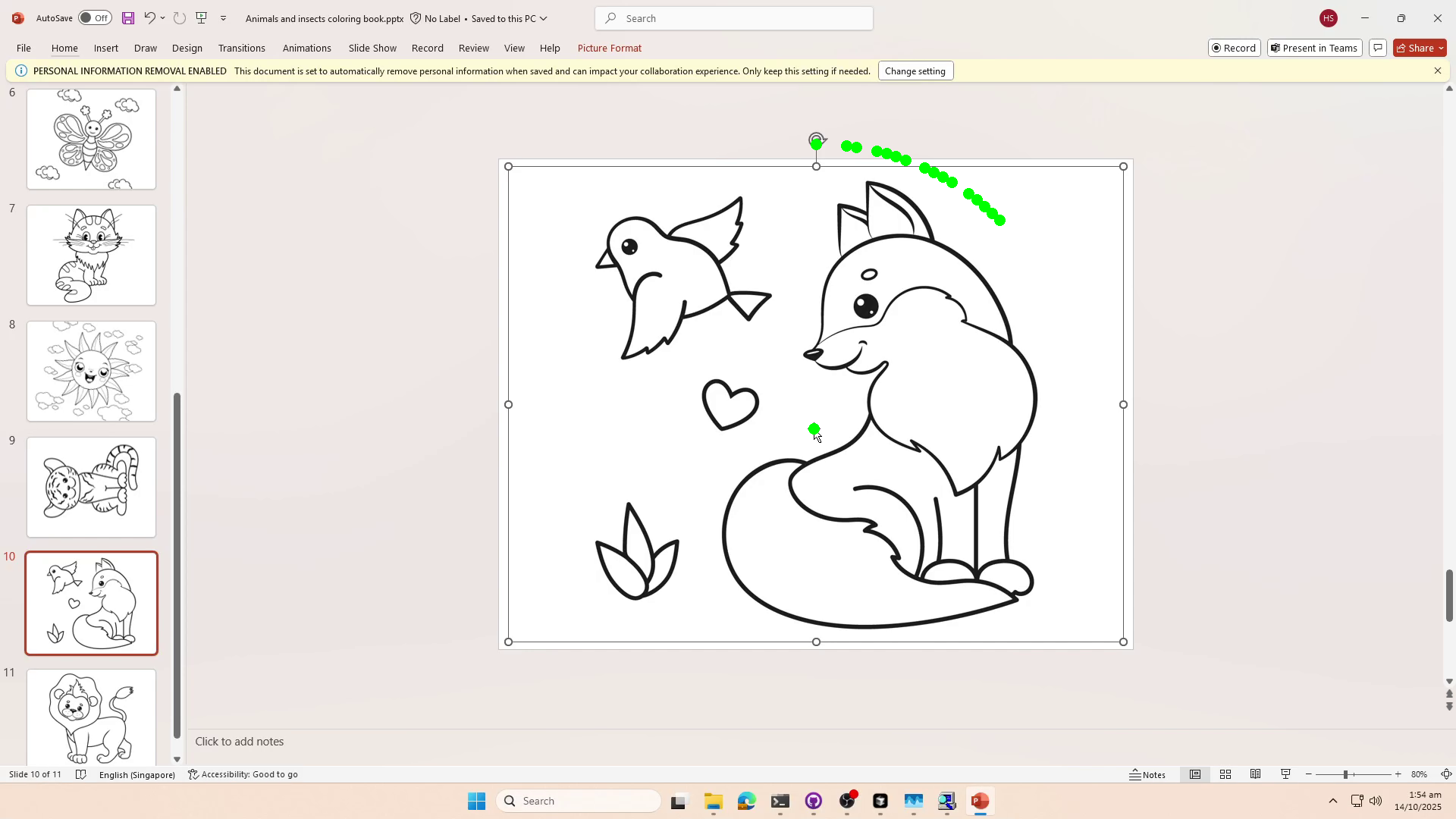} \\
\midrule
% OS Desktop & Drag Q2Draft.docx to ClientHub &
% \includegraphics[width=\linewidth]{chart/visualization_assets/desktop_trajectory.png} \\
% \midrule
Handwriting & Write “Starlit grin” on the canvas &
\includegraphics[width=\linewidth]{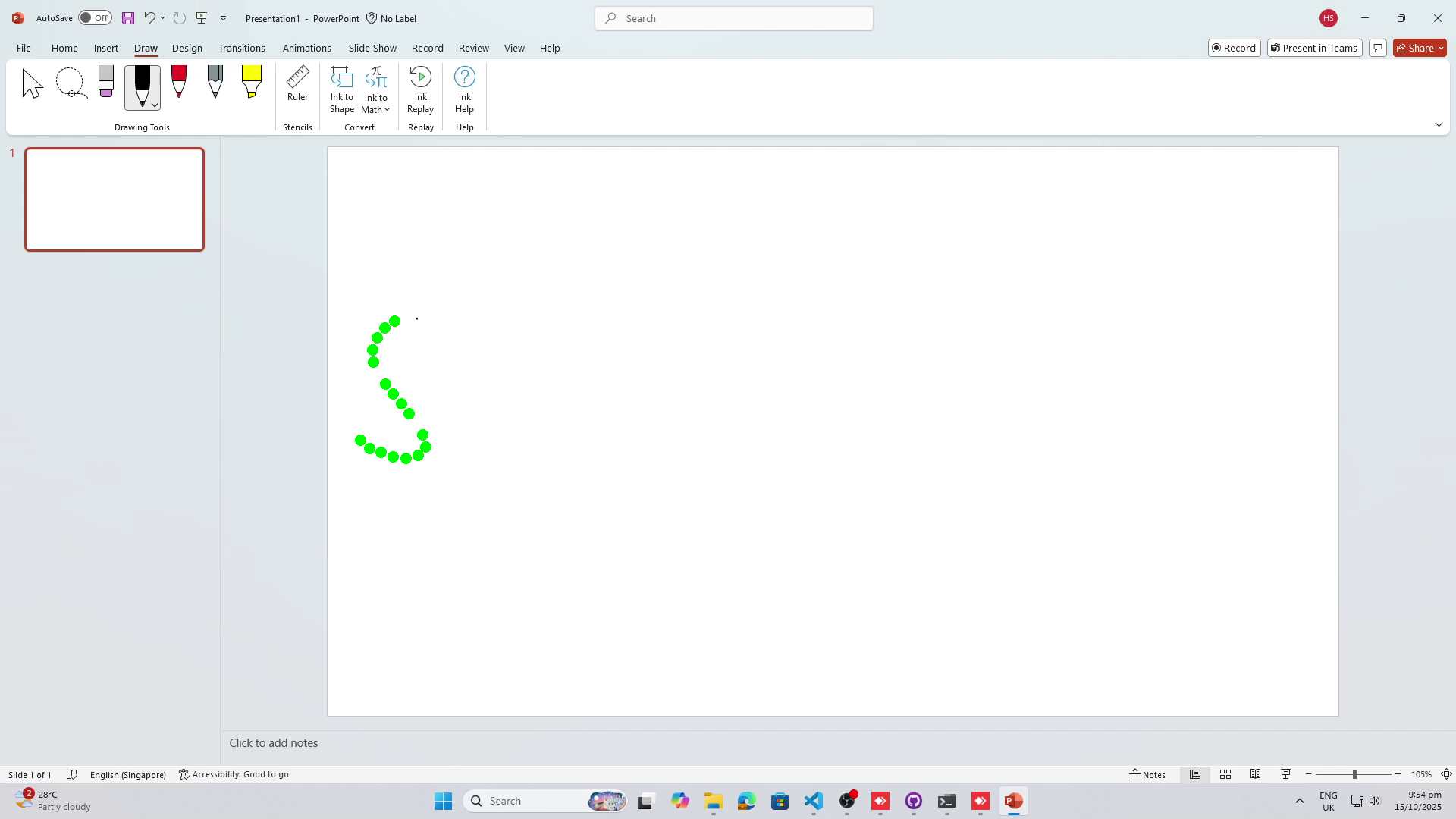} \\
\midrule
Adobe Premiere Pro & Apply Unsharp Mask effect to Space clip &
\includegraphics[width=\linewidth]{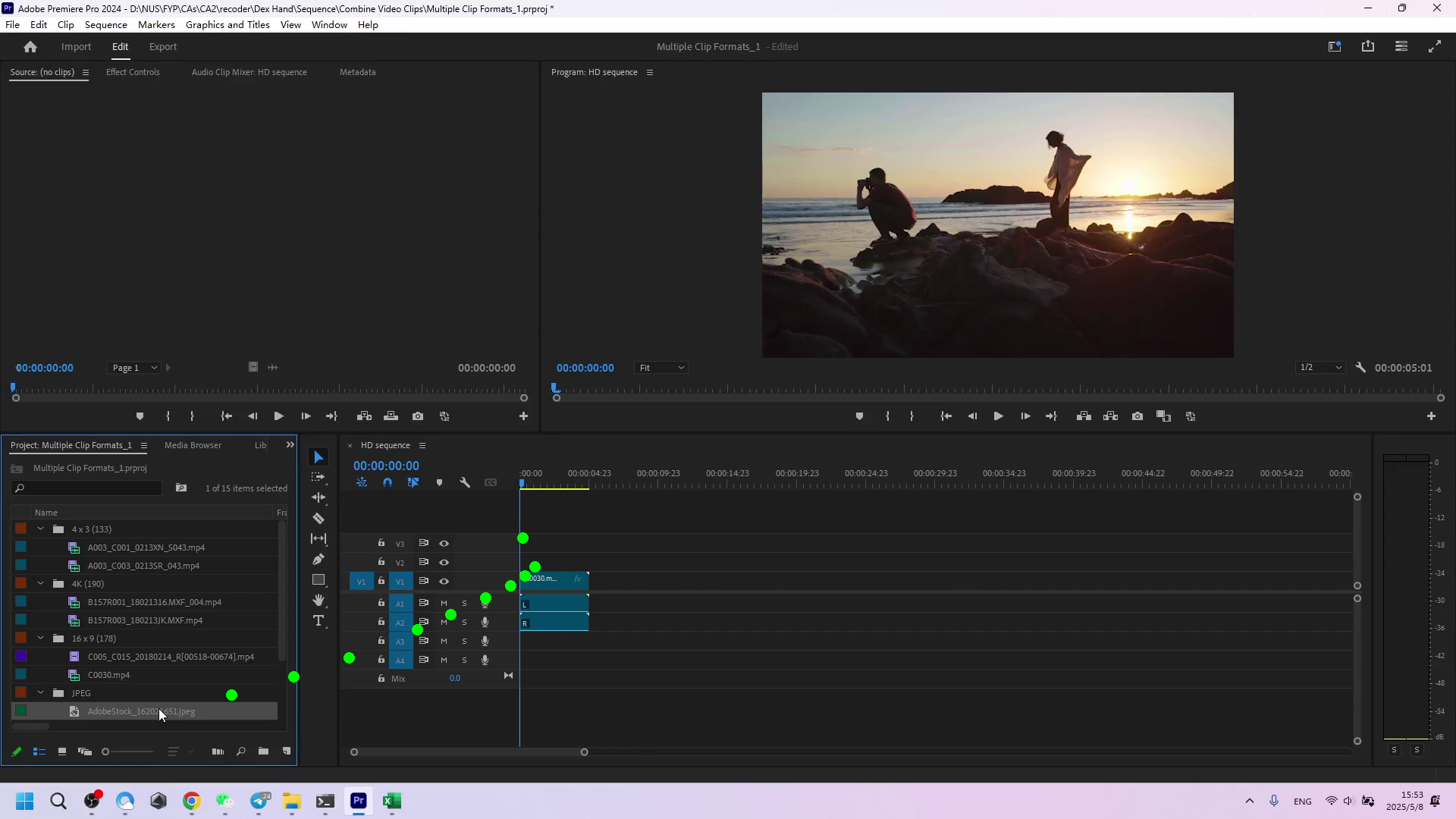} \\
\midrule
Captcha & Solve the rotate Captcha &
\includegraphics[width=\linewidth]{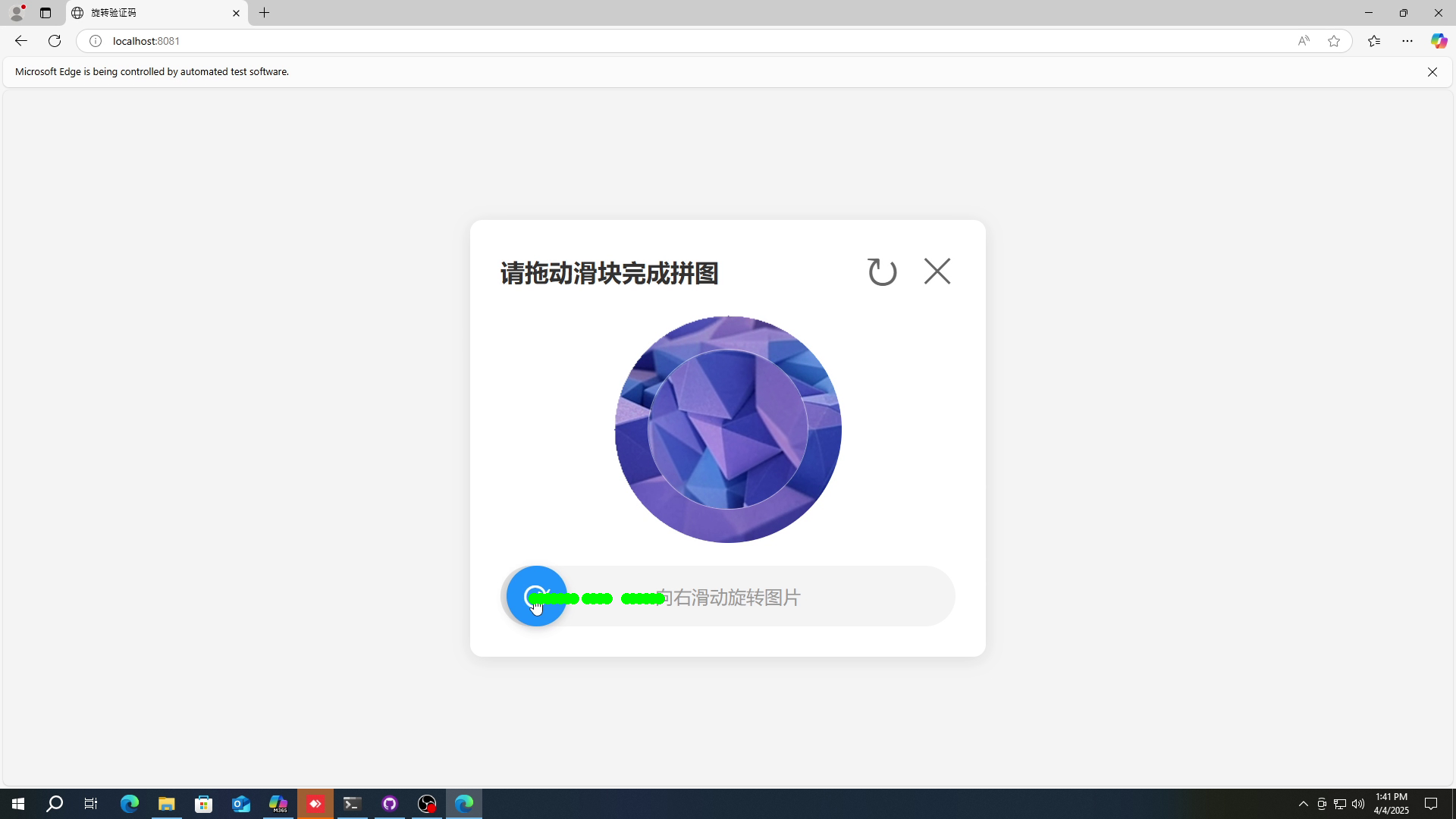} \\
\bottomrule
\end{tabular}%
}
\vspace{-1em}
\caption{\textbf{Illustration of \our~predicted trajectories across four domains.} \our~generates smooth, human-like trajectories that closely follow the instructed paths.}
\vspace{-2em}
\label{tab:vis_table}
\end{table}

\subsection{Qualitative Analysis}
In Fig.\ref{tab:vis_table}, we demonstrate how \our produces precise trajectory generation aligned with task intent across different domains.
In PowerPoint, the model accurately performs the required 45-degree clockwise rotation, producing a trajectory consistent with the instructed geometric transform.
In Handwriting, it generates a smooth, continuous stroke that clearly forms the intended “S” shaped curve in “Starlit grin.”
For Captcha, the model exhibits fine-grained control, stopping at the exact rotation angle needed to solve the puzzle rather than overshooting.
These results collectively show that \our~can follow continuous, high-degree-of-freedom motions with instruction-following.
\section{Conclusion}
We introduce \our, the first flow-based generative GUI agent designed for continuous trajectory control. By unifying discrete clicks and continuous drags within a shared flow-based formulation, and by training on a large corpus of 20K dense trajectories, \our enables precise, real-time cursor adjustment that existing discrete-action GUI agents cannot achieve. To support evaluation, we release ScreenDrag, a comprehensive benchmark covering five domains with both offline and online rollout protocols. Extensive experiments show that current proprietary GUI agents still struggle with continuous manipulation, whereas \our sets a leading results with only 450M parameters. We hope this work lays the foundation for building GUI agents with truly human-like dexterity.

\begin{sloppypar}
\noindent\textbf{Limitations and Future works.}
We trained ShowUI-$\pi$ at a small model size and limited training data scale. In our future work, we plan to scale up the model size with more parameters and also larger training data scale from our data collection pipeline and external data. Meanwhile, We will explore text-centric planning integration with ShowUI-$\pi$.
\end{sloppypar}
{
    \small
    \bibliographystyle{ieeenat_fullname}
    \bibliography{main}
}
\appendix
% \clearpage
% \setcounter{page}{1}
% \maketitlesupplementary

% \section{Implementation Details}

% \subsection{Model Architecture}
% We provide the detailed architecture specifications of our proposed model in Table~\ref{tab:model_arch}.

% \subsection{Training and Hyperparameters}
% We summarize the hyperparameter settings and training details used throughout our experiments in Table~\ref{tab:train_hyperparams}.

% \section{Dataset}
% We describe the datasets used for training and evaluation in detail. Specifically, we discuss data collection methodologies, preprocessing steps, and statistics in Table~\ref{tab:dataset_stat}.

% \section{Additional Experimental Results}
% In this section, we provide supplementary experimental results, including ablation studies and additional comparisons with baselines.

% \subsection{Extended Baseline Comparison}
% Additional comparisons against existing baselines are summarized in Table~\ref{tab:extended_baseline}.

\section{More Results}

\subsection{Grounding Performance of \our}

\begin{table*}[htbp]
\centering
\small
\setlength{\tabcolsep}{0.8pt}
\renewcommand{\arraystretch}{1.18}
\arrayrulecolor{black}

\resizebox{\textwidth}{!}{
\begin{tabular}{l|ccc|ccc|ccc|ccc|ccc|ccc|ccc}
\toprule
\rowcolor{white}
\multirow{2}{*}{\textbf{Model}} &
\multicolumn{3}{c|}{\includegraphics[width=0.015\textwidth]{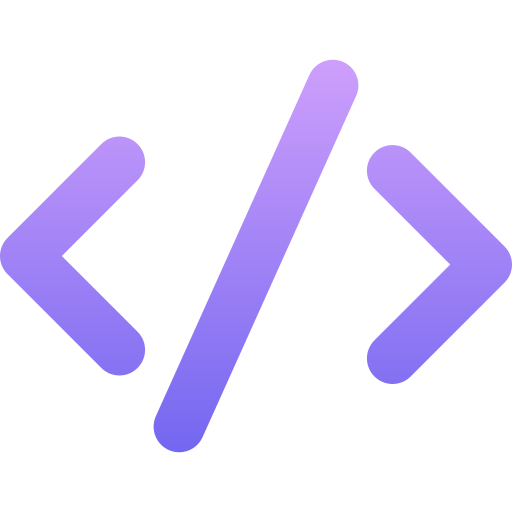} \textbf{Development}} &
\multicolumn{3}{c|}{\includegraphics[width=0.015\textwidth]{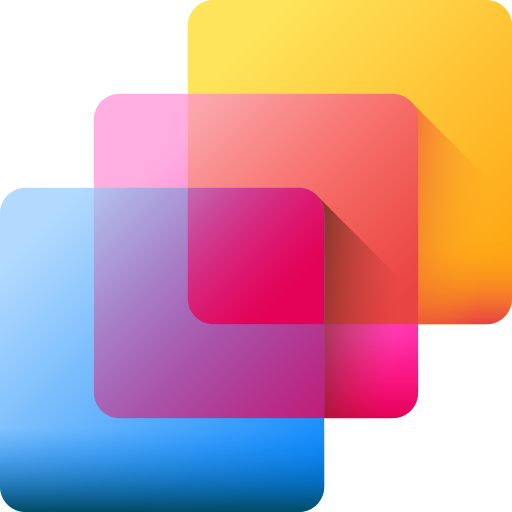} \textbf{Creative}} &
\multicolumn{3}{c|}{\includegraphics[width=0.015\textwidth]{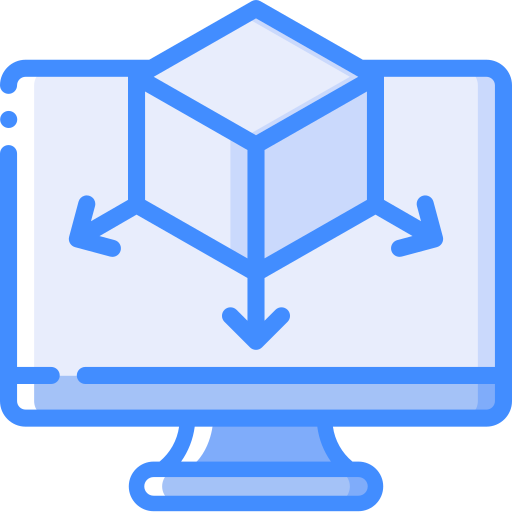} \textbf{CAD}} &
\multicolumn{3}{c|}{\includegraphics[width=0.015\textwidth]{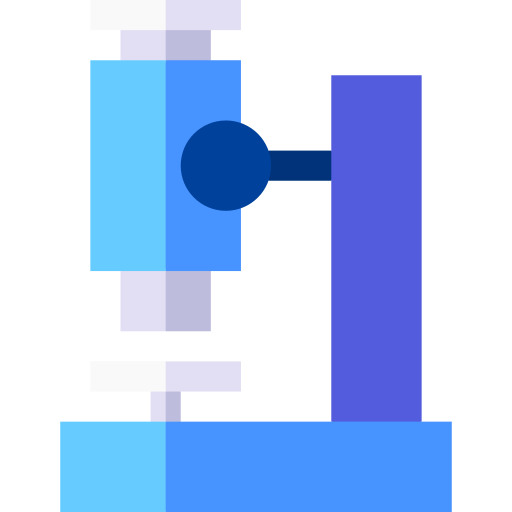} \textbf{Scientific}} &
\multicolumn{3}{c|}{\includegraphics[width=0.015\textwidth]{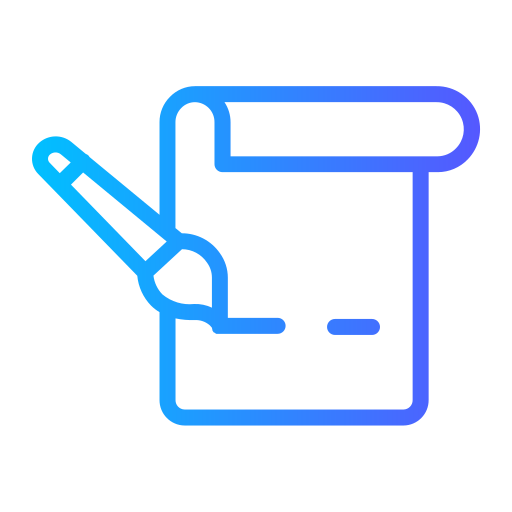} \textbf{Office}} &
\multicolumn{3}{c|}{\includegraphics[width=0.015\textwidth]{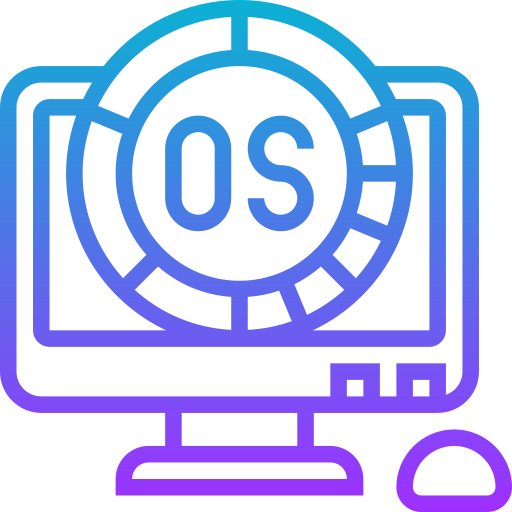} \textbf{OS}} &
\multicolumn{3}{c}{\textbf{Avg}} \\
\rule{0pt}{2.3ex} &
\textbf{Text} & \textbf{Icon} & \textbf{Avg} &
\textbf{Text} & \textbf{Icon} & \textbf{Avg} &
\textbf{Text} & \textbf{Icon} & \textbf{Avg} &
\textbf{Text} & \textbf{Icon} & \textbf{Avg} &
\textbf{Text} & \textbf{Icon} & \textbf{Avg} &
\textbf{Text} & \textbf{Icon} & \textbf{Avg} &
\textbf{Text} & \textbf{Icon} & \textbf{Avg} \\
\midrule

% \rowcolor{BaselineGray}
OSAtlas-7B & 33.1 & 1.4 & 17.7 & 28.8 & 2.8 & 17.9 & 12.2 & 4.7 & 10.3 & 37.5 & 7.3 & 24.4 & 33.9 & 5.7 & 27.4 & 27.1 & 4.5 & 16.8 & 28.1 & 4.0 & 18.9 \\

% \rowcolor{BaselineGray}
UGround (7B) & 26.6 & 2.1 & 14.7 & 27.3 & 2.8 & 17.0 & 14.2 & 1.6 & 11.1 & 31.9 & 2.7 & 19.3 & 31.6 & 11.3 & 27.0 & 17.8 & 0.0 & 9.7 & 25.0 & 2.8 & 16.5 \\

% \rowcolor{BaselineGray}
AriaUI (MOE, 3.9B active) & 16.2 & 0.0 & 8.4 & 23.7 & 2.1 & 14.7 & 7.6 & 1.6 & 6.1 & 27.1 & 6.4 & 18.1 & 20.3 & 1.9 & 16.1 & 4.7 & 0.0 & 2.6 & 17.1 & 2.0 & 11.3 \\

% \rowcolor{BaselineGray}
CogAgent (18B) & 14.9 & 0.7 & 8.0 & 9.6 & 0.0 & 5.6 & 7.1 & 3.1 & 6.1 & 22.2 & 1.8 & 13.4 & 13.0 & 0.0 & 10.0 & 5.6 & 0.0 & 3.1 & 12.0 & 0.8 & 7.7 \\

% \rowcolor{BaselineGray}
ShowUI (2B) & 16.9 & 1.4 & 9.4 & 9.1 & 0.0 & 5.3 & 2.5 & 0.0 & 1.9 & 13.2 & 7.3 & 10.6 & 15.3 & 7.5 & 13.5 & 10.3 & 2.2 & 6.6 & 10.8 & 2.6 & 7.7 \\

\rowcolor{BandBlue}
\textcolor{OursBlue}{\textbf{ShowUI-\ensuremath{\pi} (450M)}} &
9.7 & 5.5 & 7.7 & 10.1 & 2.8 & 7.0 & 1.0 & 0.0 & 0.8 & 9.7 & 6.4 & 8.3 & 10.2 & 1.9 & 8.3 & 3.7 & 3.4 & 3.6 & 7.5 & 3.8 & 6.1 \\

% \rowcolor{BaselineGray}
OSAtlas-4B & 7.1 & 0.0 & 3.7 & 3.0 & 1.4 & 2.3 & 2.0 & 0.0 & 1.5 & 9.0 & 5.5 & 7.5 & 5.1 & 3.8 & 4.8 & 5.6 & 0.0 & 3.1 & 5.0 & 1.7 & 3.7 \\

% \rowcolor{BaselineGray}
MiniCPM-V (7B) & 7.1 & 0.0 & 3.7 & 2.0 & 0.0 & 1.2 & 4.1 & 1.6 & 3.4 & 8.3 & 0.0 & 4.7 & 2.8 & 3.8 & 3.0 & 3.7 & 1.1 & 2.6 & 4.5 & 0.7 & 3.0 \\

% \rowcolor{BaselineGray}
Qwen2-VL-7B & 2.6 & 0.0 & 1.3 & 1.5 & 0.0 & 0.9 & 0.5 & 0.0 & 0.4 & 6.3 & 0.0 & 3.5 & 3.4 & 1.9 & 3.0 & 0.9 & 0.0 & 0.5 & 2.5 & 0.2 & 1.6 \\

% \rowcolor{BaselineGray}
SeeClick (7B) & 0.6 & 0.0 & 0.3 & 1.0 & 0.0 & 0.6 & 2.5 & 0.0 & 1.9 & 3.5 & 0.0 & 2.0 & 1.1 & 0.0 & 0.9 & 2.8 & 0.0 & 1.5 & 1.8 & 0.0 & 1.1 \\

% \rowcolor{BaselineGray}
GPT-4o & 1.3 & 0.0 & 0.7 & 1.0 & 0.0 & 0.6 & 2.0 & 0.0 & 1.5 & 2.1 & 0.0 & 1.2 & 1.1 & 0.0 & 0.9 & 0.0 & 0.0 & 0.0 & 1.3 & 0.0 & 0.8 \\

% \rowcolor{BaselineGray}
QwenVL-7B & 0.0 & 0.0 & 0.0 & 0.0 & 0.0 & 0.0 & 0.0 & 0.0 & 0.0 & 0.7 & 0.0 & 0.4 & 0.0 & 0.0 & 0.0 & 0.0 & 0.0 & 0.0 & 0.1 & 0.0 & 0.1 \\
\bottomrule
\end{tabular}
}

\arrayrulecolor{black}
\vspace{-1em}
\caption{\textbf{Performance breakdown of various models across application categories on ScreenSpot-Pro.}}
\vspace{-2em}
\label{tab:grounding:screenspotpro}
\end{table*}

To evaluate the grounding performance of \our as a VLA unifying both discrete click and continuous drag actions, we evaluate \our on one of the most challenging grounding benchmarks, ScreenSpot-Pro. During the grounding evaluation, we use the first coordinate in the generated action chunk as the model prediction for grounding. As demonstrated in Tab.~\ref{tab:grounding:screenspotpro}, \our achieves performance comparable to much larger baseline models. Due to the small parameter size (450M), the vision capacity of \our is weaker than other models, thus restricting the grounding performance.

\subsection{Drag Performance of \our on public benchmark}

% \begin{table}[ht]
% \centering
% \begin{tabular}{l cc}
% \toprule
% \textbf{Model} & \multicolumn{2}{c}{\textbf{Drag}} \\
% \cmidrule(lr){2-3}
% & Dist. $\downarrow$ & Recall $\uparrow$ \\
% \midrule
% CogAgent (18B)           & 44.7 & 0.0 \\
% Qwen-VL-Max (72B)       & 42.0 & 0.3 \\
% Gemini-Pro-Vision   & 40.8 & 0.0 \\
% Claude-3-Opus       & 30.6 & 1.7 \\
% GPT-4-Turbo         & 31.3 & 1.4 \\
% \textbf{ShowUI-\ensuremath{\pi} (450M)} & 41.6 & 1.7 \\
% GPT-4o              & 21.9 & 2.5 \\
% \bottomrule
% \end{tabular}
% \caption{\textbf{Drag performance of \our and baseline models.} 
% Following VideoGUI's setting, two metrics are used in the evaluation. 
% The distance (Dist) is normalized and then multiplied by 100, reflecting the average offset between the predicted and ground-truth drag endpoints. 
% The recall (Rec) is the percentage of drags whose predicted start and end points both fall within a 100-pixel threshold of the corresponding ground-truth endpoints. Note that VideoGUI-Action is image-based and only provides the initial screenshot.}
% \label{tab:drag:videogui_action_drag}
% \end{table}

\begin{table}[htbp]
\centering
\small
\setlength{\tabcolsep}{6pt}
\renewcommand{\arraystretch}{1.18}
\arrayrulecolor{black}
\begin{tabular}{l cc}
\toprule
\rowcolor{white}
\textbf{Model} & \multicolumn{2}{c}{\textbf{Drag}} \\
\cmidrule(lr){2-3}
\rule{0pt}{2.3ex} & Dist. $\downarrow$ & Recall $\uparrow$ \\
\midrule
% \rowcolor{BaselineGray}
CogAgent (18B)           & 44.7 & 0.0 \\
% \rowcolor{BaselineGray}
Qwen-VL-Max (72B)        & 42.0 & 0.3 \\
% \rowcolor{BaselineGray}
Gemini-Pro-Vision        & 40.8 & 0.0 \\
% \rowcolor{BaselineGray}
Claude-3-Opus            & 30.6 & 1.7 \\
% \rowcolor{BaselineGray}
GPT-4-Turbo              & 31.3 & 1.4 \\
\rowcolor{BandBlue}
\textcolor{OursBlue}{\textbf{ShowUI-\ensuremath{\pi} (450M)}} & 41.6 & 1.7 \\
% \rowcolor{BaselineGray}
GPT-4o                   & 21.9 & 2.5 \\
\bottomrule
\end{tabular}
\arrayrulecolor{black}
\vspace{-0.8em}
\caption{\textbf{Drag performance of \our and baseline models.}
Following VideoGUI's setting, two metrics are used in the evaluation.
The distance (Dist) is normalized and then multiplied by 100, reflecting the average offset between the predicted and ground-truth drag endpoints.
The recall (Rec) is the percentage of drags whose predicted start and end points both fall within a 100-pixel threshold of the corresponding ground-truth endpoints. Note that VideoGUI-Action is image-based and only provides the initial screenshot.}
\vspace{-1.2em}
\label{tab:drag:videogui_action_drag}
\end{table}

To evaluate the drag performance of \our on public dataset, we evaluate \our on the \texttt{Drag} subset of VideoGUI-Action benchmark where drag performance is specifically reported. As the benchmark is image-based, only one screenshot is provided as the visual input. Therefore, we let \our generate the entire action chunk at once and use it as the model prediction for dragging, without observations on-the-fly or continuous actions. This restricts \our's performance, as its drag capability is trained on the continuous observations from \bench only, without the drag data from public datasets, thus leading to a gap between training and public benchmark evaluation. However, \our still achieves performance comparable to much larger baseline models, as demonstrated in Tab.~\ref{tab:drag:videogui_action_drag}. 

\subsection{Effects of Co-training with Drag Data and Grounding Data}

\begin{table*}[htbp]
\centering
\small
\setlength{\tabcolsep}{0.8pt}
\renewcommand{\arraystretch}{1.18}
\arrayrulecolor{black}

\resizebox{\textwidth}{!}{
\begin{tabular}{l|ccc|ccc|ccc|ccc|ccc|ccc|ccc}
\toprule
\rowcolor{white}
\multirow{2}{*}{\textbf{Data Recipe}} &
\multicolumn{3}{c|}{\includegraphics[width=0.015\textwidth]{appendix/symbols/dev.png} \textbf{Development}} &
\multicolumn{3}{c|}{\includegraphics[width=0.015\textwidth]{appendix/symbols/creative.png} \textbf{Creative}} &
\multicolumn{3}{c|}{\includegraphics[width=0.015\textwidth]{appendix/symbols/cad.png} \textbf{CAD}} &
\multicolumn{3}{c|}{\includegraphics[width=0.015\textwidth]{appendix/symbols/scientific.png} \textbf{Scientific}} &
\multicolumn{3}{c|}{\includegraphics[width=0.015\textwidth]{appendix/symbols/office.png} \textbf{Office}} &
\multicolumn{3}{c|}{\includegraphics[width=0.015\textwidth]{appendix/symbols/os.png} \textbf{OS}} &
\multicolumn{3}{c}{\textbf{Avg}} \\
\rule{0pt}{2.3ex} &
\textbf{Text} & \textbf{Icon} & \textbf{Avg} &
\textbf{Text} & \textbf{Icon} & \textbf{Avg} &
\textbf{Text} & \textbf{Icon} & \textbf{Avg} &
\textbf{Text} & \textbf{Icon} & \textbf{Avg} &
\textbf{Text} & \textbf{Icon} & \textbf{Avg} &
\textbf{Text} & \textbf{Icon} & \textbf{Avg} &
\textbf{Text} & \textbf{Icon} & \textbf{Avg} \\
\midrule
% \rowcolor{BaselineGray}
Drag & 1.3 & 0.0 & 0.7 & 0.5 & 2.8 & 1.5 & 1.0 & 1.6 & 1.1 & 0.7 & 0.0 & 0.4 & 0.6 & 3.8 & 1.3 & 2.8 & 0.0 & 1.5 & 1.0 & 1.2 & 1.1 \\
% \rowcolor{BaselineGray}
Grounding & 8.4 & 4.8 & 6.7 & 8.1 & 2.1 & 5.6 & 0.5 & 0.0 & 0.4 & 12.5 & 5.5 & 9.4 & 8.5 & 1.9 & 7.0 & 7.5 & 7.9 & 7.7 & 7.3 & 4.0 & 6.0 \\
\rowcolor{BandBlue}
\textcolor{OursBlue}{\textbf{Drag + Grounding}} &
9.7 & 5.5 & 7.7 & 10.1 & 2.8 & 7.0 & 1.0 & 0.0 & 0.8 & 9.7 & 6.4 & 8.3 & 10.2 & 1.9 & 8.3 & 3.7 & 3.4 & 3.6 & 7.5 & 3.8 & 6.1 \\
\bottomrule
\end{tabular}
}

\arrayrulecolor{black}
\vspace{-1em}
\caption{\textbf{Performance breakdown of models trained using different data recipe across application categories on ScreenSpot-Pro.}}
% \vspace{-2em}
\label{tab:data_recipe:screenspotpro}
\end{table*}
\begin{table*}[htbp]
\centering
\small
\setlength{\tabcolsep}{6pt}
\renewcommand{\arraystretch}{1.18}
\arrayrulecolor{black}
\begin{tabular}{lcccccc}
\toprule
\rowcolor{white}
\multirow{2}{*}{\textbf{Data Recipe}} &
\multicolumn{6}{c}{\textbf{Online Evaluation.} Avg. Success Rate($\uparrow$).} \\
% \rowcolor{white}
\rule{0pt}{2.3ex} &
OS & PowerPoint & Premiere & Captcha & Handwriting & Overall \\
\midrule
% \rowcolor{BaselineGray}
Grounding  & 0.00  & 0.00 & 0.00 & 0.00 & 0.00 & 0.00 \\
% \rowcolor{BaselineGray}
Drag & 20.79 & 10.34 & 4.62 & 48.15 & 25.74 & 21.92 \\
\rowcolor{BandBlue}
\textcolor{OursBlue}{\textbf{Drag + Grounding}} &
\best{13.11} & \best{22.93} & \best{8.64} & \best{55.91} & \best{34.32} & \best{26.98} \\
\bottomrule
\end{tabular}
\arrayrulecolor{black}
\vspace{-1em}
\caption{\textbf{Performance breakdown of models trained using different data recipe across different domains on \bench online evaluation.}}
% \vspace{-2em}
\label{tab:data_recipe:drag}
\end{table*}

We also evaluate the effects of training \our on the combination of \texttt{Drag} data and \texttt{Grounding} data. As shown in Fig.~\ref{tab:data_recipe:drag} and Fig.~\ref{tab:data_recipe:screenspotpro}, training on both types of actions leads to the highest performance on both \bench online evaluation and Screenshot-Pro benchmarks. Interestingly, after training on \bench drag data, the performance on the \texttt{Creative} and \texttt{Office} categories in ScreenSpot-Pro are boosted, as there is a large amount of data in \bench within the corresponding domains,~\eg, PowerPoint and Premiere Pro. Moreover, training on grounding data can help the model learn a better representation, thus bringing the gain. As the grounding data contains only \texttt{Click} data, the model trained on grounding data solely is unable to generate valid drag trajectories, thus failing the drag tasks.
\section{Setup}
\subsection{Training Details}
We utilize 4 H200 GPUs for training. The batch size per GPU is set to 64, without gradient accumulation. We use \texttt{bfloat16} precision for training. The vision encoder and language model part are initialized with the weights from SmolVLM, and other components,~\eg, action state embedding, action expert,~\etc, are randomly initialized. The model is trained end-to-end thanks to its small parameter size. To enhance efficiency, we resize the visual observation to $(1024, 576)$ from the common $(1920, 1080)$, reducing the vision encoder overhead, while maintaining the $16:9$ scale. 
We leverage DeepSpeed Zero-2 as the training framework.
The learning rate is configured to 1e-4.

\subsection{Training Data}
We use a smaller training corpus for \our compared to other models. Specifically, in addition to the training data from \bench, we use the desktop \texttt{Click} subset of GUIAct, WaveUI, UGround, and ShowUI-Desktop. We did not use any \texttt{non-Click} data from any public dataset, nor did we use any mobile data for training. 
\section{Dataset Construction}
\subsection{How we collect raw data}
We construct raw demonstration data across five domains: PowerPoint, OS Desktop and File Manager, Captcha, Handwriting, as well as Adobe Premiere Pro.  
Across all domains, data are collected on Windows machines where we record high-frequency mouse events and screen video recordings, using our \bench data pipeline. To obtain the UI metadata, the DOM is used for Captcha on webpages and the UIA framework is used for other domains, so that each trajectory can be paired with precise element locations and attributes.

\noindent\textbf{PowerPoint.}
For the PowerPoint domain, we start from the official Microsoft template gallery.  
We crawl and download a diverse set of slide templates spanning different layouts, color schemes, and typography.  
For each template, we automatically parse the slide metadata,~\eg, textboxes, images and shapes, using UIA, and then design manipulation tasks such as rotating and different types of resizing. The position of rotation handle is calculated using heuristics, as its position is not in the UIA metadata.

\noindent\textbf{Captcha.}
For Captcha tasks, we build the automated data collection pipeline on an open-source library Go-Captcha.  
We configure the pipeline to generate interactive Captchas such as sliders and puzzle pieces embedded in a webpage, and the Captcha will be refreshed and regenerated after the previous one is solved.  
We modify the Captcha library codebase so that the task metadata,~\eg, puzzle piece position, target position,~\etc, and the task status,~\eg, success or failure are accessible at real-time. Therefore, we can filter the successfully solved tasks and collect their trajectories. 

\noindent\textbf{Handwriting.}
For the Handwriting domain, we build upon an open-source handwriting synthesis library.  
We sample diverse names, short phrases from a Qwen2.5-72B endpoint deployed using VLLM, then the handwriting trajectory from the sampled text will be generated using the handwriting synthesis library with varied stroke styles. Afterwards, the trajectory will be written on the canvas through Win32 mouse interface for fine-grained control.  

\noindent\textbf{Adobe Premiere Pro.}
In the Premiere Pro domain, we capture human demonstration trajectories in creative workflows instead of automation, due to the limitation that Premiere Pro UI metadata is not fully accessible.  
We recruit two student annotators experienced with video editing to design tasks that reflect common real-world operations, such as trimming clips, adjusting layers on the timeline, arranging clips, and applying effects.
The expert demonstrations are recorded using our recorder that can capture high-frequency mouse events with low latency, aligned with the video recording timestamps.

\noindent\textbf{OS Desktop and File Manager.}
For OS Desktop and File Manager tasks, we build an automated pipeline that creates different types of files and folders on the desktop or file manager windows, then performs drags and records. To increase the task difficulty, we modify the Windows registry so that files and folders created can be placed anywhere on the desktop without automatic arrangement, and each time multiple files and folders with different names will be generated. UIA is used to obtain bounding boxes for desktop icons, folders, and window controls.

\noindent\textbf{Data Recording.}
Across all domains, we use OBS for the screen recording, as it is well optimized and has less latency compared to other screen capture methods,~\eg, FFmpeg. 

\noindent\textbf{Data Generation Codebase.}
To contribute to the community, we will open-source the data generation and recording codebase. Moreover, some of the data crawl sources,~\eg, the Microsoft PowerPoint Template Gallery, have changed and make the template collection more difficult,~\ie, only a limited number of templates will be shown without search queries, therefore, we will also provide our collected raw data.

\subsection{Data visualization}

\newcolumntype{D}[1]{>{\centering\arraybackslash}m{#1}}
\newcolumntype{L}[1]{>{\raggedright\arraybackslash}m{#1}}
\newcolumntype{C}[1]{>{\centering\arraybackslash}m{#1}}

\begin{table*}[t]
\centering
\caption{\textbf{Examples of task trajectories from five domains.}
Three frames from the episode are shown for each task.}
\label{tab:data_observations}
\setlength{\tabcolsep}{3pt}
\renewcommand{\arraystretch}{1.1}
\begin{tabular}{D{0.09\textwidth} L{0.35\textwidth} C{0.18\textwidth} C{0.18\textwidth} C{0.18\textwidth}}
\toprule
\textbf{Domain} & \textbf{Task} & \textbf{Initial} & \textbf{Intermediate} & \textbf{Final} \\
\midrule
%------------------ Captcha (3 rows of images, adjusted multirow count to ~14) ------------------
\multirow{9}{*}{Captcha} &
Solve the Drag-and-Drop Captcha &
\includegraphics[width=\linewidth]{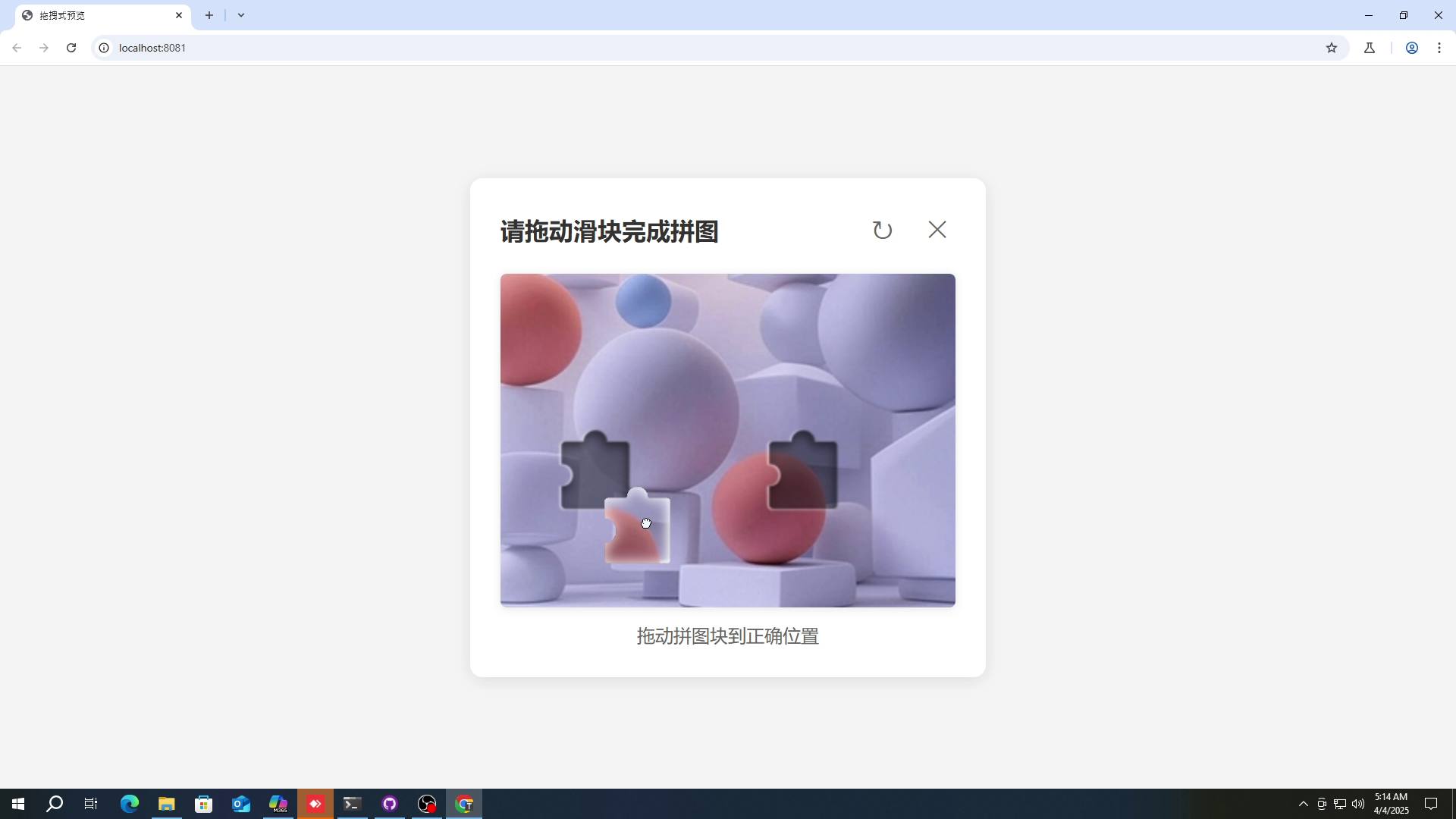} &
\includegraphics[width=\linewidth]{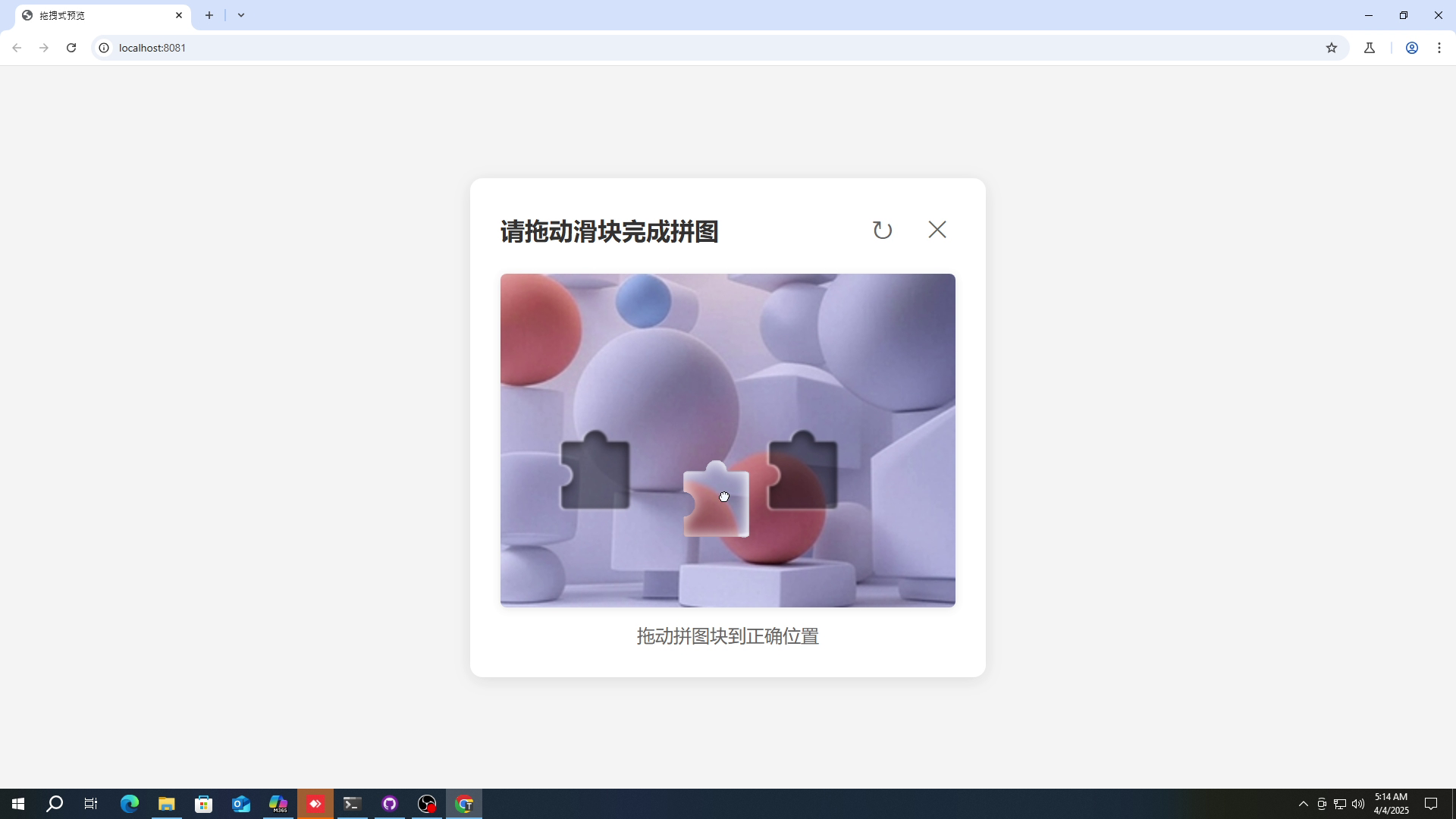} &
\includegraphics[width=\linewidth]{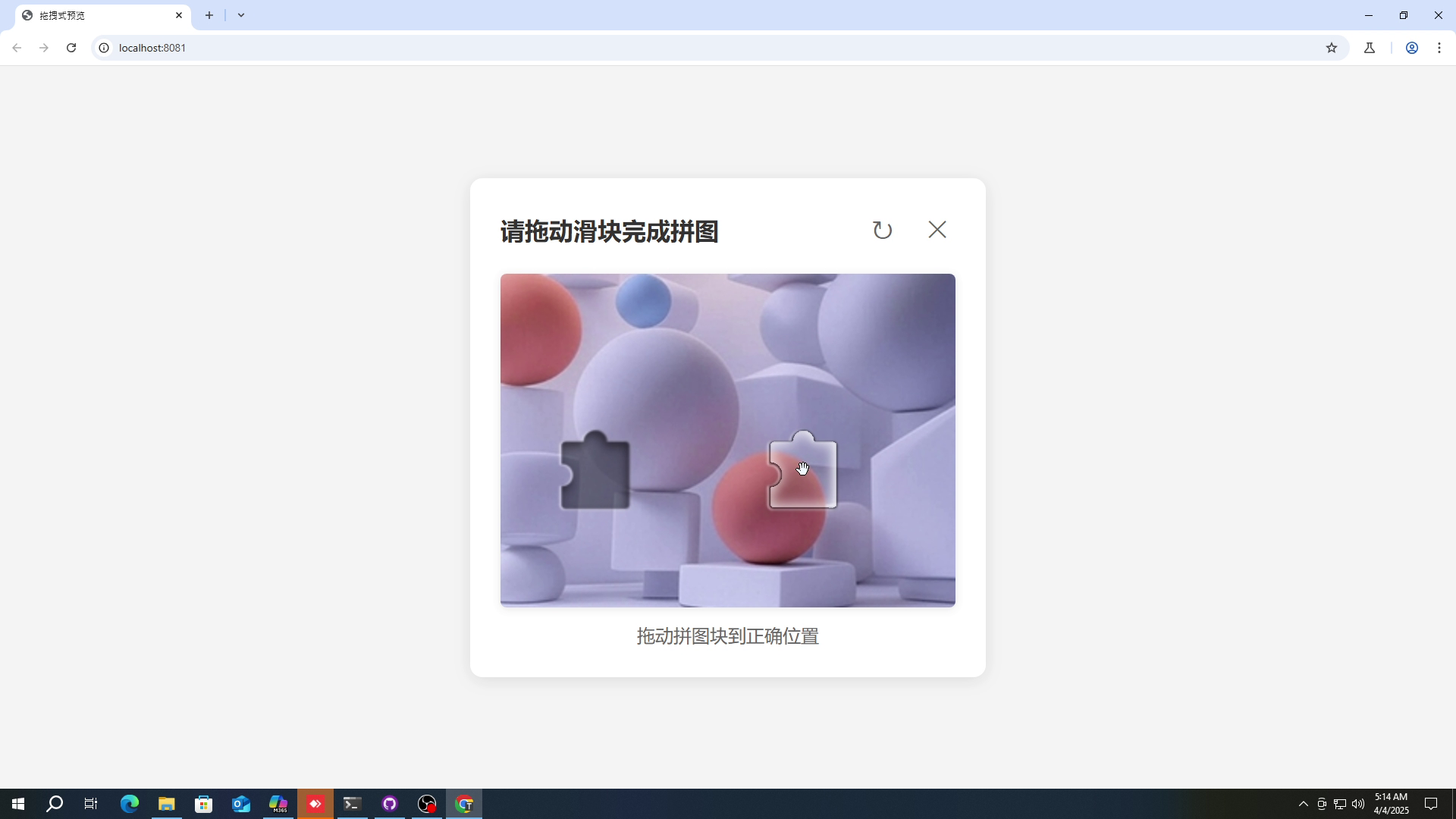} \\
& Solve the Rotate Captcha &
\includegraphics[width=\linewidth]{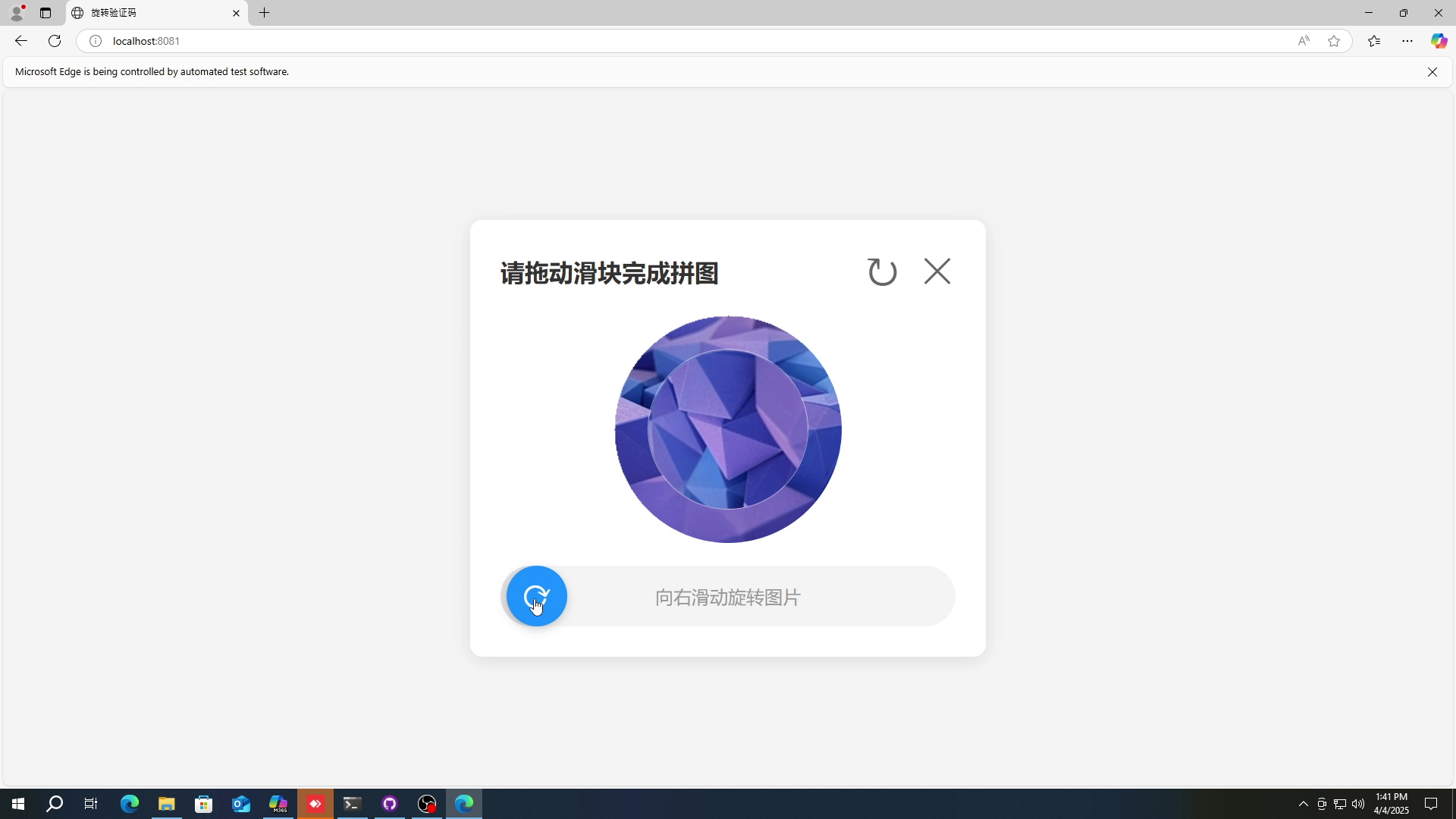} &
\includegraphics[width=\linewidth]{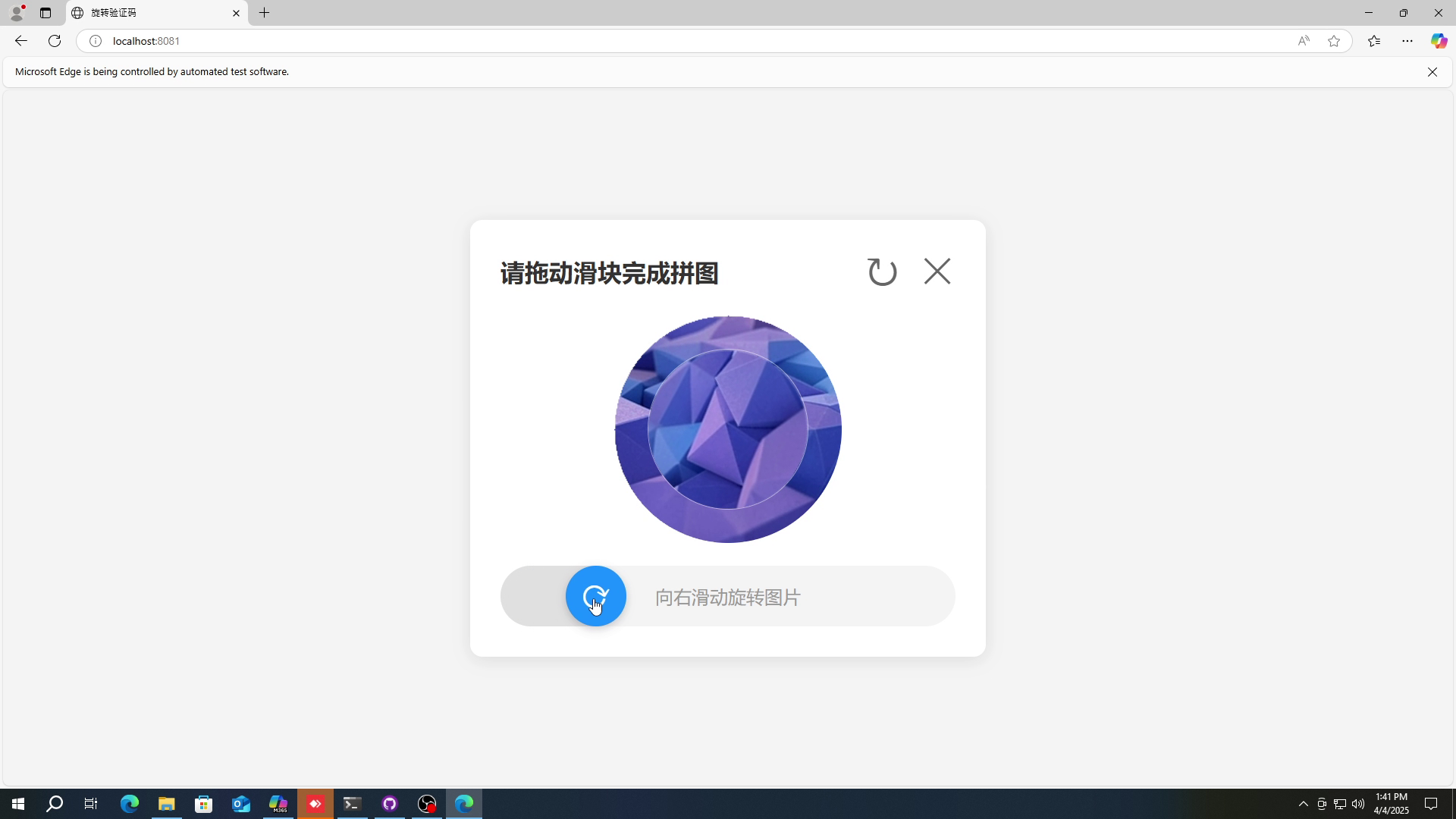} &
\includegraphics[width=\linewidth]{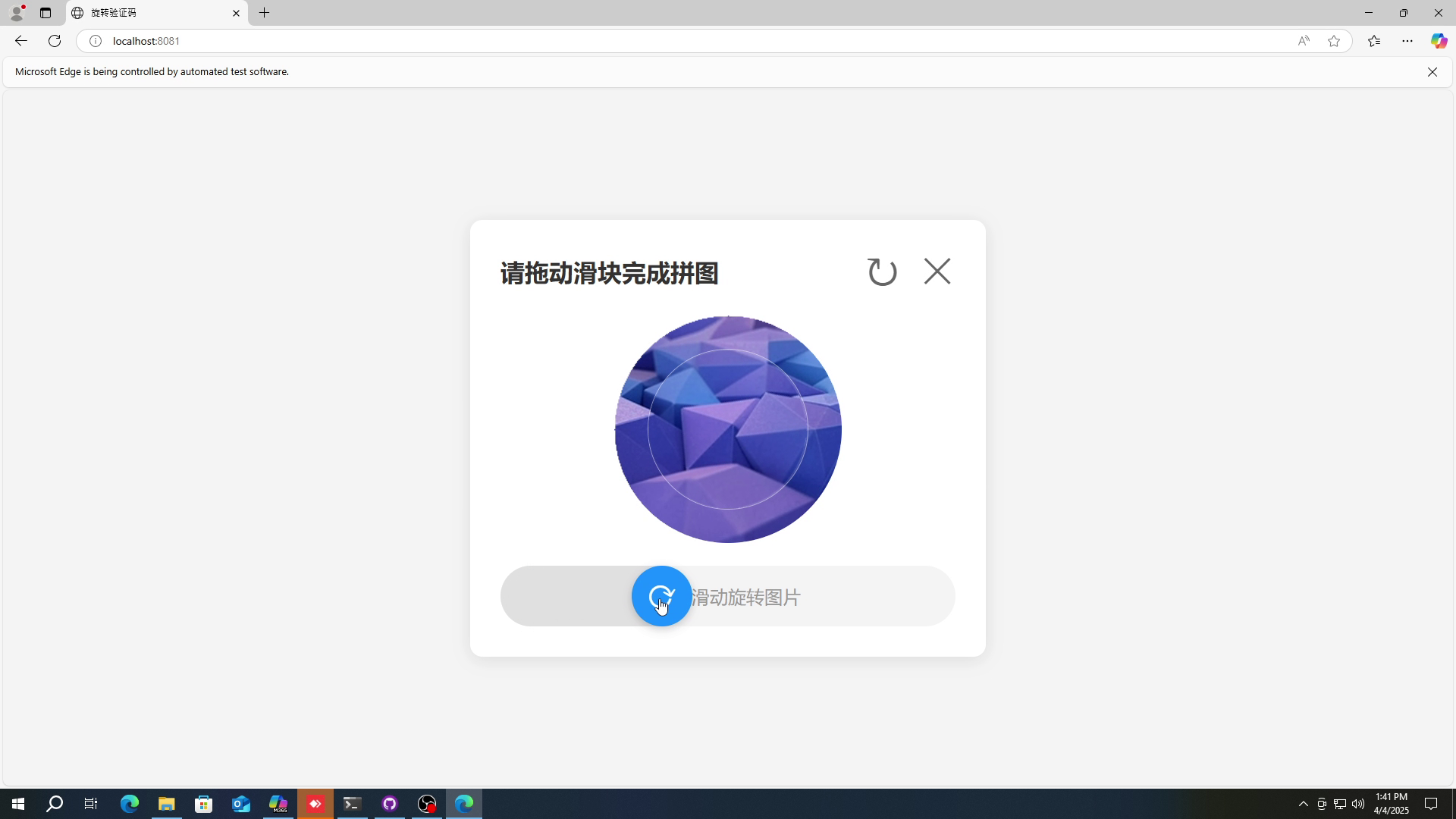} \\
& Solve the Slider Captcha &
\includegraphics[width=\linewidth]{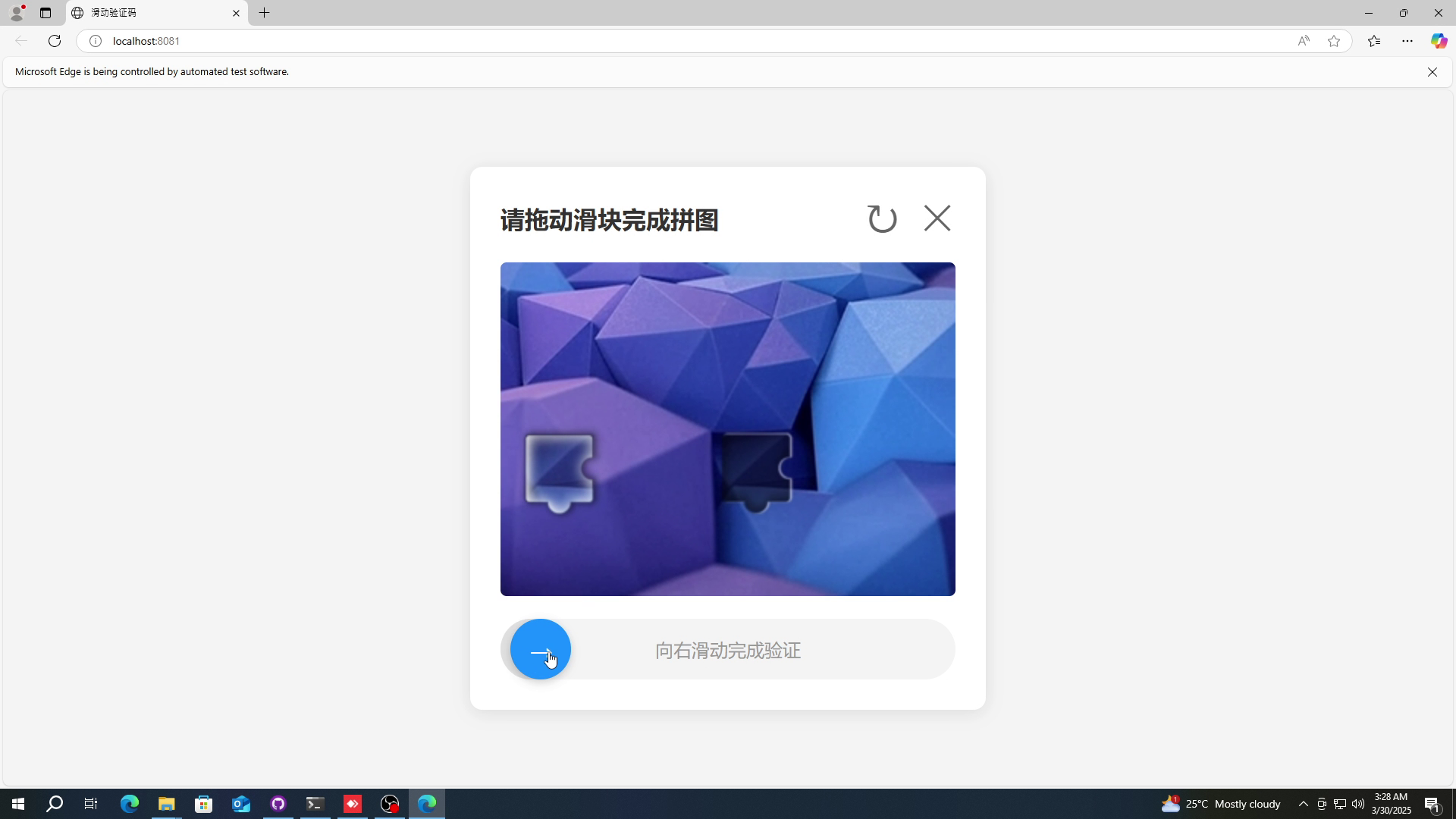} &
\includegraphics[width=\linewidth]{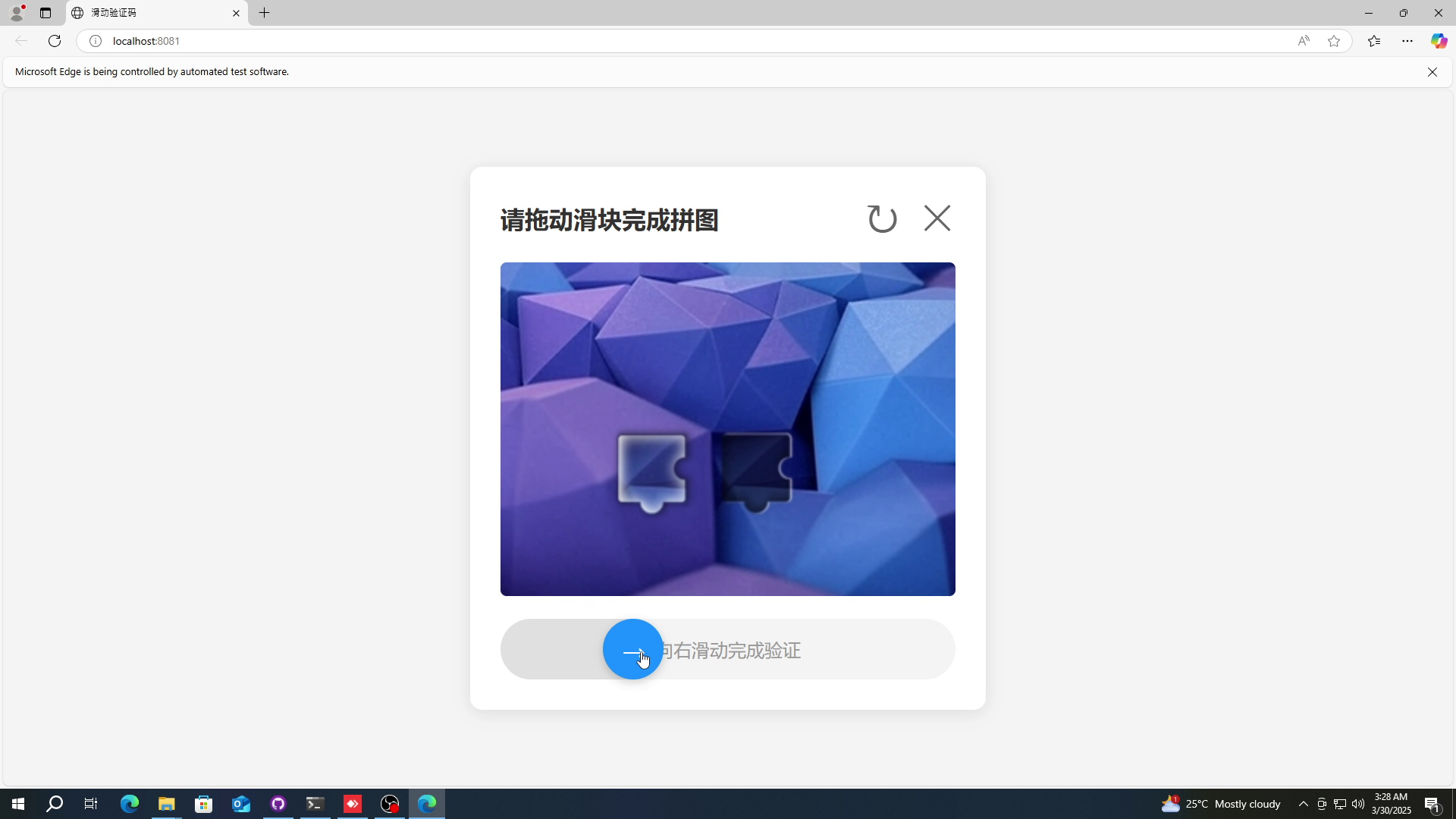} &
\includegraphics[width=\linewidth]{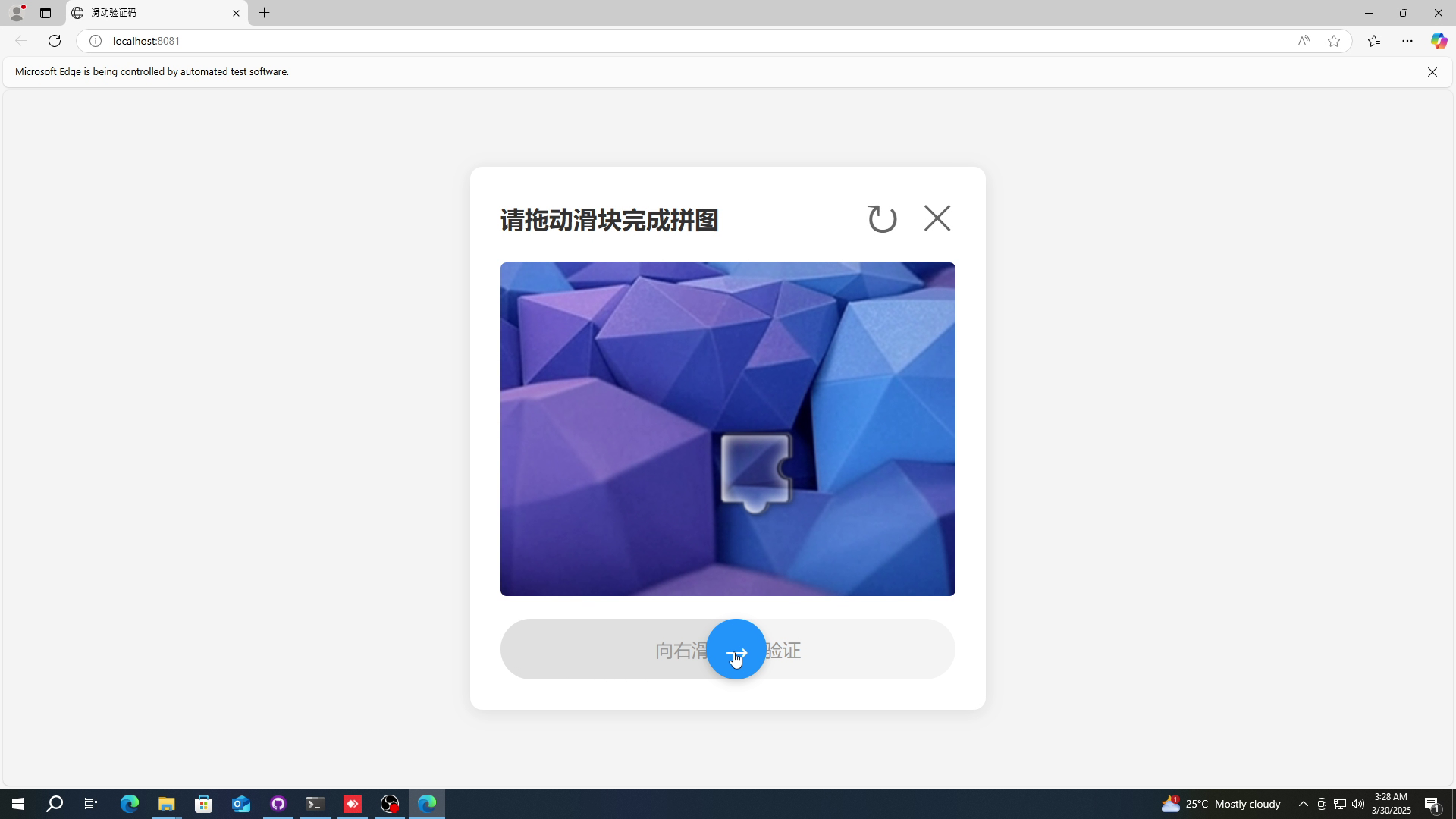} \\
\midrule
%------------------ Desktop (2 rows of images, adjusted multirow count to ~9) ------------------
\multirow{5}{*}{Desktop} &
Drag Analysis.xlsx to MyProject &
\includegraphics[width=\linewidth]{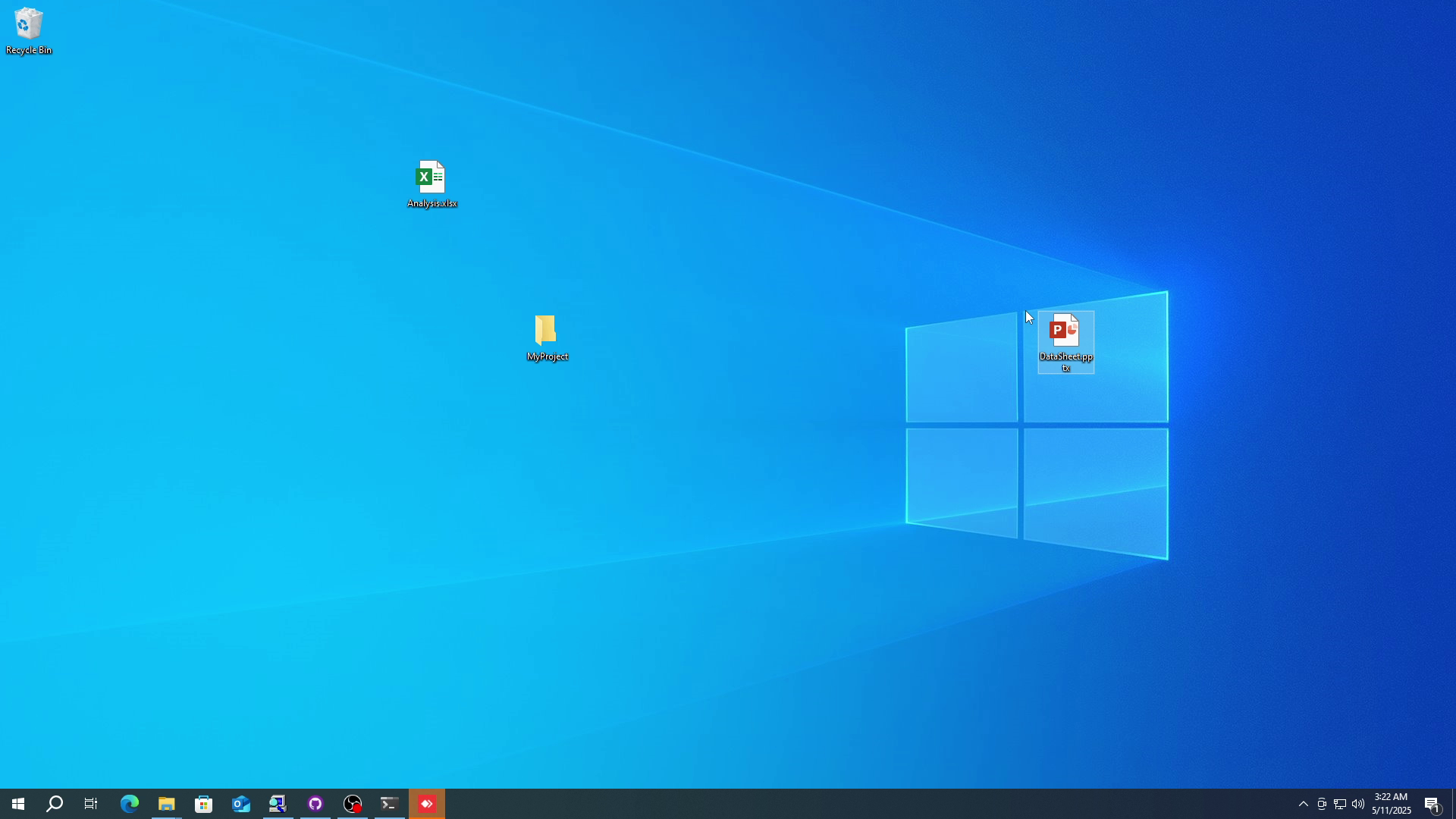} &
\includegraphics[width=\linewidth]{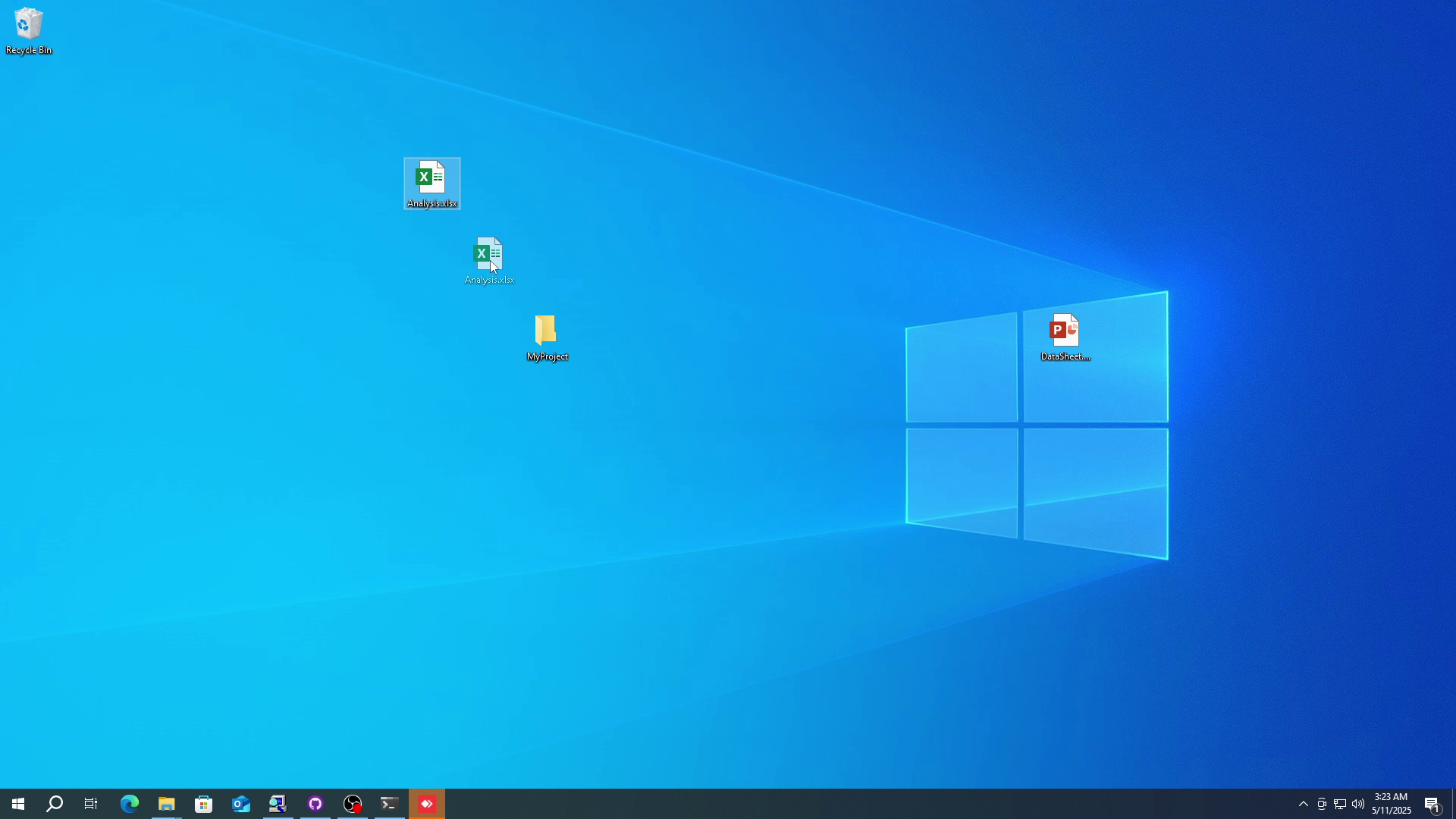} &
\includegraphics[width=\linewidth]{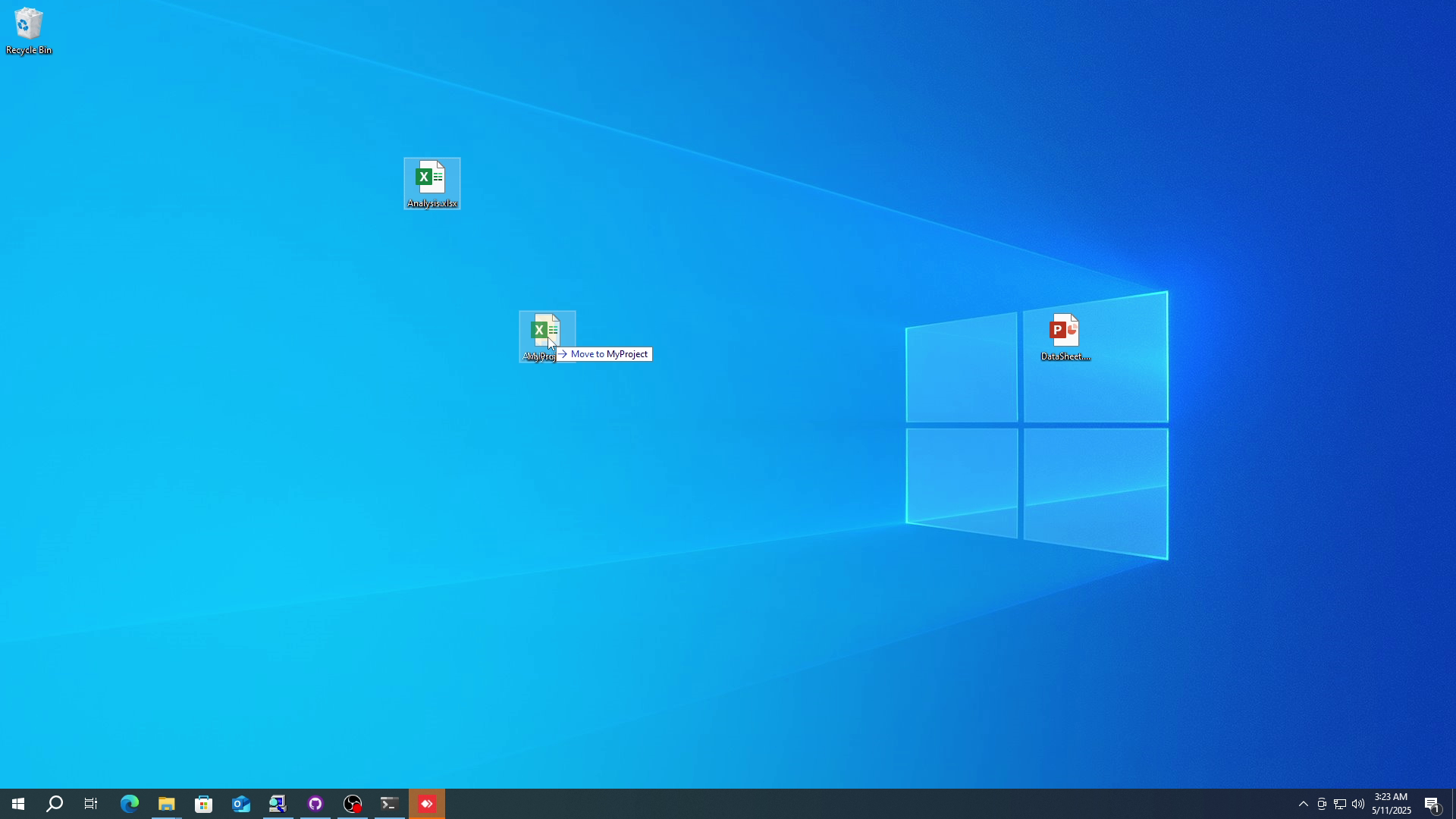} \\
& Drag Q1Report to projectDocs &
\includegraphics[width=\linewidth]{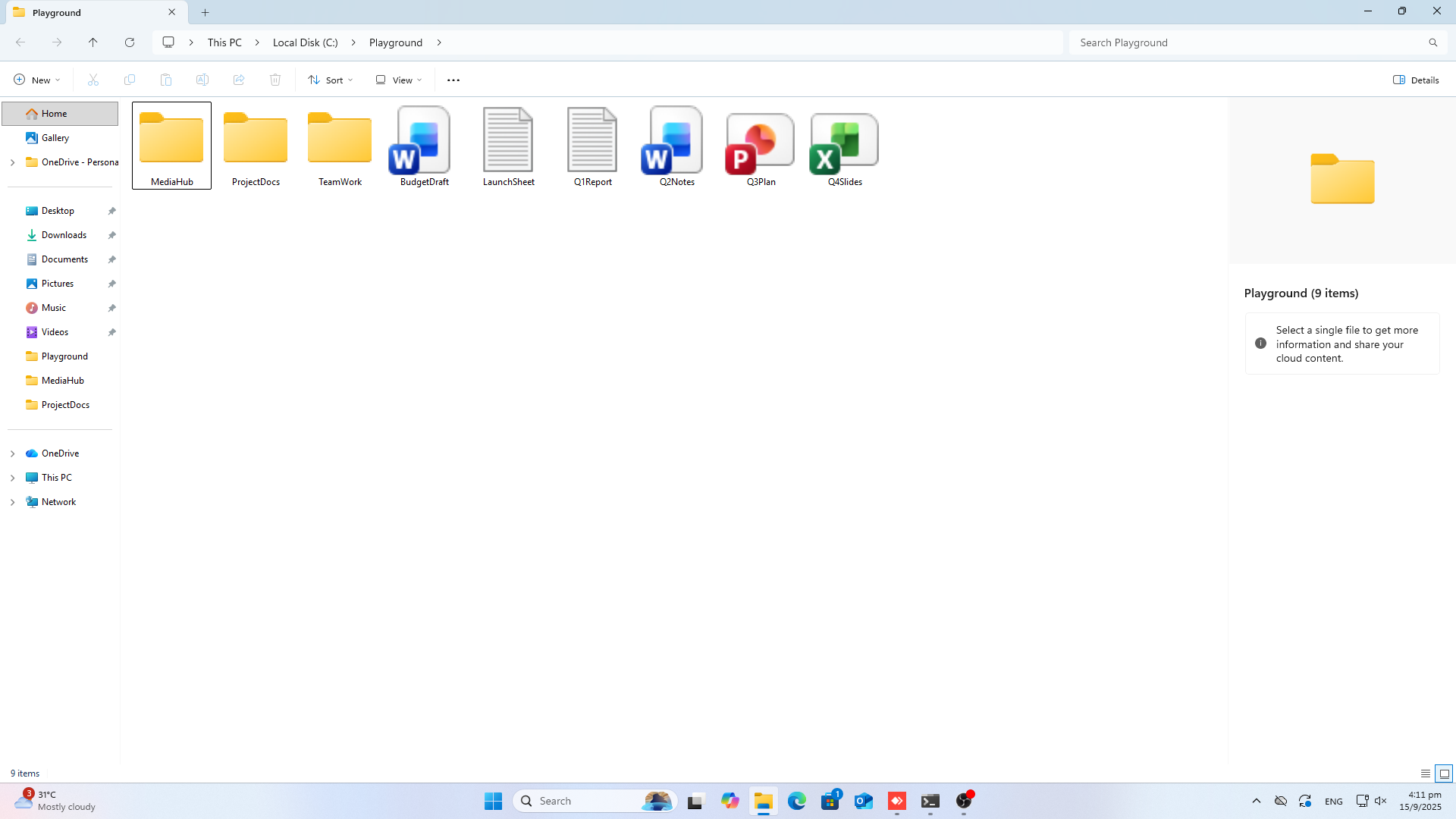} &
\includegraphics[width=\linewidth]{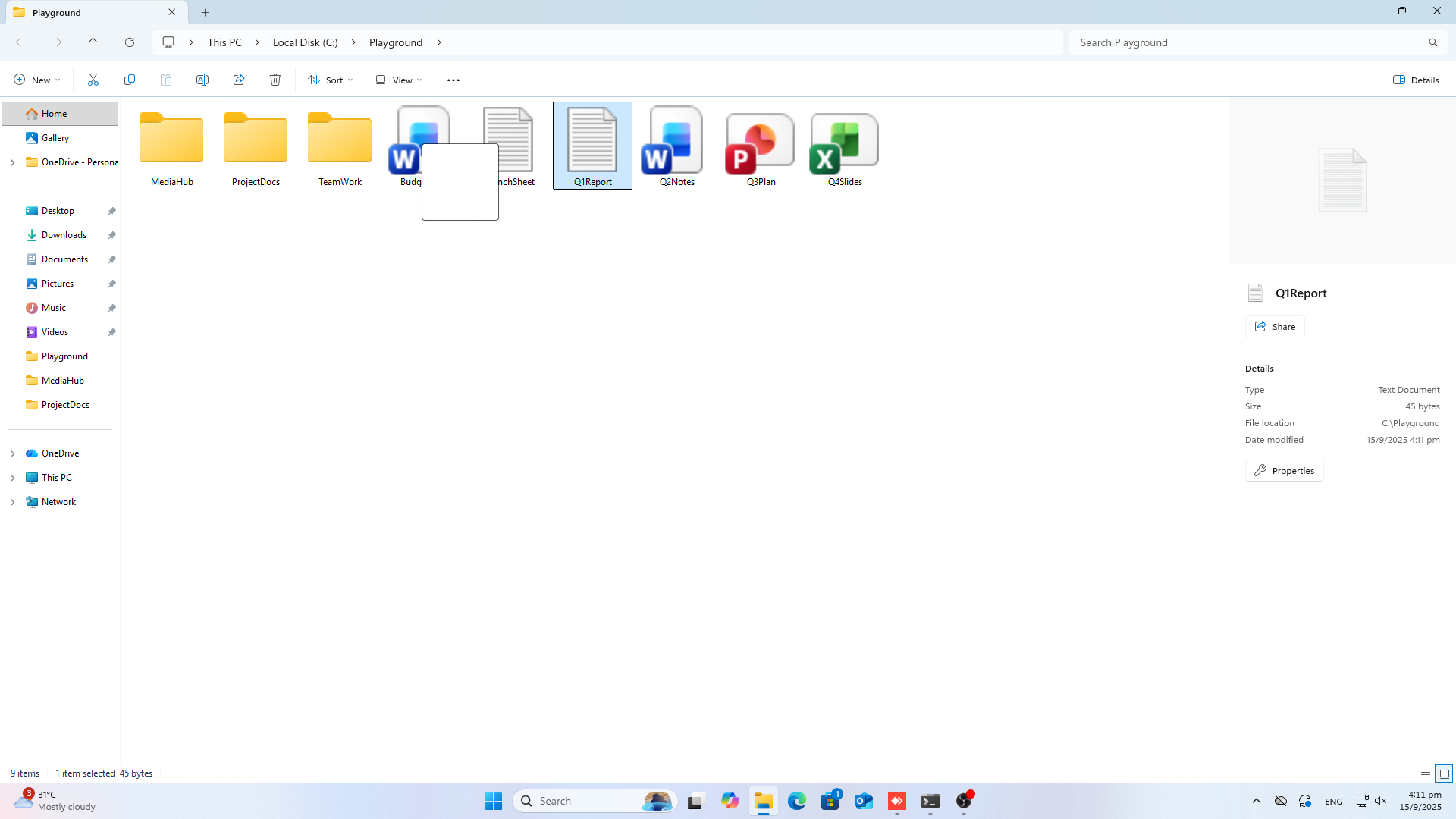} &
\includegraphics[width=\linewidth]{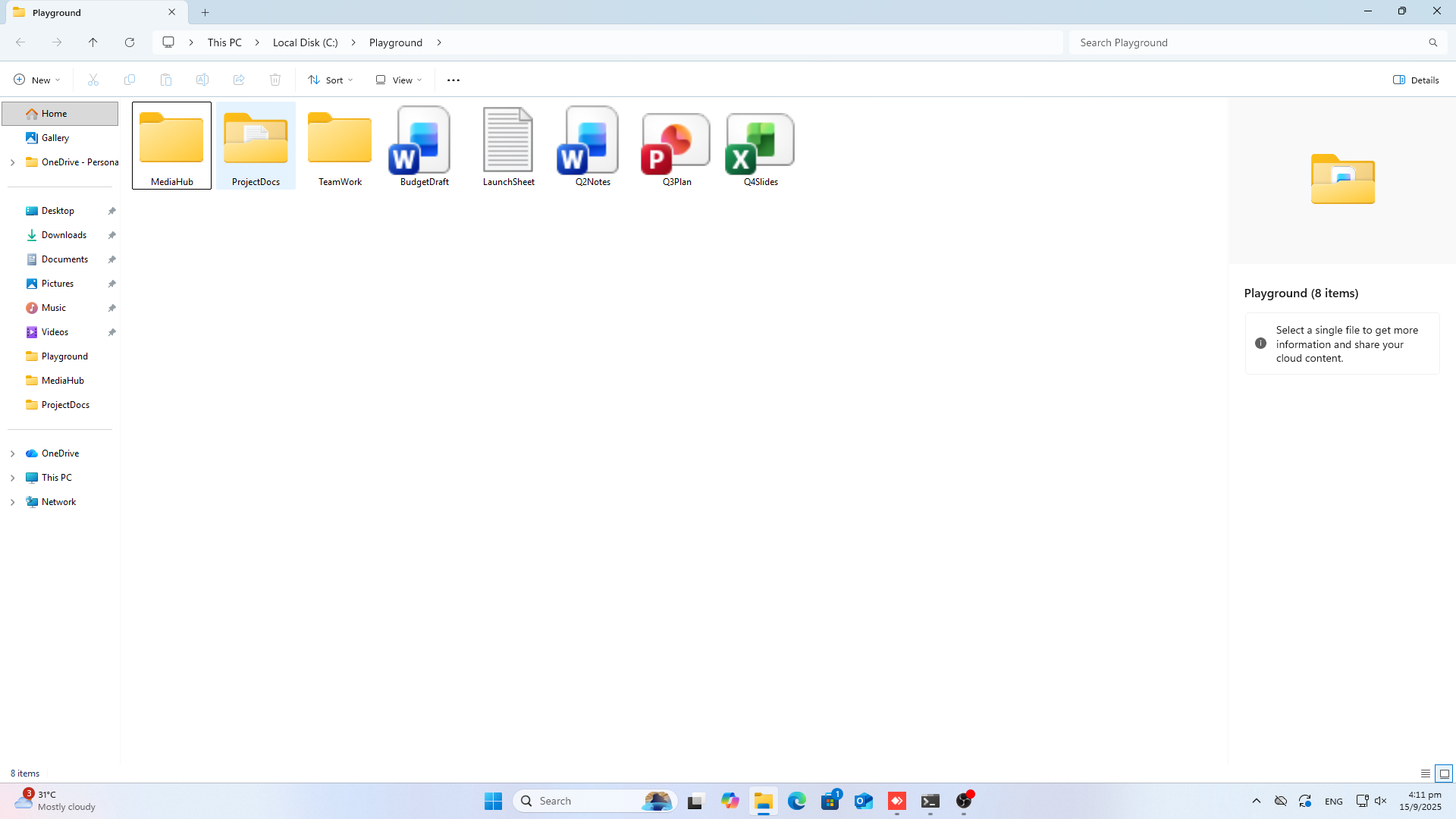} \\
\midrule
%------------------ Handwriting (1 row, REMOVED multirow) ------------------
Handwriting &
Write ``Hello World'' on canvas &
\includegraphics[width=\linewidth]{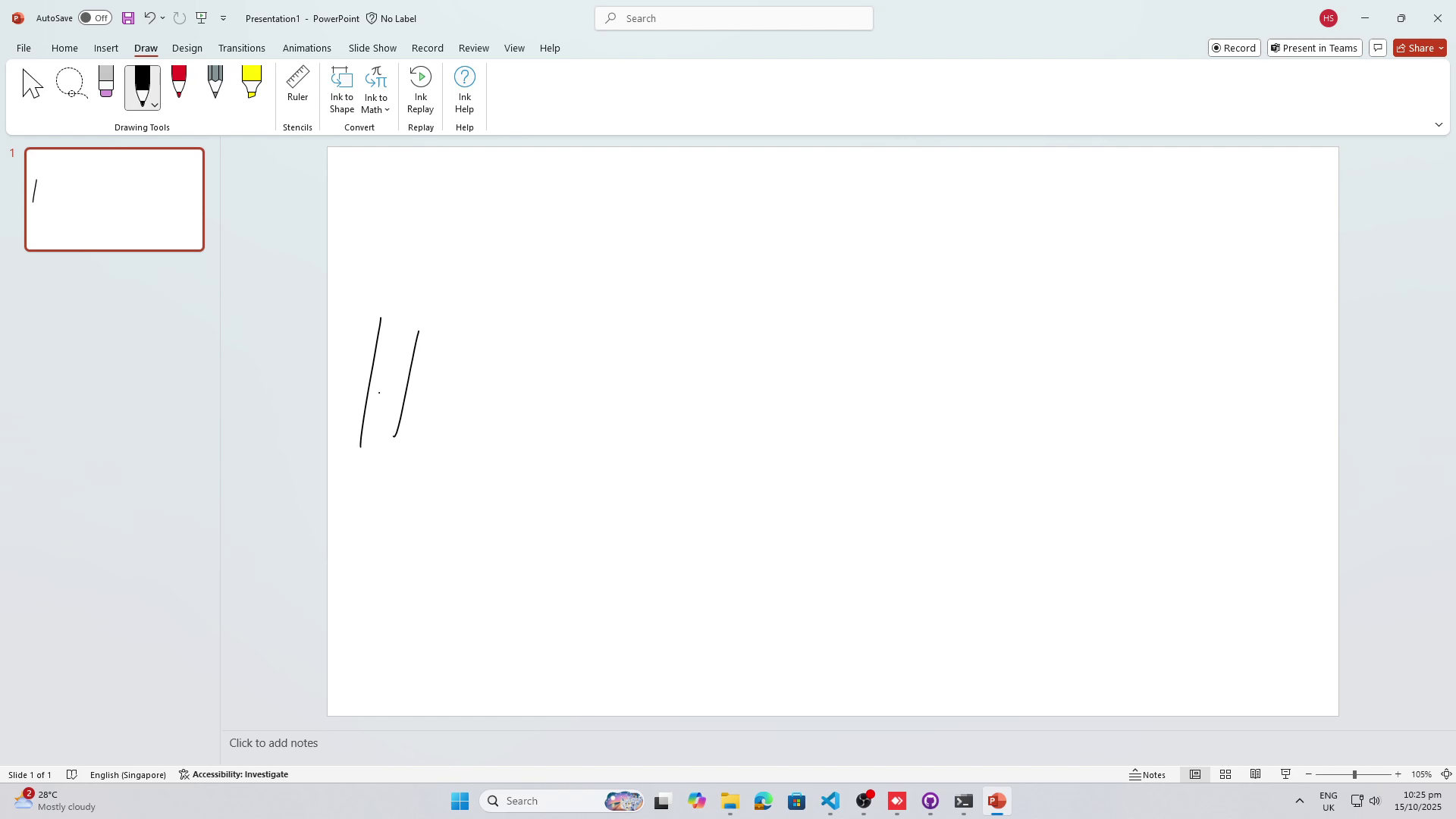} &
\includegraphics[width=\linewidth]{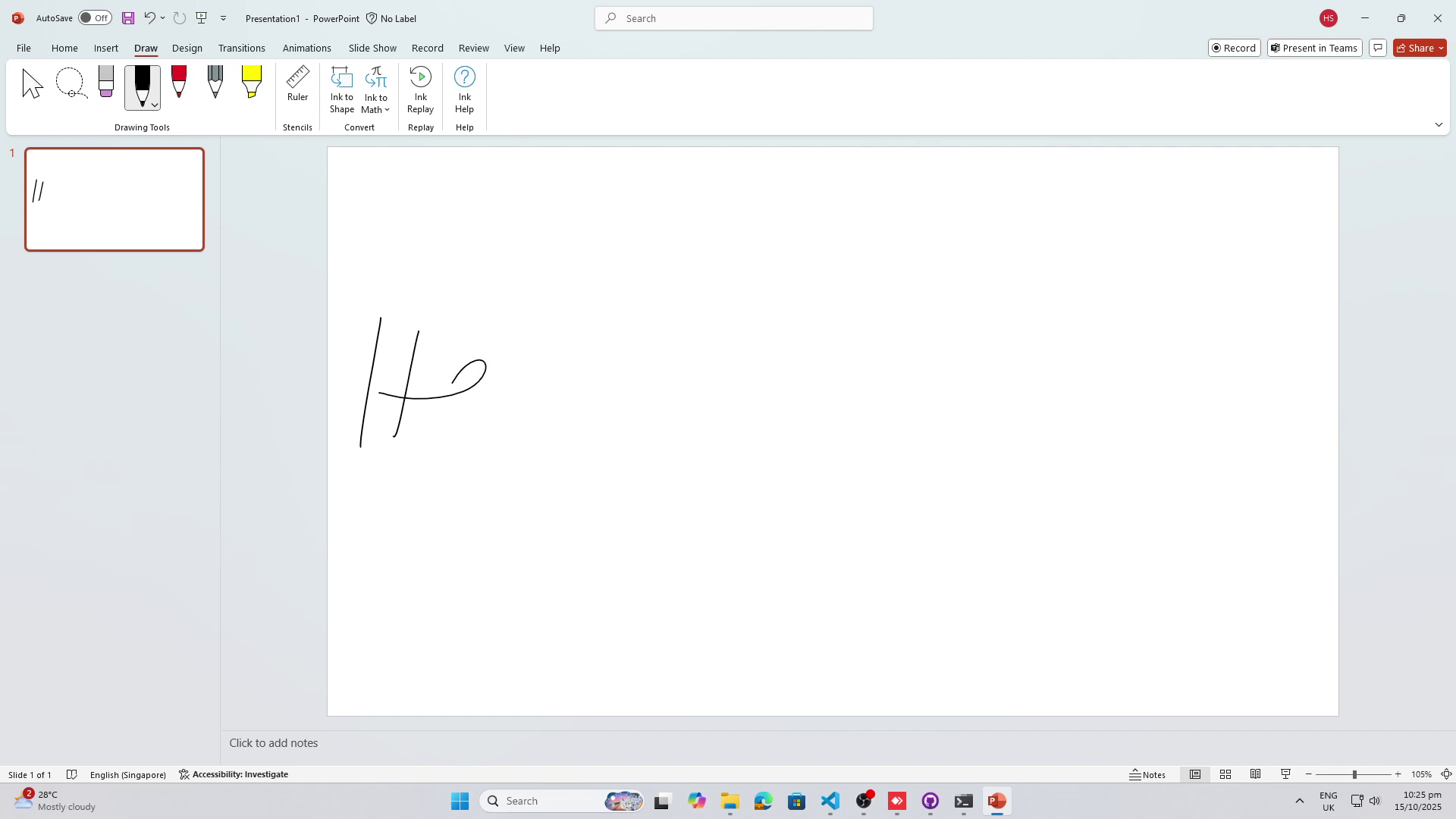} &
\includegraphics[width=\linewidth]{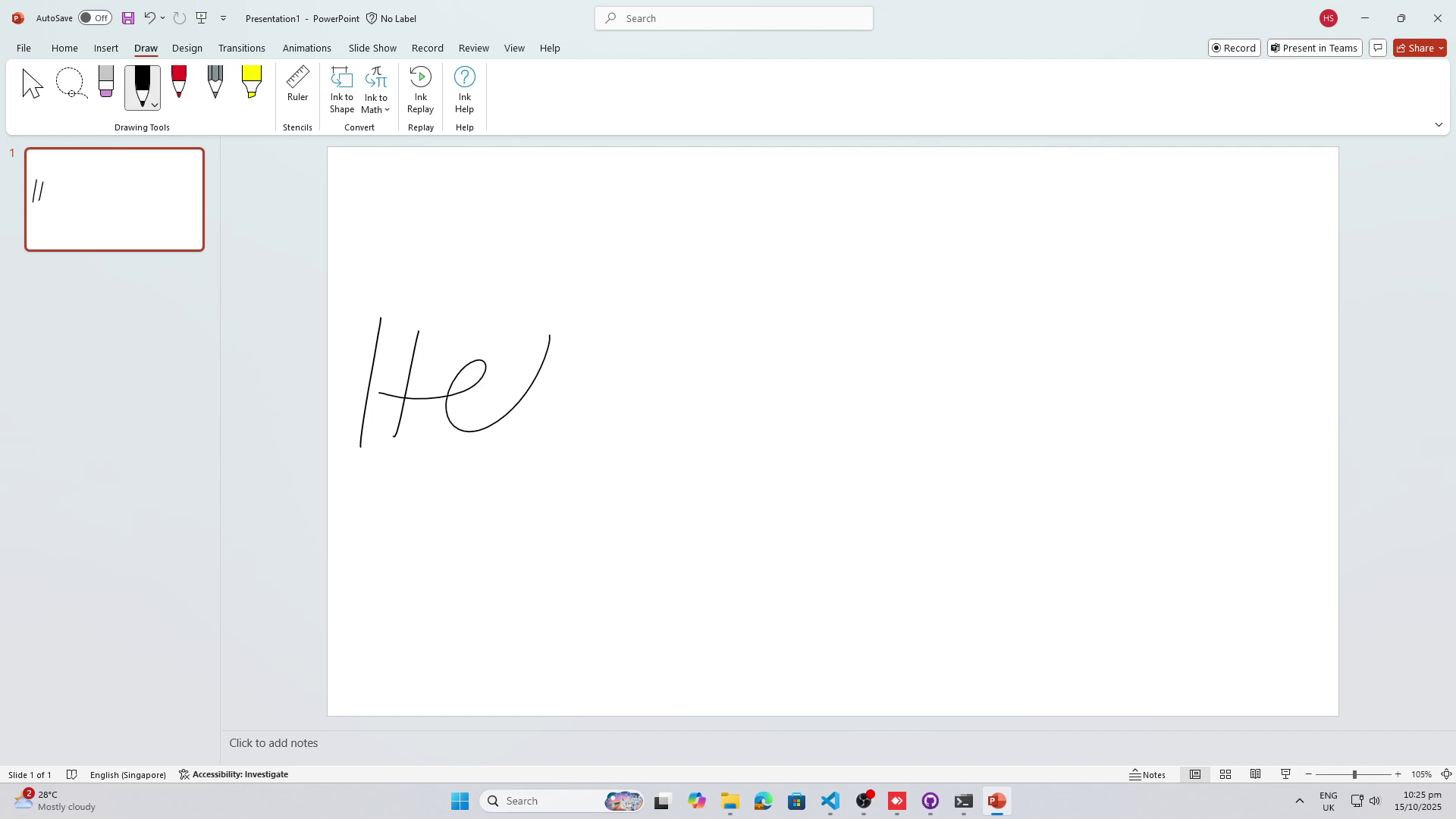} \\
\midrule
%------------------ PowerPoint (3 rows, adjusted multirow count to ~14) ------------------
\multirow{9}{*}{PowerPoint} &
Rotate the Lion counter-clockwise by 45 degrees &
\includegraphics[width=\linewidth]{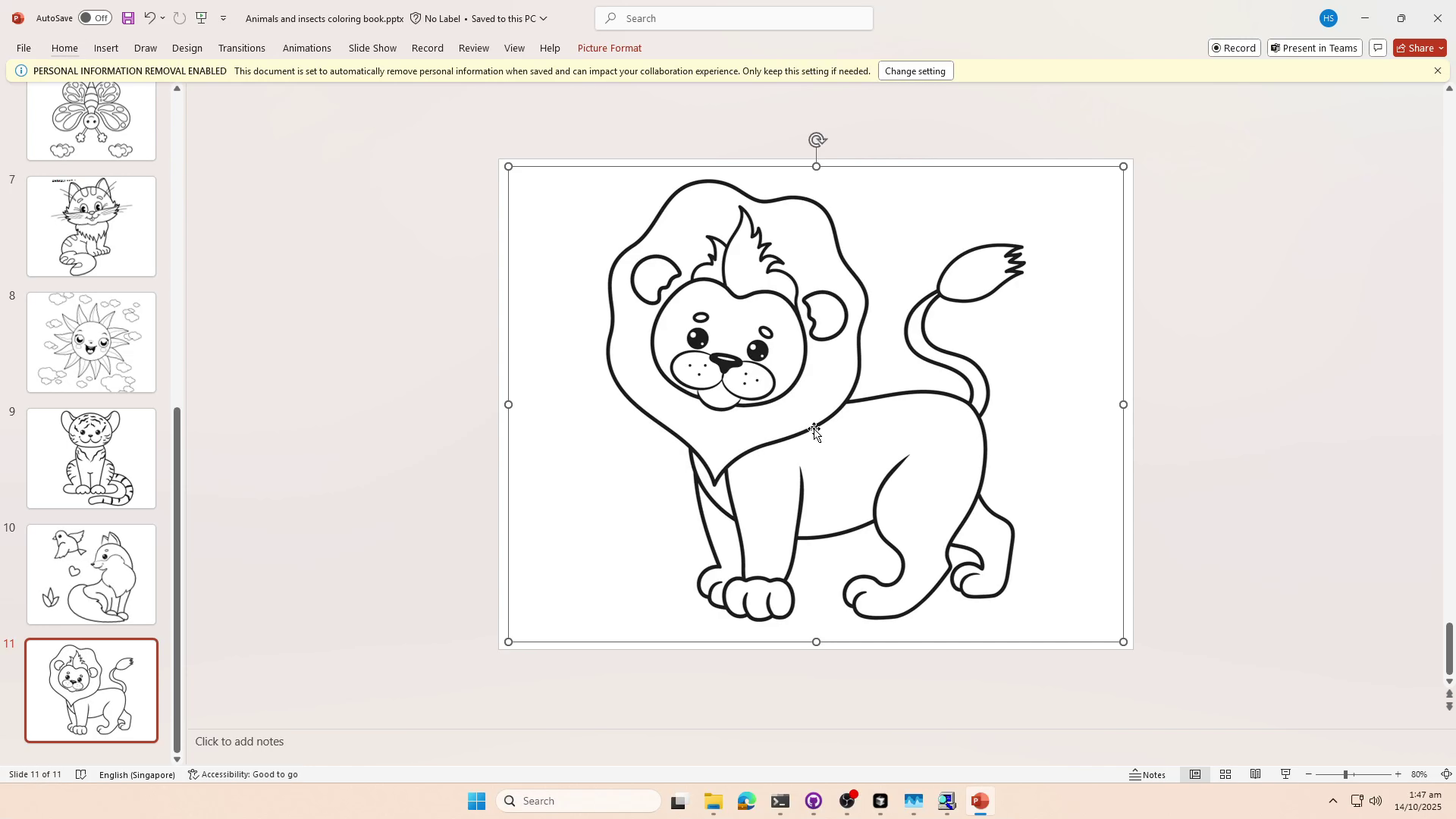} &
\includegraphics[width=\linewidth]{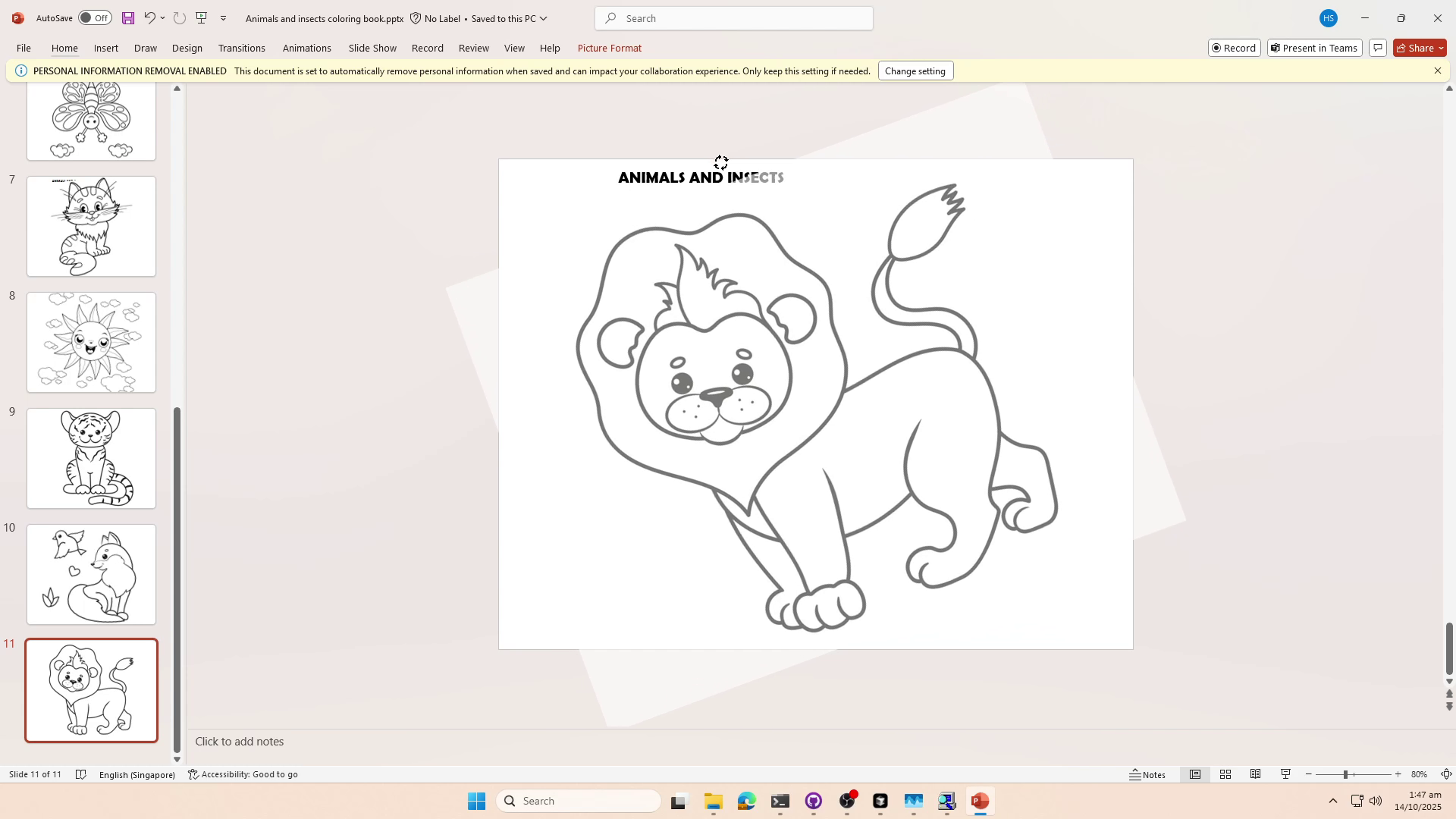} &
\includegraphics[width=\linewidth]{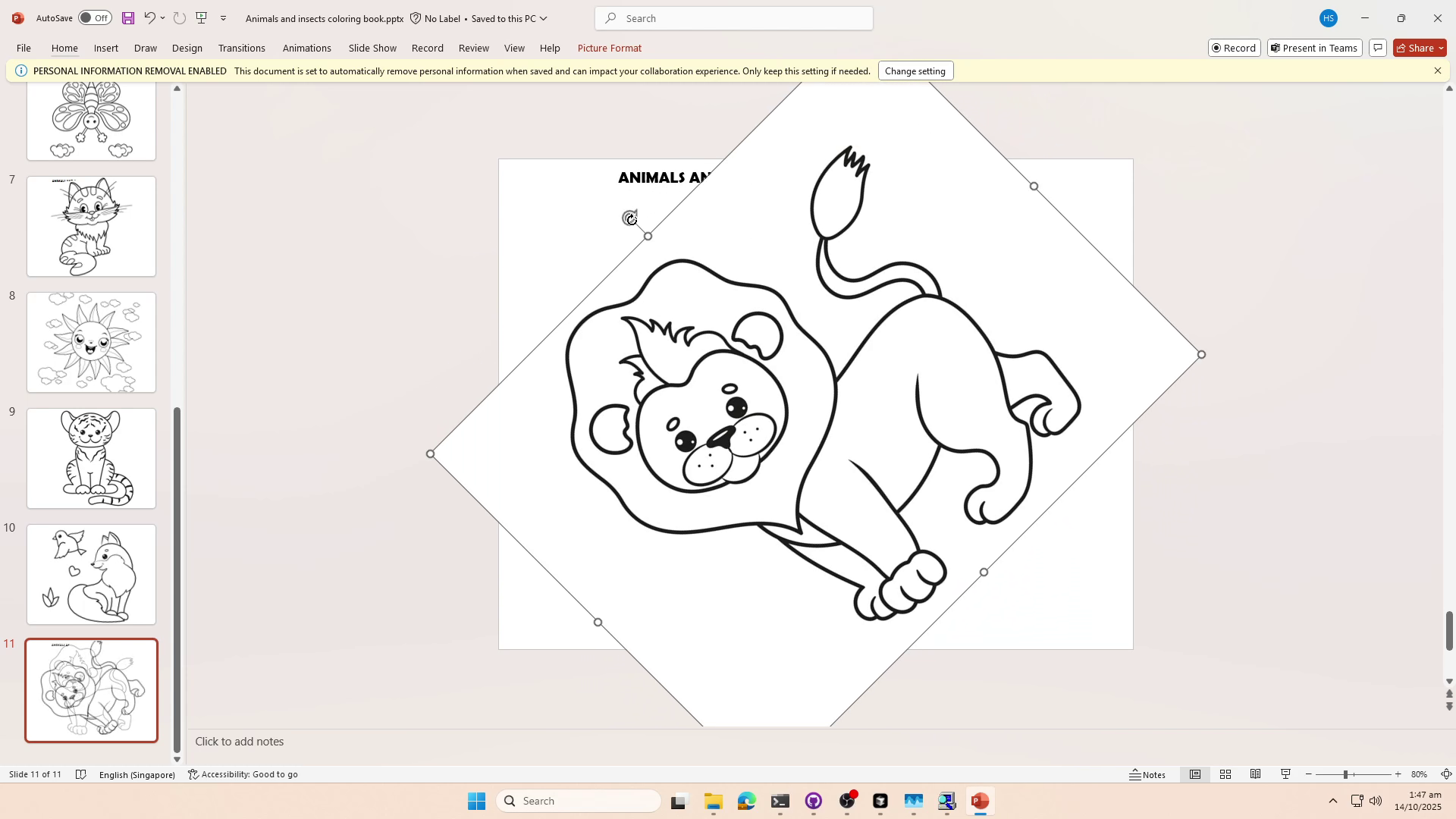} \\
& Resize the title Basic Presentation diagonally by 0.5 from top-left corner &
\includegraphics[width=\linewidth]{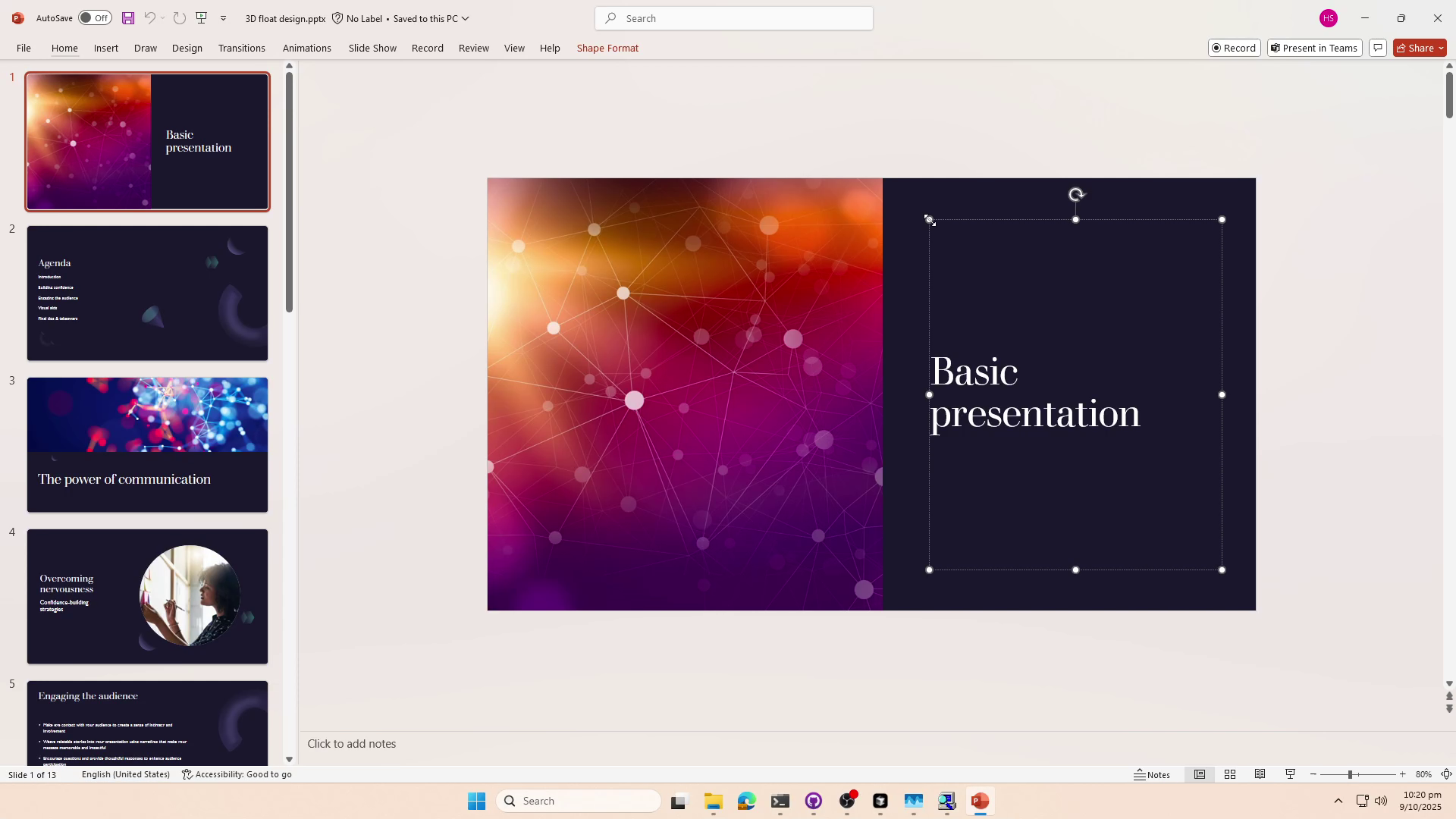} &
\includegraphics[width=\linewidth]{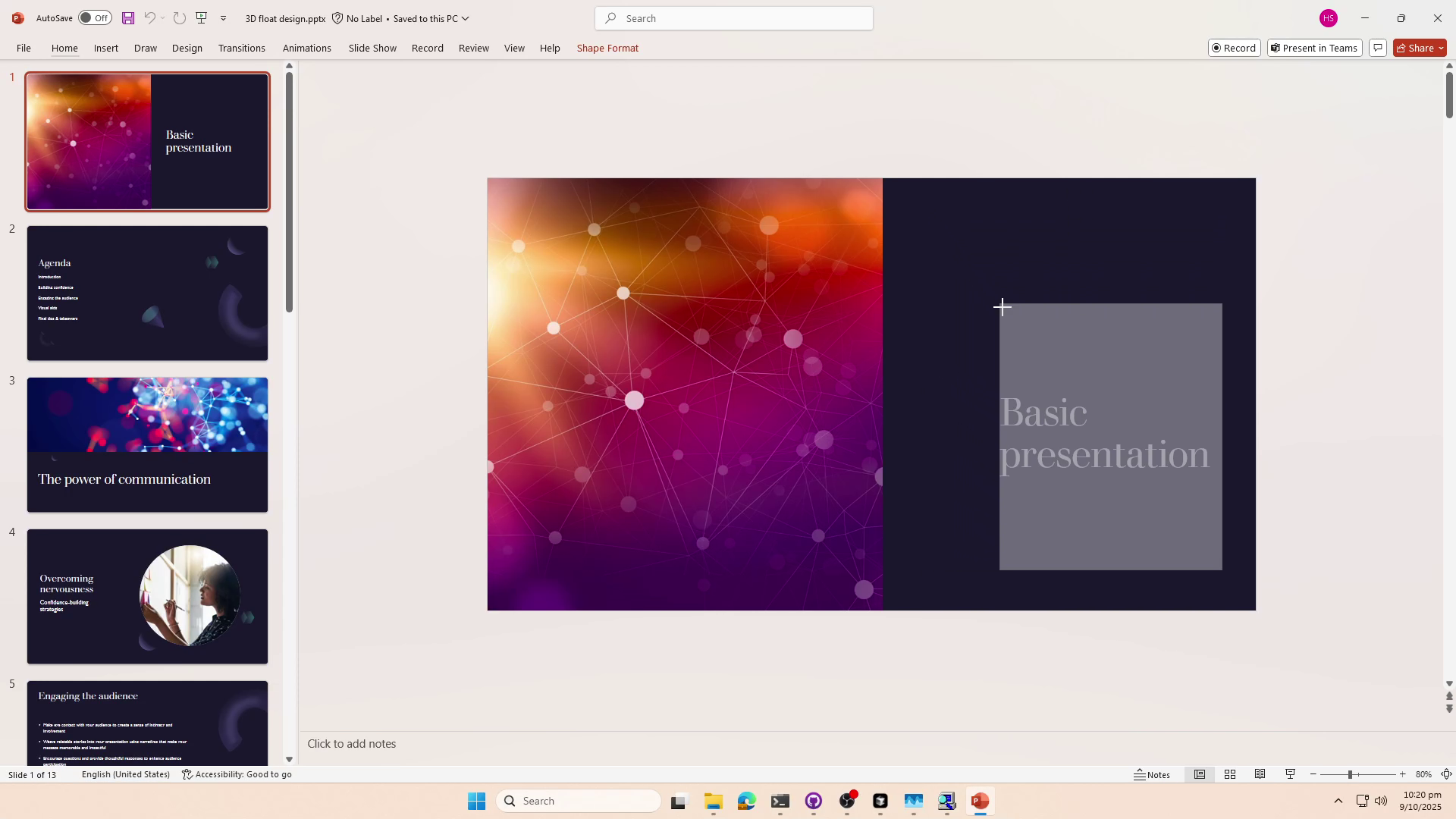} &
\includegraphics[width=\linewidth]{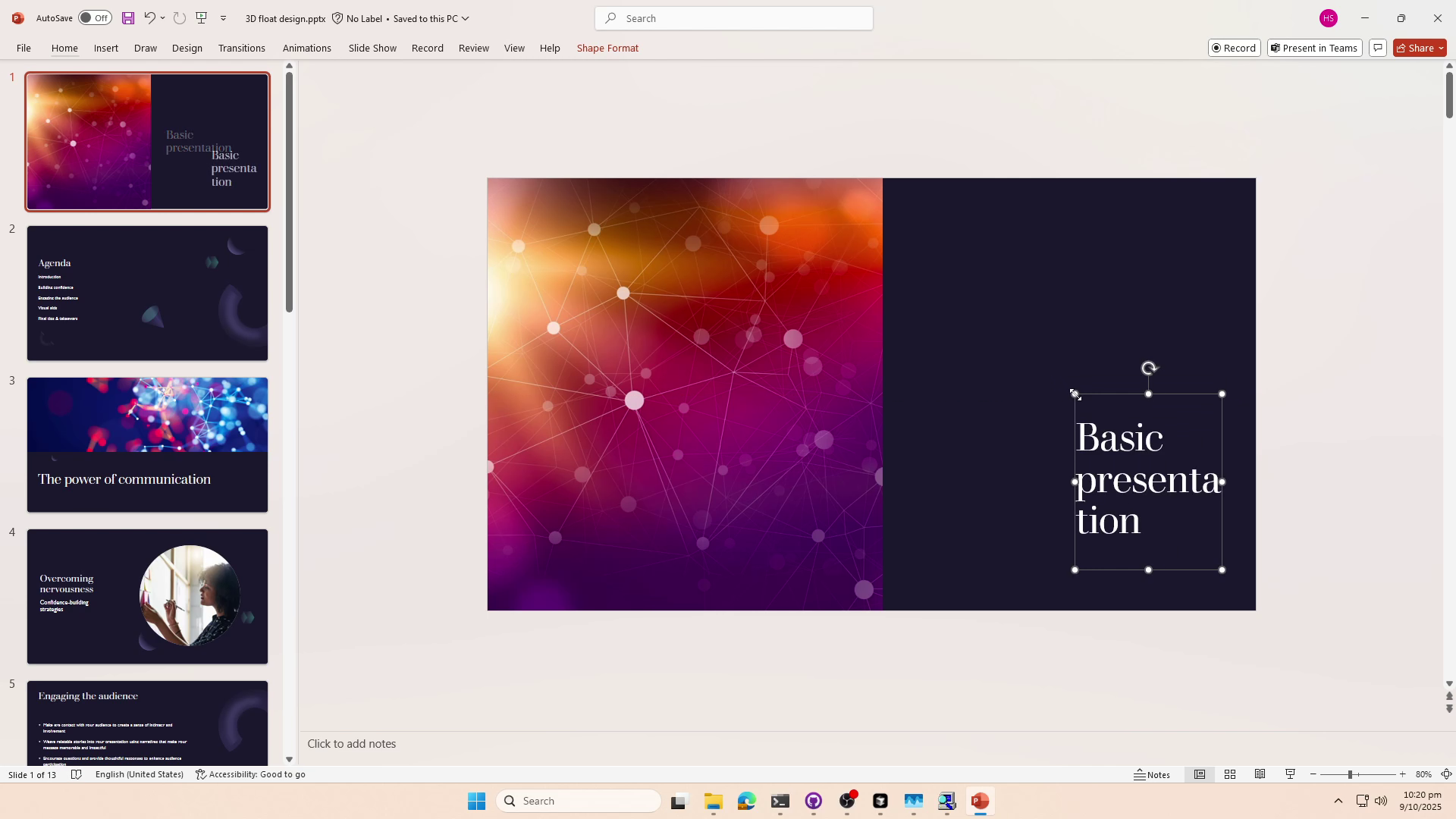} \\
& Resize width of the textbox December to 0.2 from its right &
\includegraphics[width=\linewidth]{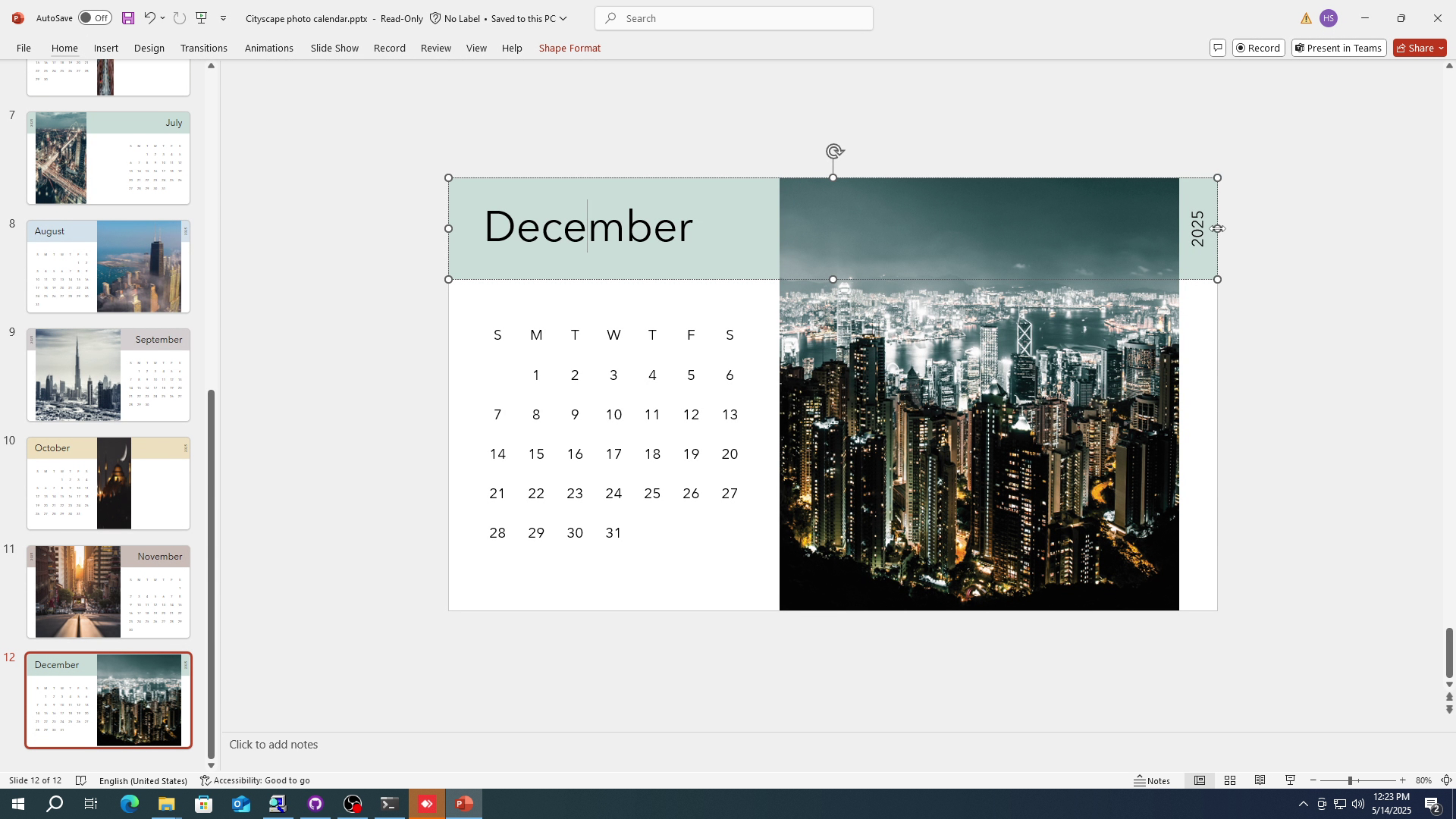} &
\includegraphics[width=\linewidth]{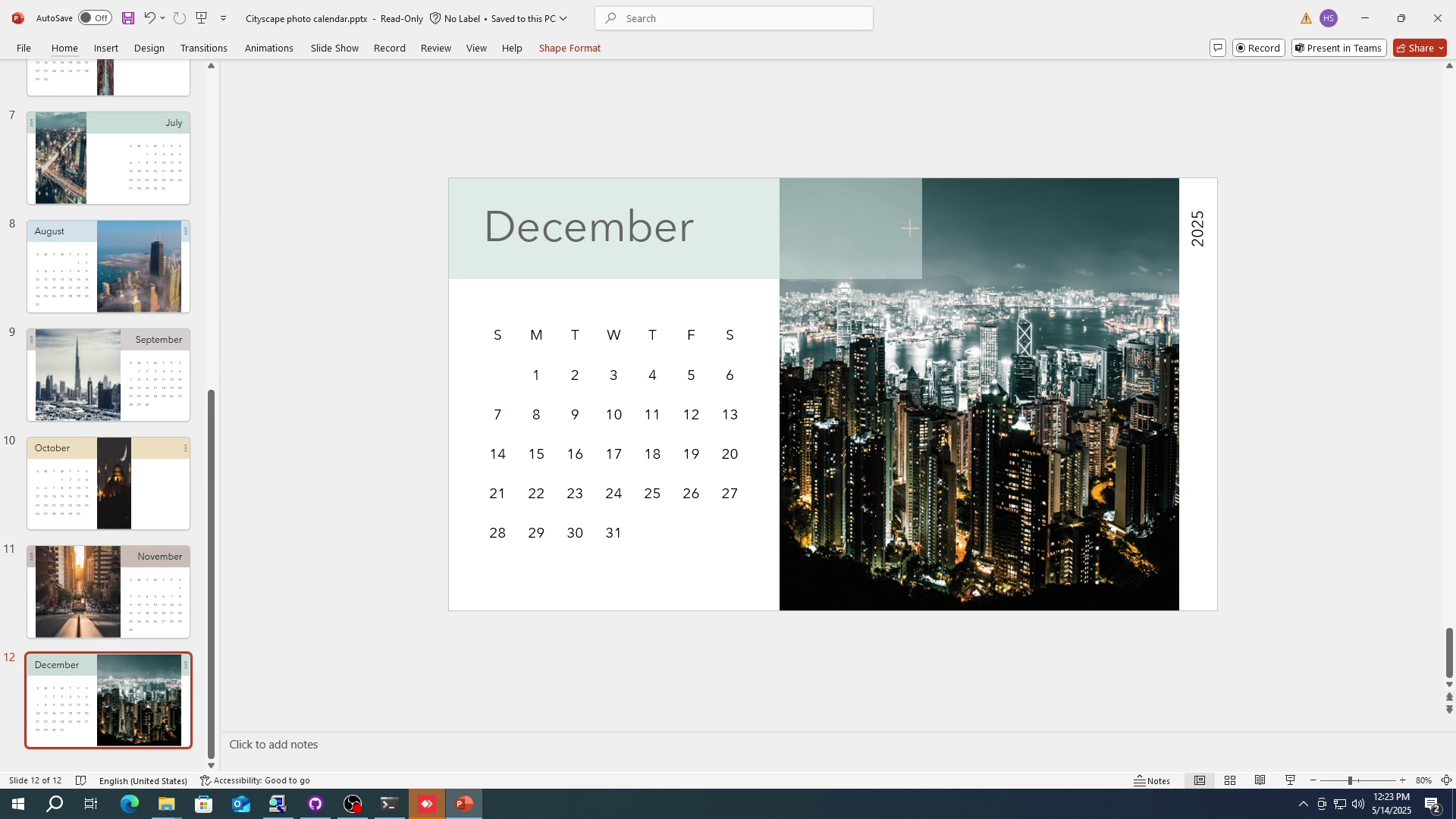} &
\includegraphics[width=\linewidth]{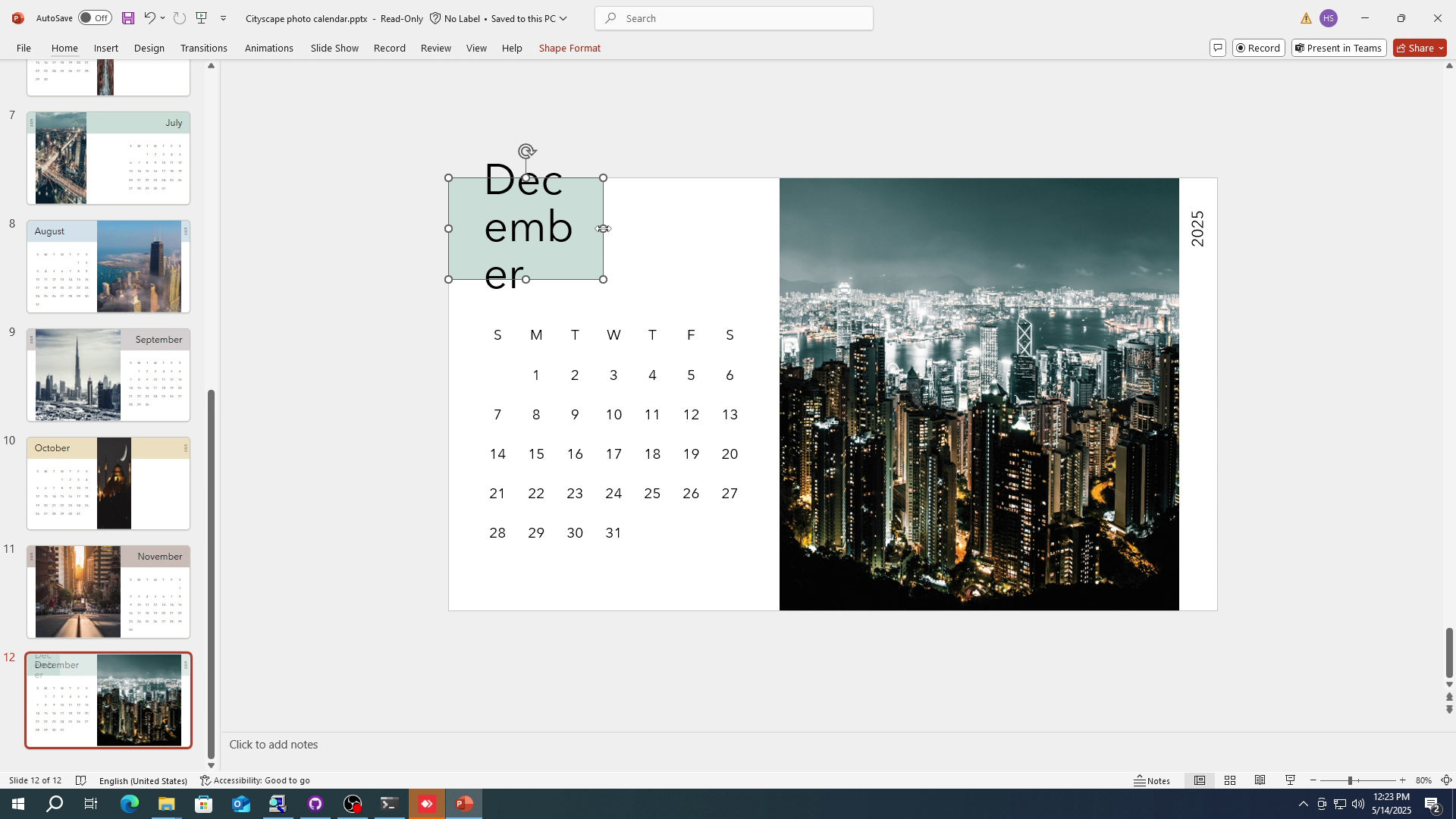} \\
\midrule
%------------------ Premiere (1 row, REMOVED multirow) ------------------
Premiere &
Apply Vertical Flip effect to Great Forest clip &
\includegraphics[width=\linewidth]{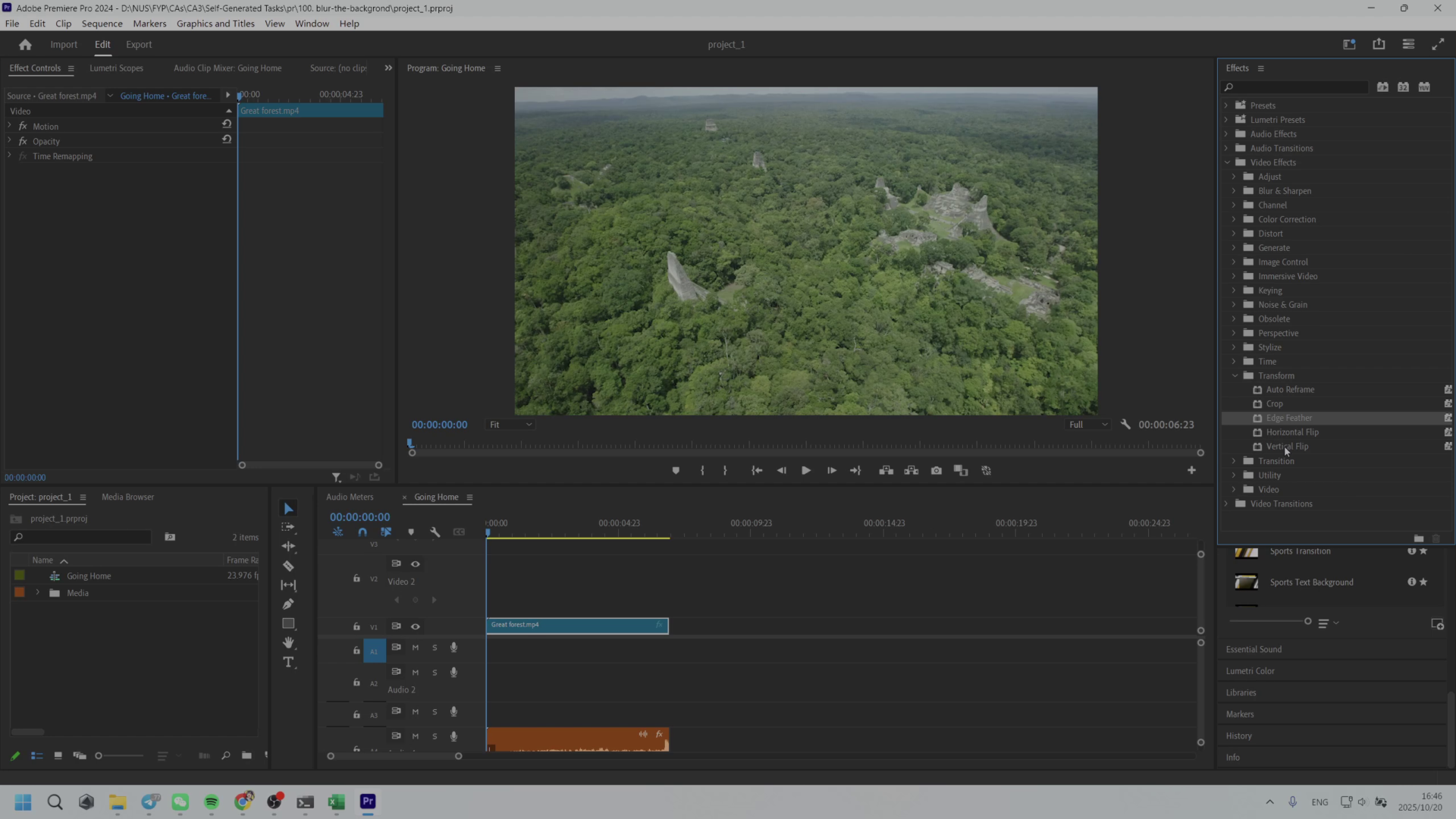} &
\includegraphics[width=\linewidth]{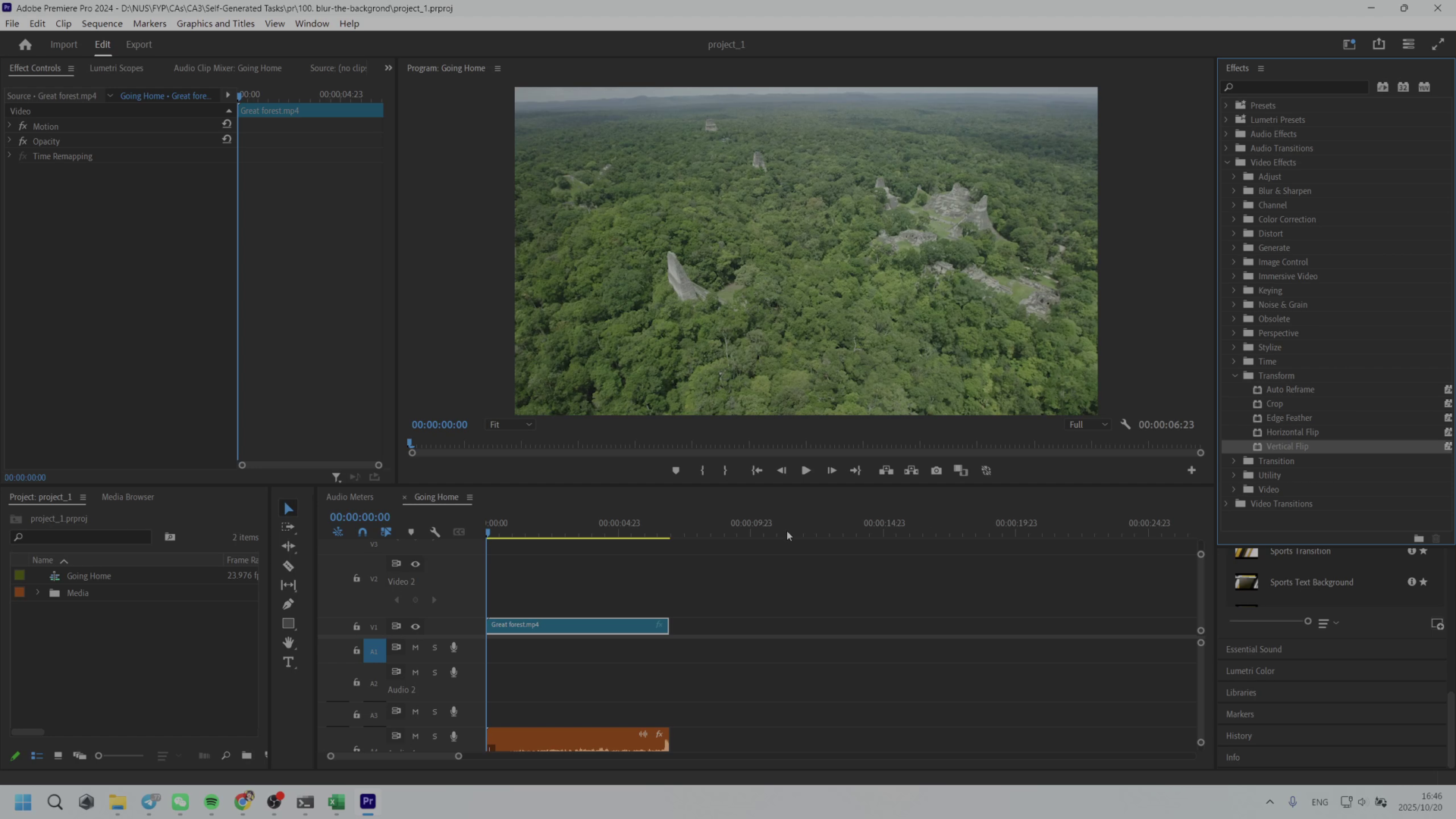} &
\includegraphics[width=\linewidth]{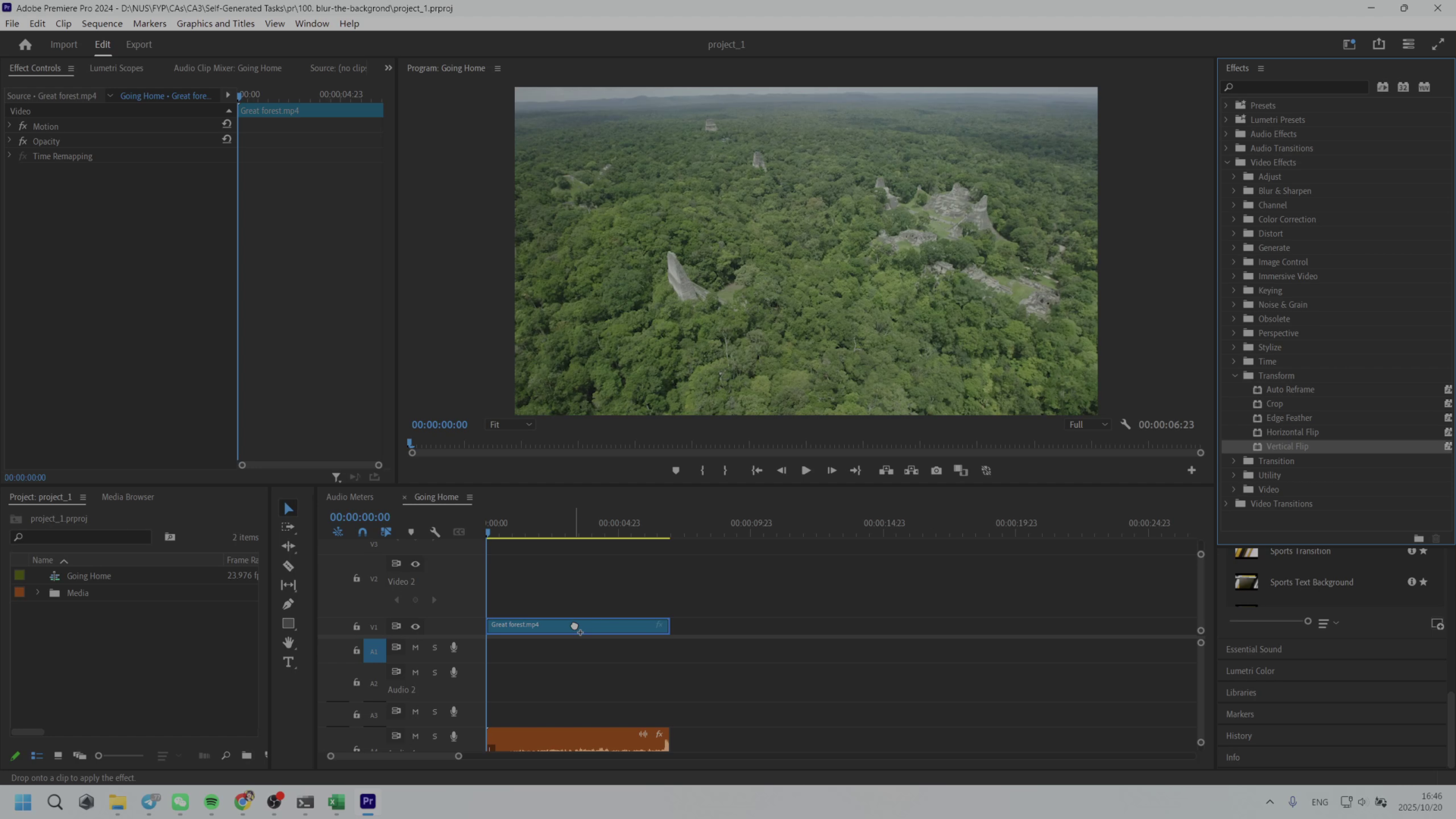} \\
\bottomrule
\end{tabular}
\end{table*}

The visualization of some task trajectories are shown in Tab.~\ref{tab:data_observations}

\subsection{Data-driven Closed-loop Online Evaluation}

\begin{figure*}[ht]
    \centering
    % 第一个图
    \begin{minipage}[b]{0.32\textwidth}
        \centering
        \includegraphics[width=\textwidth]{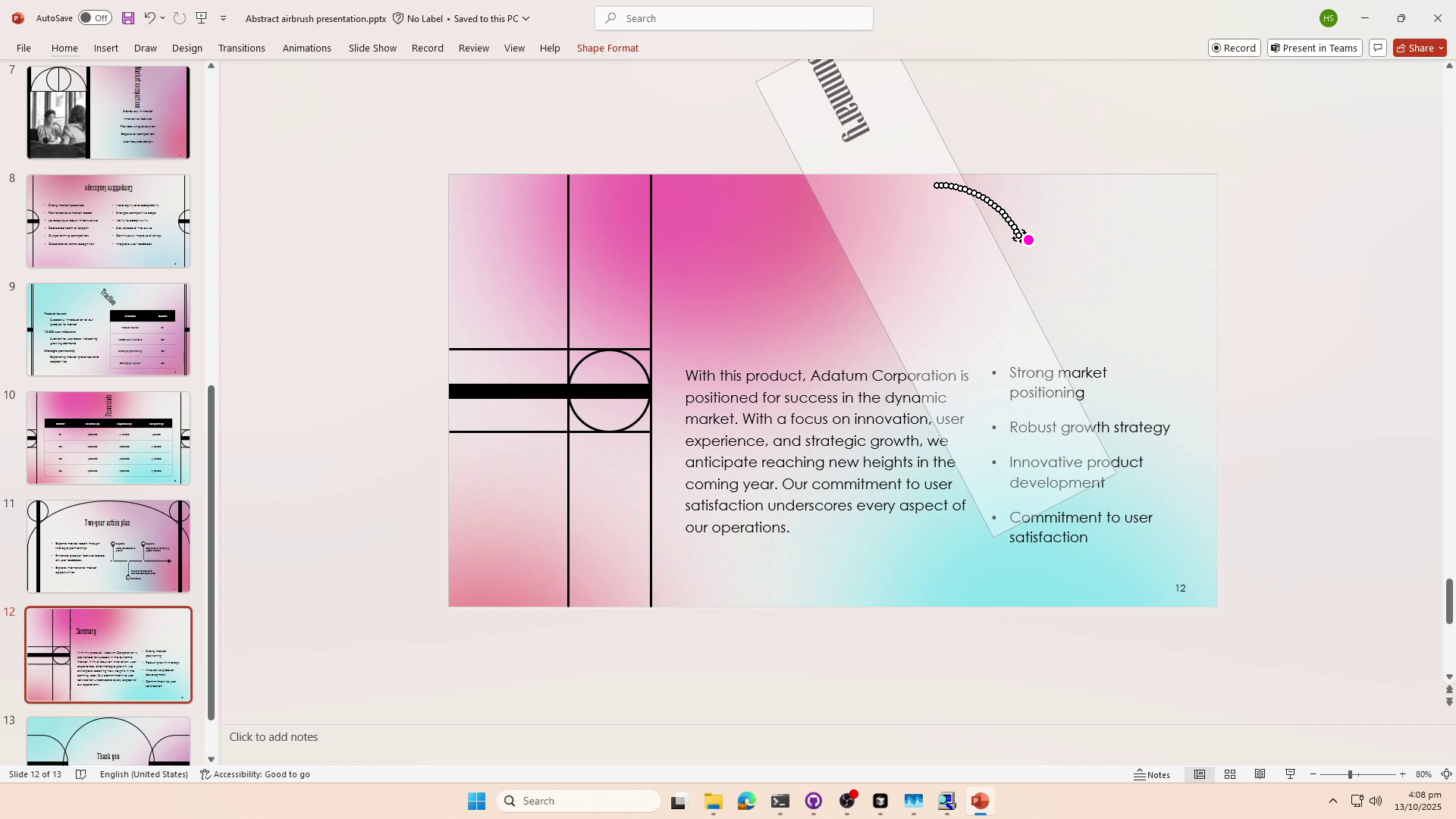}
        \caption*{(a) The model predicts a coordinate close to the dense trajectory points.}
    \end{minipage}
    \hfill
    % 第二个图
    \begin{minipage}[b]{0.32\textwidth}
        \centering
        \includegraphics[width=\textwidth]{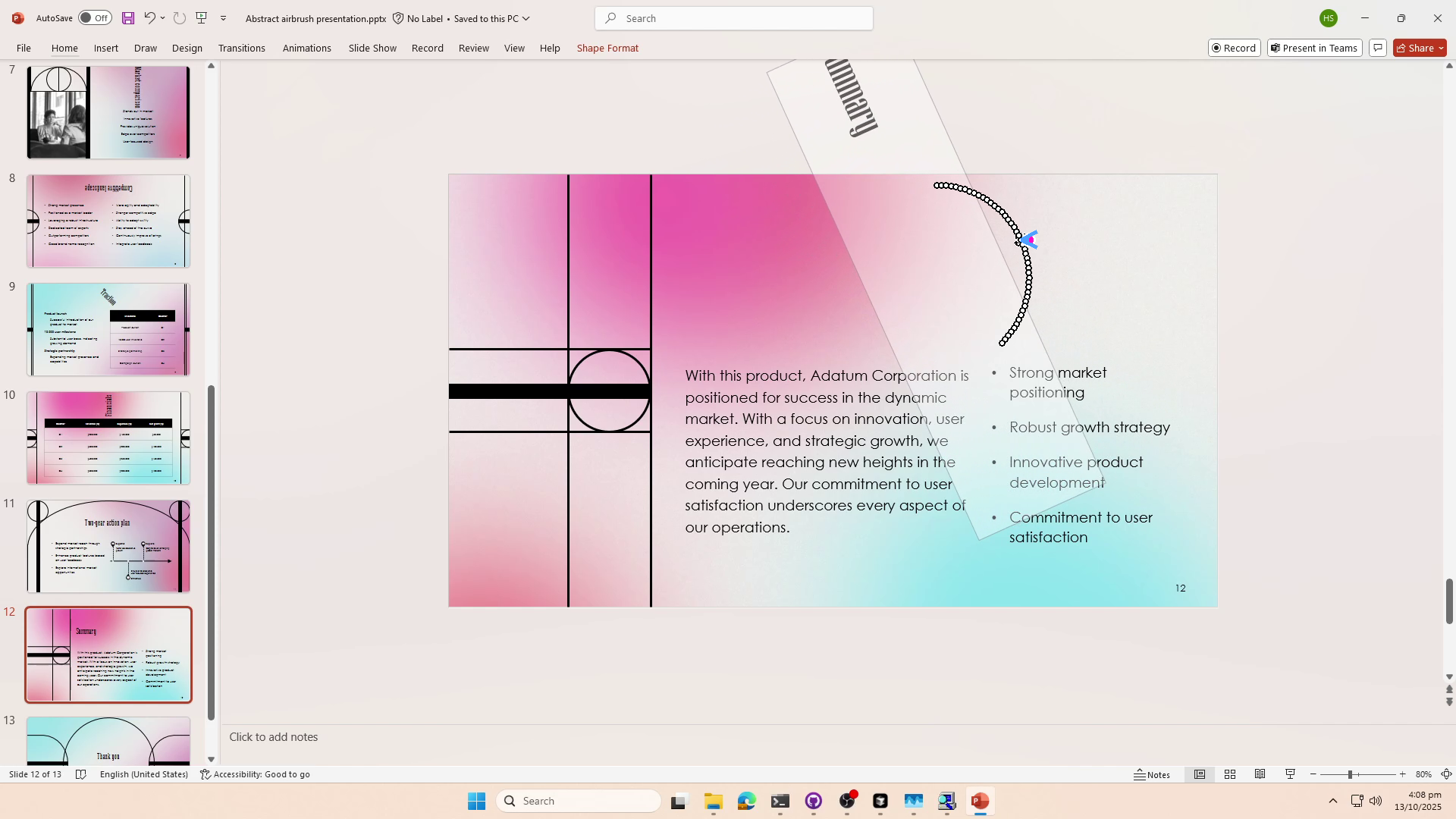}
        \caption*{(b) The prediction is mapped to its closest trajectory point.}
    \end{minipage}
    \hfill
    % 第三个图
    \begin{minipage}[b]{0.32\textwidth}
        \centering
        \includegraphics[width=\textwidth]{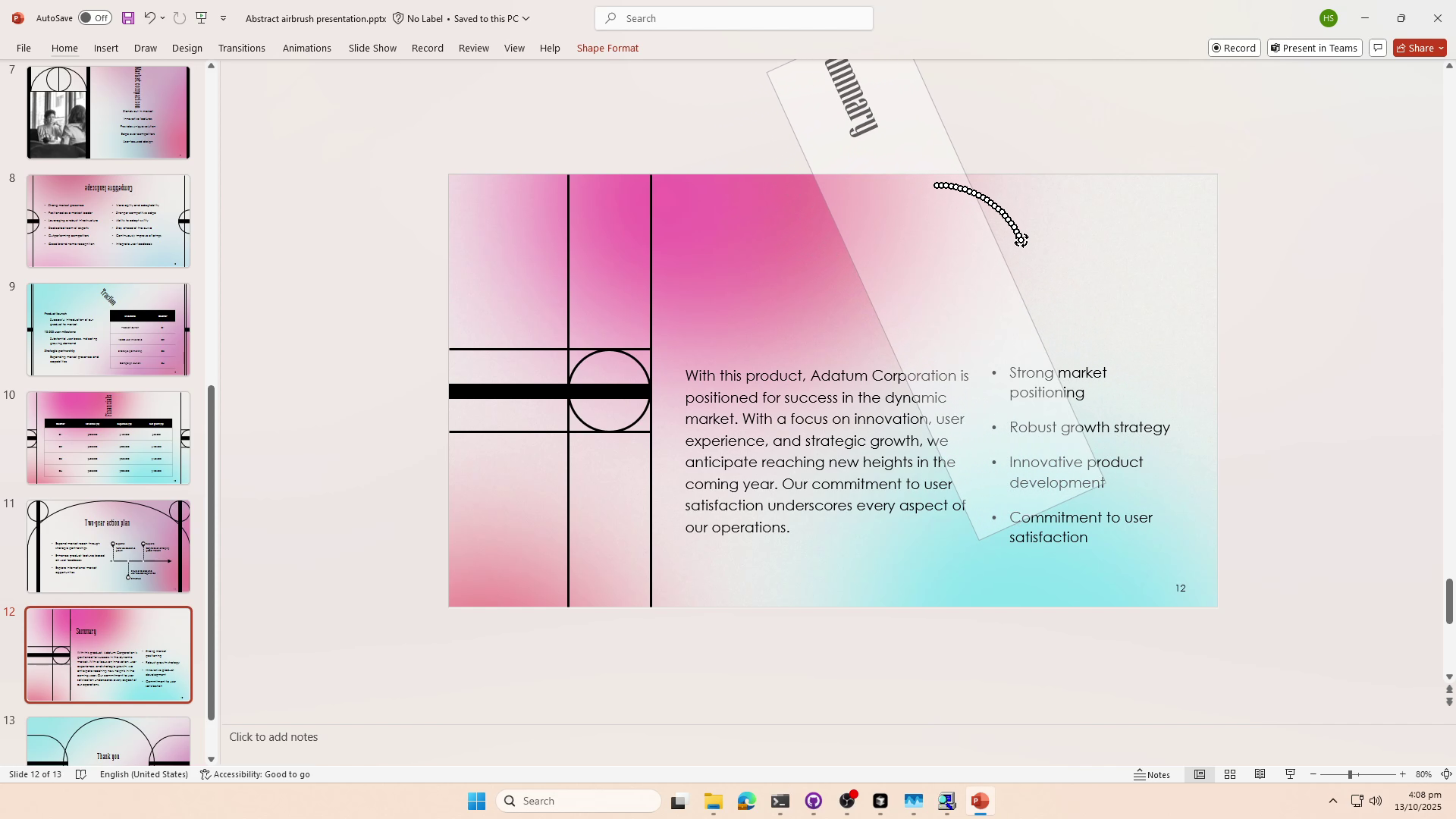}
        \caption*{(c) The model receives the next observation at the mapped trajectory point.}
    \end{minipage}

    \caption{\textbf{The visualization of the data-driven closed-loop online evaluation process.} This approach enables models to perform continuous actions with observations on-the-fly, without the complexity to set up OS and software applications, enhancing reproducibility.}
    \label{fig:data-driven}
\end{figure*}

As mentioned in the main paper, a \textbf{data-driven} approach is designed to enable closed-loop rollouts in online evaluation. For each drag task, we store the video recording, task specification, and dense drag trajectory, providing extensive possible GUI states encountered during dragging. During rollouts, the model’s predicted action is matched to the nearest recorded state if it falls within a tolerance $\epsilon$ (\eg~within 20 pixels of a ground-truth waypoint), upon which the corresponding next observation is retrieved. As shown in Fig.~\ref{fig:data-driven}, when the model prediction midway can be mapped to a trajectory point, its corresponding UI state will be retrieved from the extensive pre-collected UI states as the next observation, thus the model can perform continuous actions with observations on-the-fly. This data-driven approach largely reduces the complexity of setting up OS and software applications, enhancing reproducibility, while still enabling a closed-loop rollout manner.
\section{Failure Cases of Baseline Models}
To better illustrate the behavior of baseline models in drag tasks, we analyze eight failure cases observed during evaluation. These qualitative examples highlight problems encountered by existing GUI agents based on language-modeling.

\begin{figure}[h]
\centering
\includegraphics[width=\linewidth]{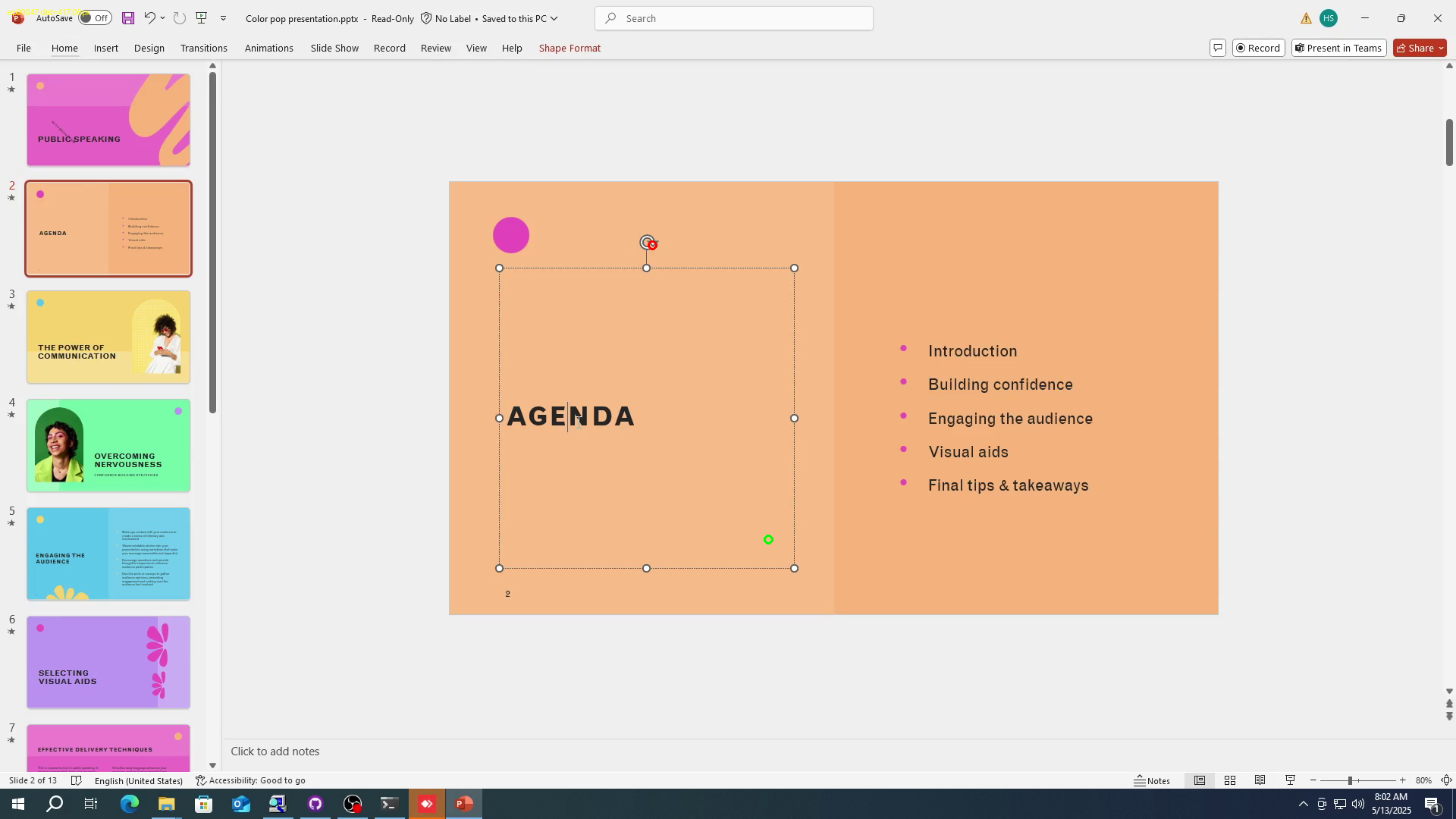}
\caption{\textbf{Know-How but does not have the tool.} The baseline formulates a correct plan to rotate the textbox by dragging the handle above the textbox "AGENDA" with an arc trajectory, however, the baseline is not equipped with such a drag tool, it is only trained and equipped with linear drags, thus failing the task.}
\label{fig:case_study:no_tool}
\end{figure}

% \noindent\textbf{(i) Know-How but does not have the tool: OpenCUA-7B and Operator.}
\noindent\textbf{(i) Know-How but does not have the tool.}
\label{case1}
% Task: Drag the file icon into the folder.
% Source Image: aiassist_desktop/.../screenshot_0_initial.png
As shown in Fig.~\ref{fig:case_study:no_tool}, the PowerPoint rotation task expects the model to rotate elements using the white rotation handle on the target elements with an arc trajectory. Interestingly, both OpenCUA-7B and Operator frequently produce the correct first step, successfully locating the handle.
However, they are only equipped and trained with linear drag tools,~\eg, \texttt{Drag(($x_1$,$y_1$),($x_2,y_2$))} and \texttt{Linear\_Drag(($x_1$,$y_1$),($x_2,y_2$))},~\etc. Therefore, they are unable to perform rotation even if they know how to rotate, as they do not have the corresponding tool.
This highlights a major limitation of GUI agents based on language-modeling: Beyond drag in an arc trajectory, there are various types of free-form drags in real-world scenarios, which are extremely difficult to be fully covered by defining drag tools.

\begin{figure}[h]
\centering
\includegraphics[width=\linewidth]{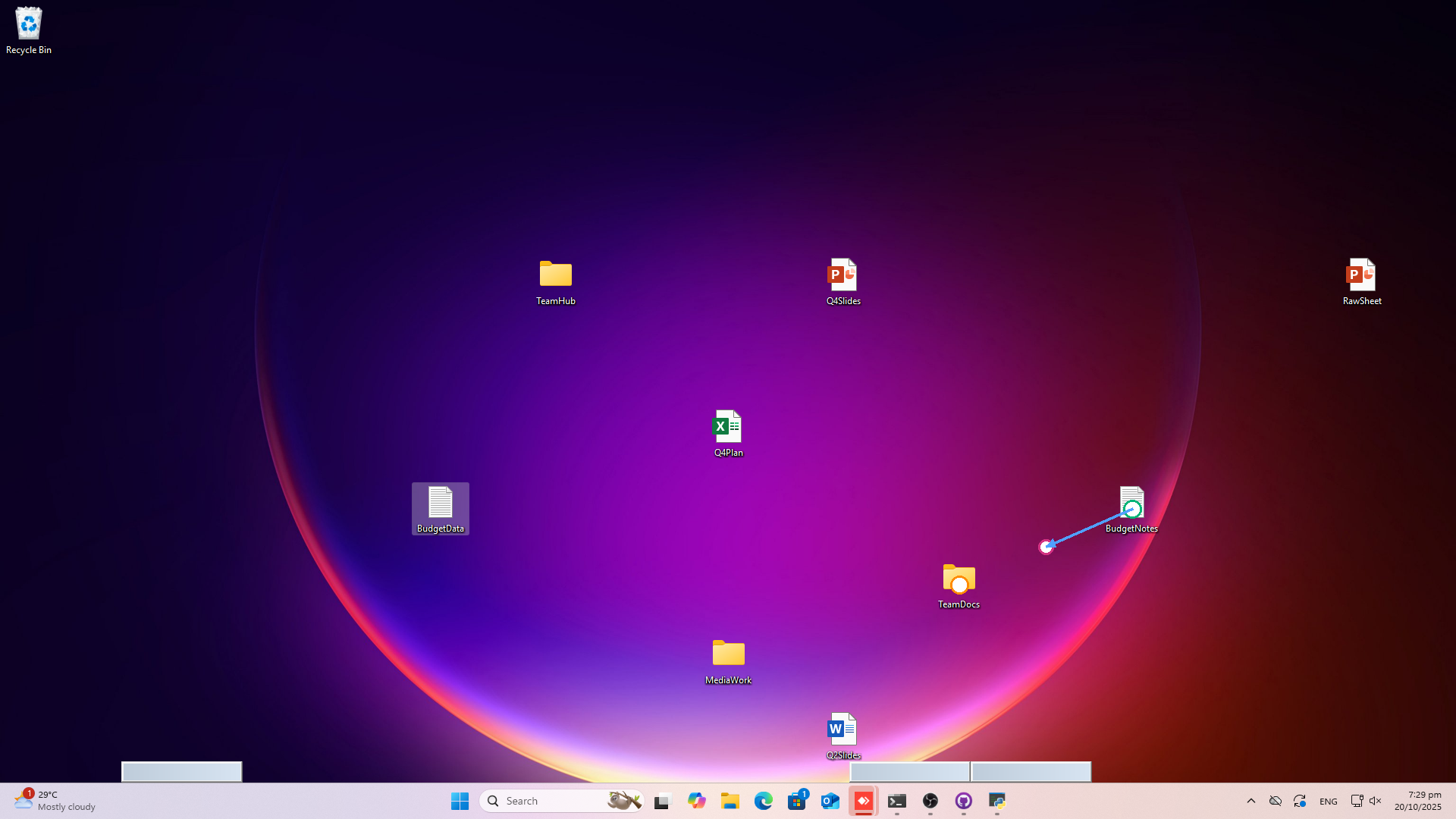}
\caption{\textbf{Gap between discrete tool use and continuous drag.} The baseline formulates a correct plan to drag the file icon to the folder and successfully locates the file icon's initial position, but the execution fails mid-trajectory, leaving the icon stranded far from the target.}
\label{fig:case_study:intent}
\end{figure}

% \noindent\textbf{(ii) Gap between discrete tool use and continuous drag: Seed and Qwen3-VL-32B.}
\noindent\textbf{(ii) Gap between discrete tool use and continuous drag.}
\label{case2}
% Task: Drag the file icon into the folder.
% Source Image: aiassist_desktop/.../screenshot_0_initial.png
As shown in Fig.~\ref{fig:case_study:intent}, on desktop drag tasks, both the Seed-1.6-Vision baseline and Qwen3-VL-32B frequently produce correct high-level plans such as ``drag \texttt{BudgetNotes.txt} into \texttt{TeamDocs}'', successfully locate the icon position, and the resulting trajectory starts in the right direction.
However, the cursor often stops mid-way or lands noticeably short of the folder, leaving a large distance to the target.
This case illustrates a gap between discrete tool use and continuous drag: language-modeling baselines are designed to predict one discrete tool call each time,~\eg, drag files using one \texttt{Drag()} call, instead of continuous actions with on-the-fly observation, thus they cannot finely adjust the cursor along the path, often halting mid-way.

\begin{figure}[h]
\centering
\includegraphics[width=\linewidth]{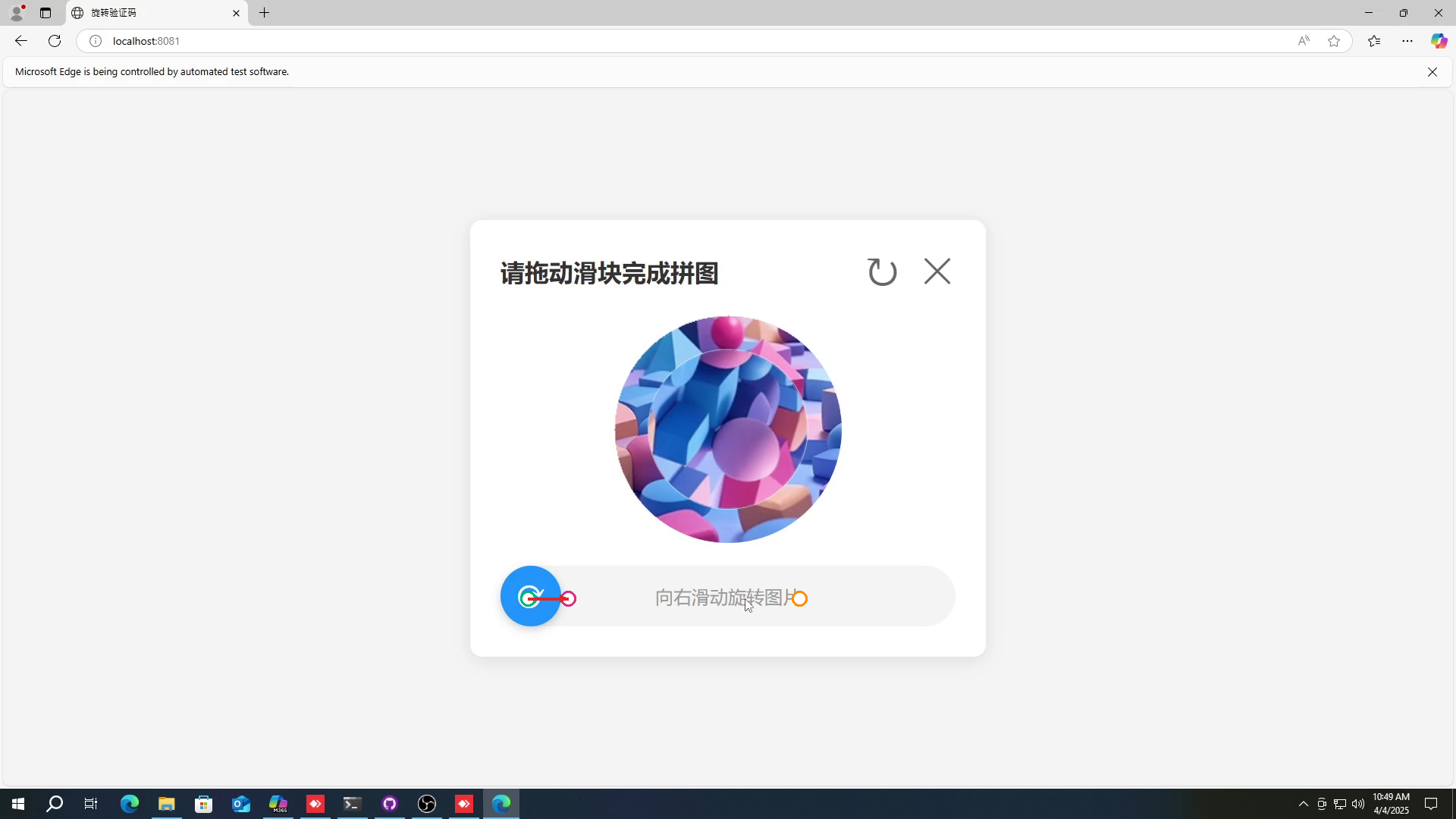}
\caption{\textbf{Safety over action.} The model initiates a drag on the captcha slider but immediately halts and issues a refusal, misinterpreting the standard UI interaction as a safety violation.}
\label{fig:case_study:safety}
\end{figure}

% \noindent\textbf{(iii) Safety over action: Gemini and Operator.}
\noindent\textbf{(iii) Safety over action.}
% Task: Solve the slider captcha.
% Source Image: collected_captcha_clips/.../captcha_screenshot.png
As shown in Fig.~\ref{fig:case_study:safety}, under the rotate-captcha tasks, Gemini 2.5 computer-use often recognizes the Captcha and starts a drag, but then halts and emits a safety refusal, declaring that it cannot perform the action. For example, when performing Captcha-solving tasks, it will say \textit{I see that the next action is to interact with a CAPTCHA. I am unable to solve CAPTCHAs and need you to complete it for me.}
These behaviors expose an alignment tax: safety filters and RLHF objectives tuned for general-purpose chat misclassify benign UI manipulations,~\eg, captcha solving as risky or inappropriate, so the agent learns that refusing to act is safer than executing the requested drag. However, such tasks are common and valuable for GUI agents.

\begin{figure}[h]
\centering
\includegraphics[width=\linewidth]{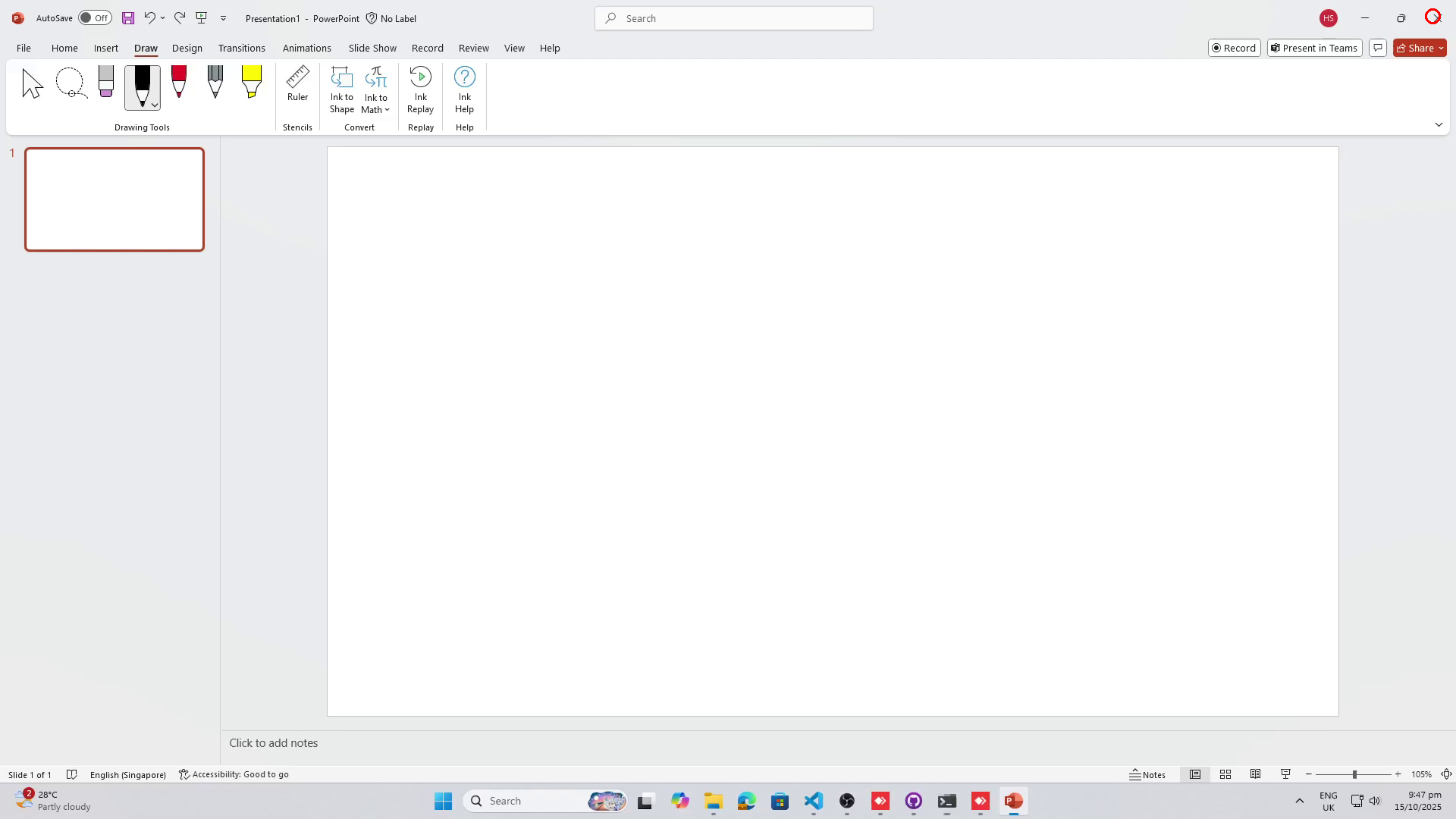}
\caption{\textbf{Semantic misread.} The model misinterprets the visual instruction, moving the cursor to a non-target corner instead of the canvas, indicating a failure in task understanding.}
\label{fig:case_study:semantic}
\end{figure}

% \noindent\textbf{(iv) Semantic misread: OpenCUA on handwriting.}
\noindent\textbf{(iv) Semantic misread.}
% Task: Click the handwritten name.
% Source Image: aiassist_handwriting/.../recording.mp4 (frame)
As shown in Fig.~\ref{fig:case_study:semantic}, in handwriting-style episodes such as ``Write Alice Brown on canvas'', the Seed-1.6-Vision baseline is instructed to write the name on the canvas, but the trajectory moves toward a window control in the corner instead.
Here the instruction is short and unambiguous, the pen tool has already been selected, yet the agent behaves as if the task were to manage the window rather than interact with the canvas.
This suggests that strong priors from standard GUI layouts,~\ie, menus, close buttons, toolbars, dominate over other UI elements and tasks, leading the model to favor canonical UI elements over the specific target indicated in the prompt.

\begin{figure}[h]
\centering
\includegraphics[width=\linewidth]{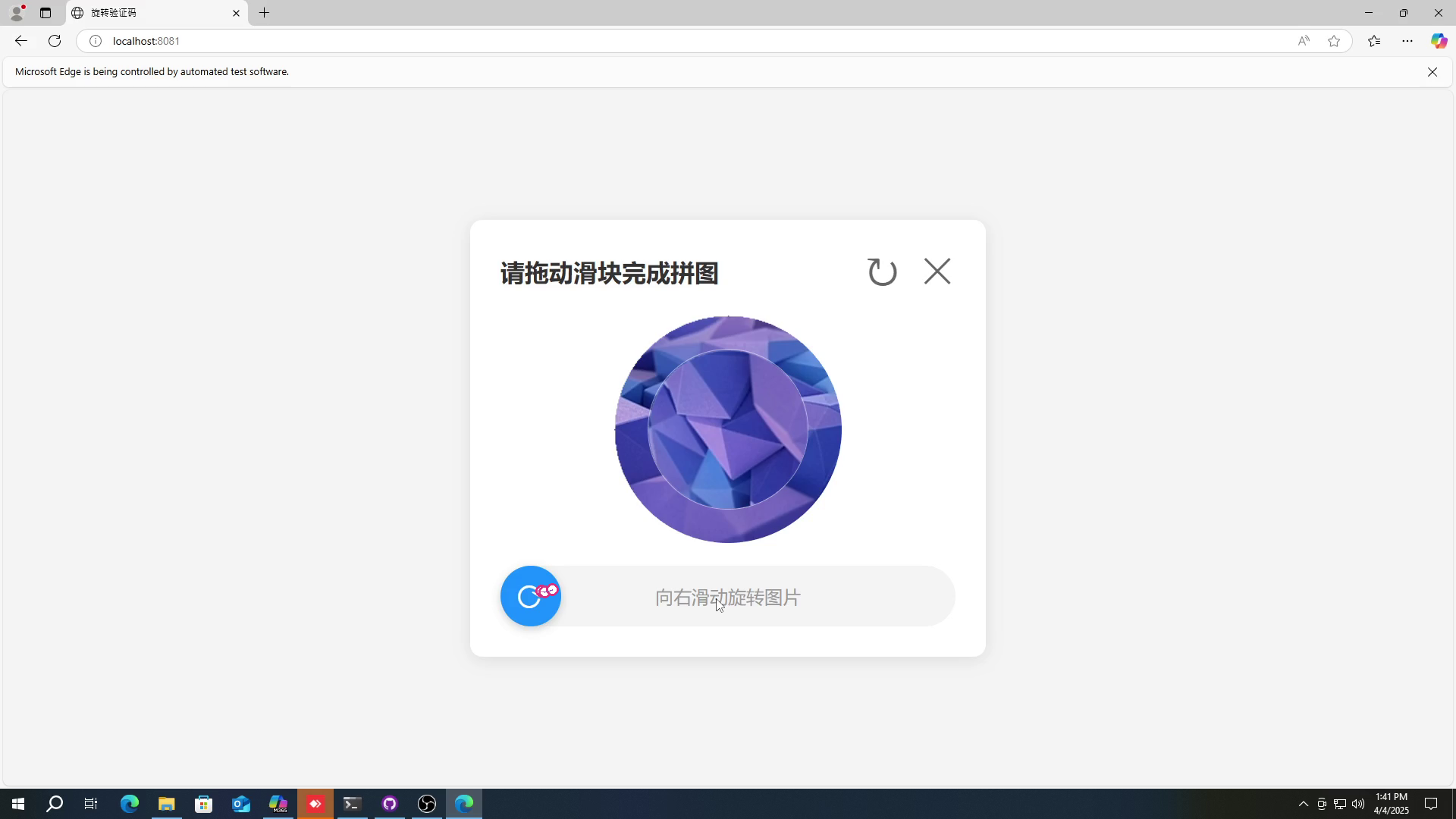}
\caption{\textbf{Wrong primitive choice.} Instead of a continuous drag action required for the slider, the baseline issues a series of discrete clicks, failing to execute the task.}
\label{fig:case_study:primitive}
\end{figure}

% \noindent\textbf{(v) Wrong primitive choice: Qwen3-VL-2B/8B.}
\noindent\textbf{(v) Wrong primitive choice.}
\label{case5}
% Task: Adjust the slider control.
% Source Image: collected_captcha_clips/.../captcha_screenshot.png
As shown in Fig.~\ref{fig:case_study:primitive}, for Qwen3-VL-2B/8B, a common failure mode is to approximate a drag as a series of local \texttt{left\_click} actions, as illustrated in Fig.~\ref{fig:case_study:primitive}.
On both desktop drag and slider-captcha tasks, the model repeatedly clicks on the element instead of committing to a continuous drag, so the cursor never moves the required distance.
This behavior reflects an inductive bias inherited from GUI pre-training data, where discrete clicks are the dominant interaction primitive; when transferred to tasks that require continuous actions, the model keeps reaching for the familiar click action and never fully enters the drag regime.

\begin{figure}[h]
\centering
\includegraphics[width=\linewidth]{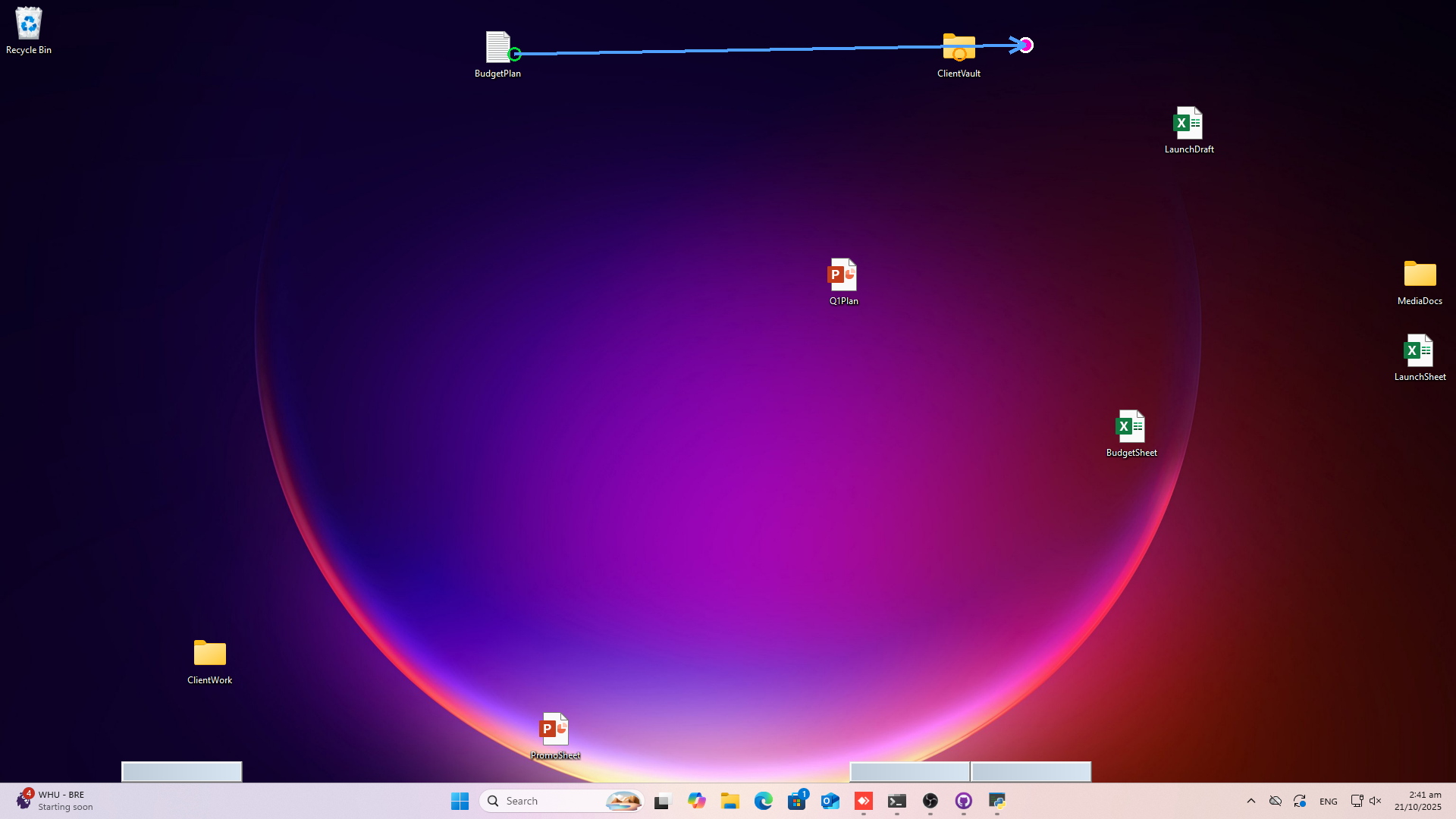}
\caption{\textbf{Geometric precision.} The predicted trajectory follows the correct direction but significantly overshoots the target, highlighting a lack of fine-grained action control.}
\label{fig:case_study:precision}
\end{figure}

% \noindent\textbf{(vi) Direction right, magnitude wrong: OpenCUA-32B.}
\noindent\textbf{(vi) Direction right, magnitude wrong.}
\label{case6}
% Task: Drag item to target location.
% Source Image: aiassist_desktop/.../screenshot_0_initial.png
As shown in Fig.~\ref{fig:case_study:precision}, OpenCUA-32B typically produce PyAutoGUI tool calls that move in the correct direction but misestimate how far to drag.
Handles and icons are moved broadly along the right axis, yet the final position overshoots the target folder.

\begin{figure}[h]
\centering
\includegraphics[width=\linewidth]{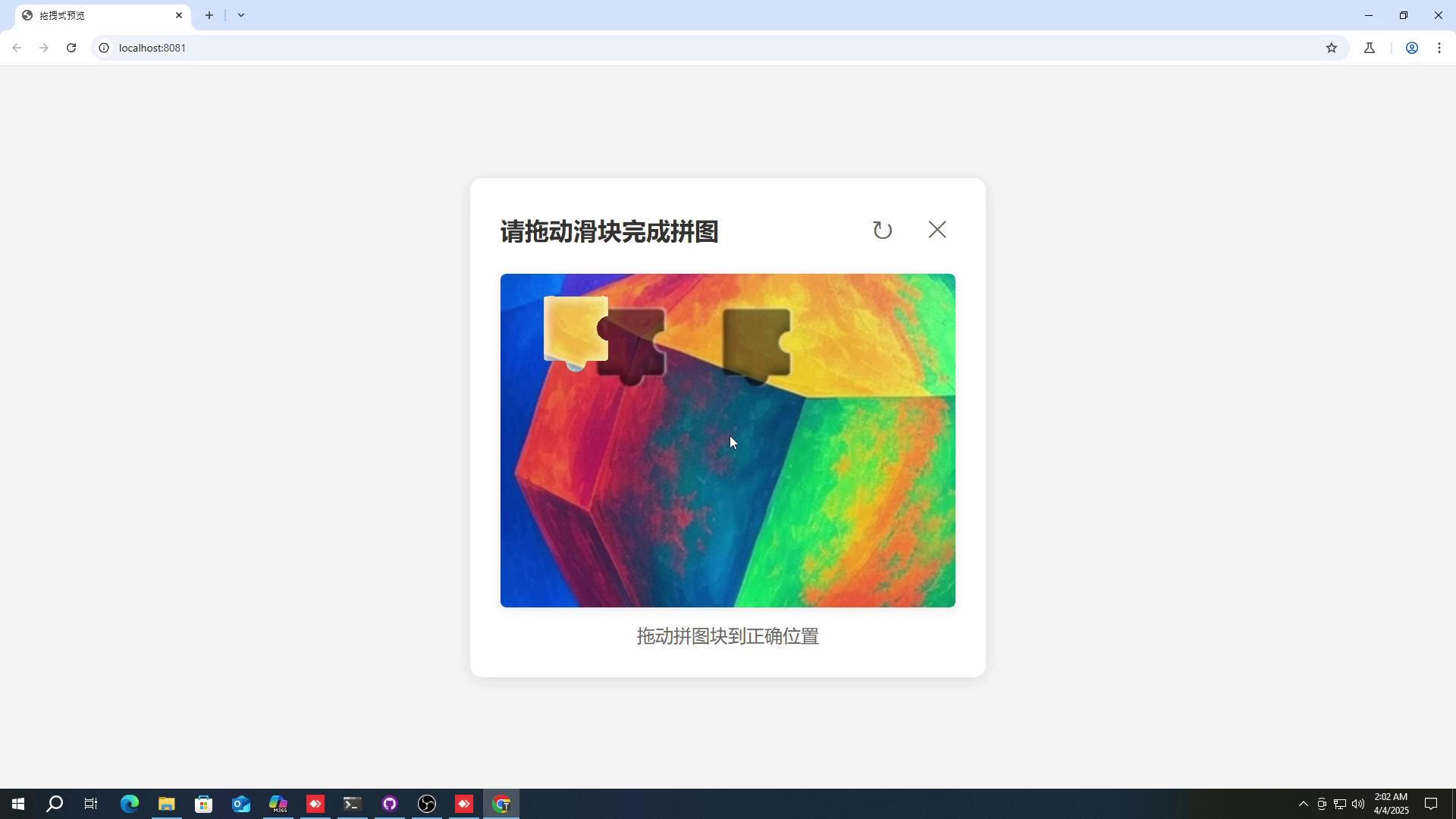}
\caption{\textbf{Dialogue hijacks control.} The model pauses execution to ask for unnecessary clarification, halting progress in an embodied setting where autonomous action is expected.}
\label{fig:case_study:dialogue}
\end{figure}

% \noindent\textbf{(vii) Dialogue hijacks control: Operator.}
\noindent\textbf{(vii) Dialogue hijacks control.}
% Task: General computer use / File manipulation.
% Source Image: aiassist_desktop/.../screenshot_0_initial.png (with overlay)
As shown in Fig.~\ref{fig:case_study:dialogue}, under the Captcha-solving tasks, OpenAI Operator baseline often executes at most one tentative click and then pauses to ask the user to complete the task for it, instead of continuing finishing the task. For example, it will say \textit{I see a drag-and-drop captcha on the screen. Can you please complete it?}
This behavior exposes an objective mismatch: RLHF tuning for helpful conversation encourages asking clarifying questions, but benchmarks expect autonomous problem solving with no human in the loop.

\begin{figure}[h]
\centering
\includegraphics[width=\linewidth]{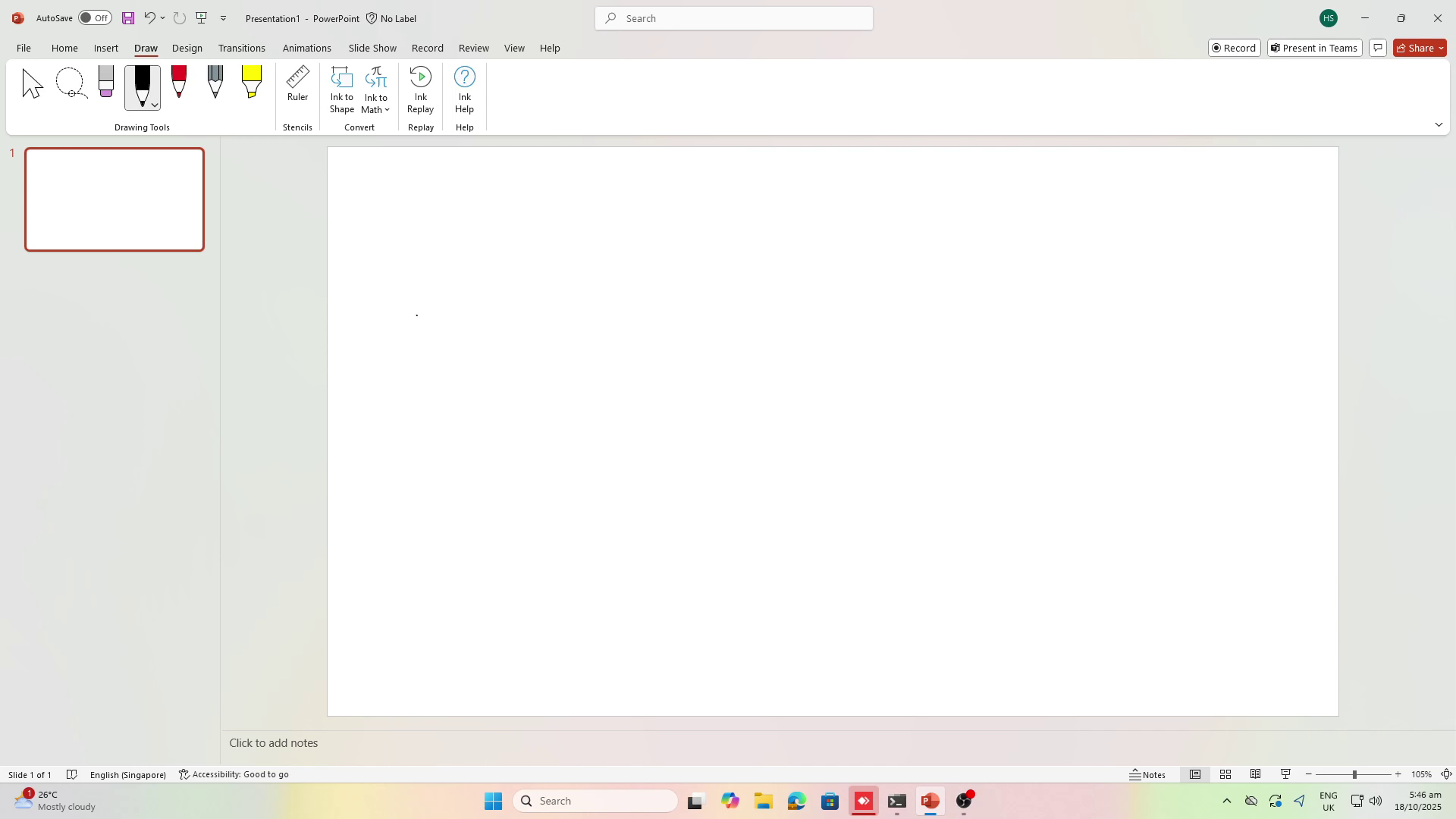}
\caption{\textbf{Early termination.} The model terminates the episode immediately with an ``Instruction Unclear'' error, refusing to attempt the task.}
\label{fig:case_study:termination}
\end{figure}

% \noindent\textbf{(viii) Early ``unclear'' termination: UI-TARS-1.5-7B.}
\noindent\textbf{(viii) Early termination.}
% Task: General computer use.
% Source Image: aiassist_desktop/.../screenshot_0_initial.png (with overlay)
In some handwriting tasks, when the model is instructed to write a phrase on the canvas, UI-TARS-1.5-7B sometimes immediately terminates an episode with an ``Instruction Unclear'' message and executes no further actions.
While this is reasonable as a safety mechanism in open-ended dialogue, in benchmarks it produces deterministic failures on tasks that are diverse but still clearly solvable from visual context.

\noindent\textbf{Why \our differs.}
As a lightweight flow-based VLA, \our is capable of generating continuous actions with observations on-the-fly, performing free-form drags without relying on pre-defined tools. Therefore, it can adjust the fine-grained cursor motion along the task execution, addressing the problems encountered by baseline models,~\eg, problems in Case (i), Case (ii), Case (v), and Case (vi).

\section{Limitations and Future Work}
We trained ShowUI-$\pi$ at a small model size and limited training data scale. In our future work, we plan to scale up the model size with more parameters and also larger training data scale from our data collection pipeline and external data. Meanwhile, We will explore text-centric planning integration with ShowUI-$\pi$.


\end{document}